\documentclass[lettersize,journal]{IEEEtran}
\usepackage{amsmath,amsfonts}
\usepackage{algorithmic}
\usepackage{algorithm}
\usepackage{array}
\usepackage[figuresright]{rotating}
\usepackage{textcomp}
\usepackage{stfloats}
\usepackage{url}
\usepackage{verbatim}
\usepackage{graphicx}
\usepackage{cite}
\usepackage{xspace}
\usepackage{soul} %
\usepackage{color, xcolor} %
\usepackage{tabularx}

\usepackage{adjustbox}
\usepackage{multirow}
\usepackage{arydshln}       
\usepackage{makecell}      

\usepackage[T1]{fontenc}
\ifCLASSOPTIONcompsoc
\usepackage[caption=false, font=normalsize, labelfont=sf, textfont=sf]{subfig}
\else
\usepackage[caption=false, font=footnotesize]{subfig}

\DeclareMathOperator*{\argmax}{arg\,max}

\begin{document}
\title{\LARGE \bf Multi-Agent Reinforcement Learning for Connected and Automated Vehicles Control: Recent Advancements and Future Prospects}

\author{Min Hua$^1$, Xinda Qi$^2$,  Dong Chen$^3$, Kun Jiang$^4$, Zemin Eitan Liu$^5$, Quan Zhou$^{1,6}$, Hongming Xu$^1$$^*$

\thanks{$^1$Min Hua, Quan Zhou, and Hongming Xu are with the School of Engineering, University of Birmingham, Birmingham, B15 2TT, UK. (e-mail: mxh623@student.bham.ac.uk, q.zhou@bham.ac.uk, h.m.xu@bham.ac.uk); $^*$ is the corresponding author.}

\thanks{$^2$Xinda Qi is with the Department of Electrical and Computer Engineering, Michigan State University, MI, USA. (e-mail: qixinda@msu.edu).}

\thanks{$^3$Dong Chen is with the Environmental Institute \& Link Lab \& Computer Science, University of Virginia, VA, USA. (e-mail: dqc4vv@virginia.edu).}

\thanks{$^4$Kun Jiang is with the School of Automation, Southeast University, Nanjing, China (e-mail: kjiang@seu.edu.cn).}

\thanks{$^5$Zemin Eitan Liu is with 
the Chemical and Petroleum Engineering Department, University of Pittsburgh, Pittsburgh, PA, USA. (e-mail: eliuzm@163.com).}

\thanks{$^6$Quan Zhou is also with the School of Automotive Studies, Tongji University, Shanghai 201804, China}

}
\markboth{Journal of \LaTeX\ Class Files,~Vol.~14, No.~8, August~2021}%
{Shell \MakeLowercase{\textit{et al.}}: A Sample Article Using IEEEtran.cls for IEEE Journals}


\maketitle

\begin{abstract}
Connected and automated vehicles (CAVs) are considered as a potential solution for future transportation challenges, aiming to develop systems that are efficient, safe, and environmentally friendly. 
However, CAV control presents significant challenges due to the complexity of interconnectivity and coordination required among vehicles.
Multi-agent reinforcement learning (MARL), which has shown notable advancements in addressing complex problems in autonomous driving, robotics, and human-vehicle interaction, emerges as a promising tool to enhance CAV capabilities.
Despite its potential, there is a notable absence of current reviews on mainstream MARL algorithms for CAVs. To fill this gap, this paper offers a comprehensive review of MARL’s application in CAV control. The paper begins with an introduction to MARL, explaining its unique advantages in handling complex and multi-agent scenarios. It then presents a detailed survey of MARL applications across various control dimensions for CAVs, including critical scenarios such as platooning control, lane-changing, and unsignalized intersections.
Additionally, the paper reviews prominent simulation platforms essential for developing and testing MARL algorithms. Lastly, it examines the current challenges in deploying MARL for CAV control, including macro-micro optimization, communication, mixed traffic, and sim-to-real challenges. Potential solutions discussed include hierarchical MARL, decentralized MARL, adaptive interactions, and offline MARL.
\vspace{5pt}

\textbf{\textit{Note to Practitioners}—This paper presents multi-agent reinforcement learning as a solution to the challenges of controlling connected and automated vehicles in complex scenarios like platooning, lane changes, and intersections. MARL offers a more adaptive approach than traditional methods, potentially improving traffic flow, safety, and fuel efficiency in real-world settings and leading to more efficient and reliable transportation systems. The paper also reviews current MARL algorithms and simulation platforms, providing a resource for implementing these advanced control strategies in practice. However, the deployment of MARL in real-world CAV systems is still in its early stages and faces challenges such as ensuring reliable communication between vehicles and managing mixed traffic environments that include both automated and human-driven vehicles. Future research is needed to address these challenges and validate the approach in diverse and dynamic traffic conditions. This paper serves as a stepping stone for practitioners aiming to develop more reliable and adaptive CAV systems in the near future.}
\end{abstract}


\section{Introduction}
\label{sec:introduction}
\IEEEPARstart{T}{he} transportation industry is experiencing a transformative shift driven by advancements in automation, artificial intelligence (AI), the Internet of Things (IoT), and sensor technologies \cite{dimitrakopoulos2010intelligent, kehoe2015survey}. A central element of this shift is the development of automated vehicles (AVs), which promise to mitigate traffic congestion, enhance road safety, and increase accessibility \cite{proia2021control, bi2021hybrid}. The evolution of this field has led to the emergence of connected and automated vehicles (CAVs), which integrate AVs into a cohesive, networked transportation system through advanced communication technologies such as vehicle-to-vehicle (V2V), vehicle-to-infrastructure (V2I), and vehicle-to-cloud (V2C) \cite{wu2023finite}. This integration significantly expands the operational capabilities of AVs, enabling extended perception, enhanced collaborative decision-making, and improved traffic efficiency \cite{li2023traffic, hua2020research}. However, the realization of an effective CAV framework that harmonizes control, computing, and communication presents substantial challenges, particularly in managing the intricate interactions between AVs and other road users, including pedestrians, cyclists, and human-driven vehicles (HDVs) \cite{he2020admittance}. Addressing these challenges is critical for advancing automation in transportation and ensuring the safe, efficient deployment of CAV systems.


Control of CAVs is pivotal in enhancing transportation safety and sustainability, integrating with disciplines like energy management, urban planning, and social contexts \cite{garg2022systematic, hua2019hierarchical}. Conventional optimization methods have addressed CAV control challenges, such as using model predictive control (MPC) to coordinate platoon behavior, minimizing control delay, and reducing traffic oscillations \cite{lin2020comparison}. Additionally, non-convex optimal control problems, such as CAV coordination at intersections, have been tackled with semidefinite relaxation techniques \cite{katriniok2022fully}. However, these approaches often depend on precise system modeling, which is not always available \cite{chen2024communication}, and require substantial computational resources, making them impractical for real-time CAV control \cite{liu2023systematic}.

Besides, reinforcement learning (RL) has gained increasing attention within the research field due to its outstanding abilities in addressing sequential decision-making tasks, such as gaming \cite{silver2016mastering}, robotics \cite{lee2020learning, bai2022group}, behavioral planning \cite{wang2024prescribed}, intelligent energy management \cite{ganesh2022review, liu2023safe, liu2024deep}. Similarly, the AV control has begun exploring the potential of employing RL for various traffic scenarios \cite{chen2020autonomous, shu2021driving}.  For instance, in \cite{chen2020autonomous}, a safe RL framework based on an improved double deep Q-Network (DDQN) \cite{van2016deep} is proposed for highway lane-changing, resulting in zero collisions. In \cite{zhou2019development}, a RL-based model for plattoning control is introduced to enhance fuel economy, driving efficiency,  and safety at signalized intersections through real-time optimization. This approach featuring an effective reward function, demonstrates strong performance under varying traffic demands and traffic light cycles with different durations. Taking into account the role of HDVs in the control of AVs, Shi et al. employ a distributed proximal policy optimization (DPPO) control strategy that allows them to learn from and respond to the behavior of the HDVs, optimizing performance at both the local subsystem level and the broader mixed traffic context \cite{shi2021connected}. Qu et al. introduce a control approach utilizing the deep deterministic policy gradient (DDPG) algorithm aimed at reducing traffic fluctuations and enhancing fuel economy. However, these approaches only consider a single agent (i.e., AV) and exhibit poor generalization when multiple agents are involved. Additionally, their study focuses solely on individual vehicle control and does not leverage shared information \cite{qu2020jointly}.

Multi-agent reinforcement learning (MARL) represents a significant research direction within artificial intelligence, extending the concepts of single-agent RL to scenarios involving multiple interacting agents. This innovative approach has found applications across a broad spectrum of fields. Notably, the gaming and simulation industry has utilized MARL to create more complex and interactive environments \cite{chandra2022gameplan}. Furthermore, within the realm of finance, MARL models are employed to simulate the dynamics of markets and the behaviors of agents \cite{amrouni2021abides}.
In traffic control and management, MARL algorithms have been instrumental in optimizing traffic flows and reducing congestion. The realm of AVs and robotics has also benefited greatly, with MARL enabling cooperative navigation and coordination among vehicles and facilitating collaborative tasks in AVs and robotics \cite{yang2022intelligent}. Extended to CAV control, each vehicle acts as an independent agent that learns from its interactions with the surrounding environment, which includes other vehicles, traffic signals, and road conditions. Each CAV learns not only to maneuver safely and efficiently on its own but also to cooperate and communicate with other CAVs \cite{palacios2023enhancing}.
Therefore, MARL offers a valuable strategy for improving the performance and efficiency of CAVs, paving the way for more sophisticated, collaborative, and flexible transportation networks.

\begin{figure*}[!ht]
\centerline{\includegraphics[width=0.8\textwidth]{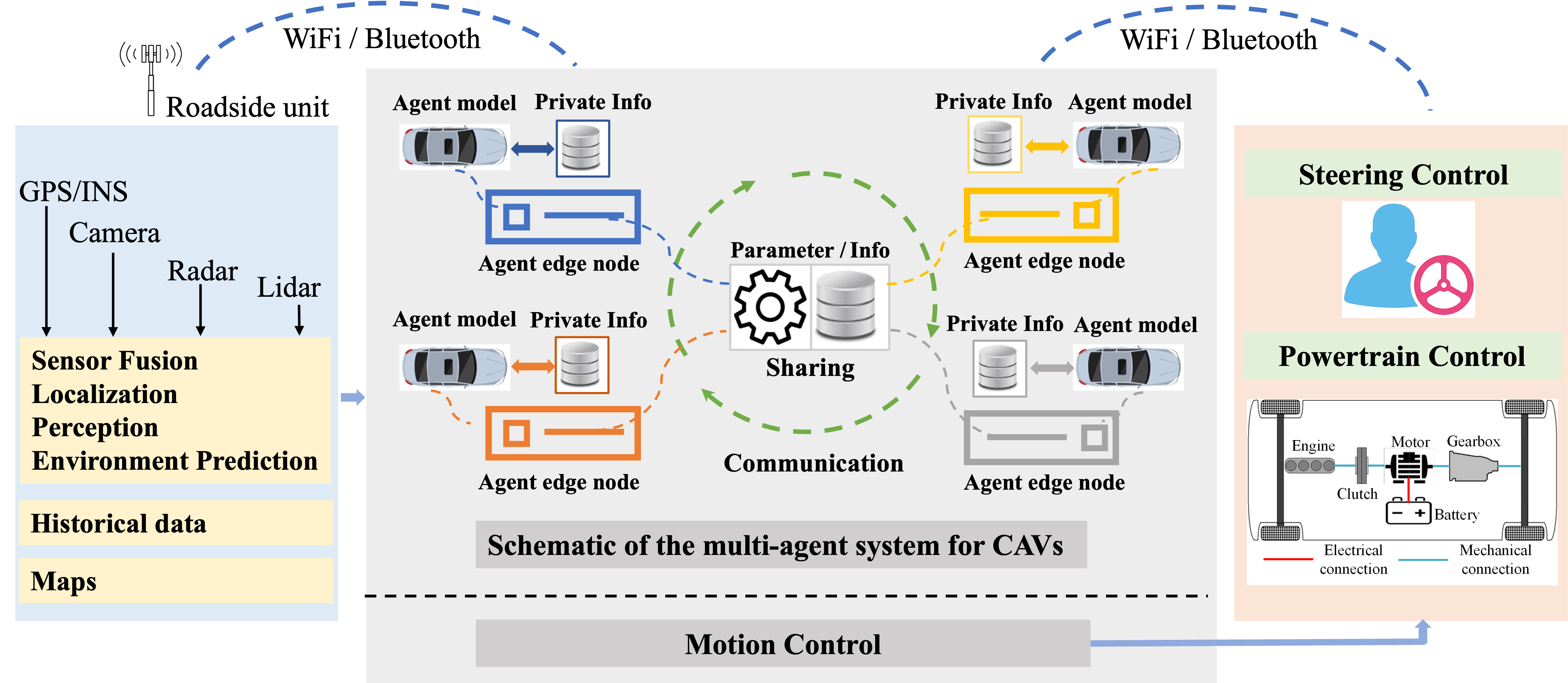}}
\caption{The multi-agent control system for CAVs: the left side represents the information inputs of the overall control system, while the right side depicts the studied multi-agent control system for CAVs.}
\label{overall}
\vspace{-10pt}
\end{figure*}

In scenarios involving multiple agents, MARL has emerged as a promising tool for optimizing the collective behaviors, showing significant potential for advancing the control strategies employed in the CAV ecosystem \cite{chen2023deep}. Fig. \ref{overall} illustrates a comprehensive multi-agent system, i.e., CAVs, and depicts a holistic approach to CAV operation, from sensory input and data processing to inter-agent communication and eventual decision-making and motion execution. On the left, real-time environmental data from various sensors, including GPS/INS, cameras, radar, and lidar, is gathered and then integrated via sensor fusion. 
Within multi-agent system, each AV maintains its own data repository while accessing shared information, facilitating inter-agent communication for enhanced decision-making and situational awareness \cite{chen2024data}. 
This system includes modules for decision-making and motion planning, which use data and inter-vehicle communication to make decisions and plan movements. Finally, the steering control and powertrain control modules demonstrate how these decisions translate into physical actions, such as turning the steering wheel or adjusting the engine's or motor operation. 

A good example of large-scale traffic signal control is developed  \cite{chu2019multi}  through a highly scalable and decentralized MARL algorithm, demonstrating superior control performance over the other leading decentralized MARL algorithms. Furthermore, in \cite{chen2023deep}, the challenge of on-ramp merging in mixed traffic scenarios has been explored with a scalable and safe MARL algorithm that utilizes a parameter-sharing technique to enhance safety and scalability. Despite recent progress, the development of efficient and scalable MARL algorithms still presents significant challenges. The primary concern is guaranteeing the robustness and safety of these algorithms, especially in the face of diverse and uncertain traffic scenarios \cite{chen2023deep}. Furthermore, the dynamic behaviors of CAVs lead to non-stationary environments, adding layers of complexity to the development of MARL algorithms \cite{chu2019multi}. Another challenge lies in the lack of realistic simulators capable of accurately modeling traffic scenarios and other road users, which are essential for effective training of MARL algorithms \cite{zhou2020smarts}. Lastly, the real-world implementations of these algorithms remain an uphill task due to sim-to-real gaps \cite{ chen2023deep}, and technological, legislative, and societal barriers . While a few articles describe the potential of MARL in CAVs \cite{hernandez2019survey,  yadav2023comprehensive}, none of them, to the best of our knowledge, are specifically devoted to the application of MARL for CAV control.

This paper is thus intended to deliver a comprehensive and systematic review of MARL within the realm of CAVs, such as lane changing, platooning control, traffic signal cooperation, and on-ramp merging.
Through this review, we not only provide a clear, up-to-date understanding of the current classic MARL algorithms applied in CAV control, but we also outline the potential direction for future work in this area. The main contributions of the work presented in this paper are summarized as follows:
\begin{enumerate}
    \item We survey the recent developments in MARL algorithms, discussing their diverse applications in aspects of CAV control based on the extent of control dimensions. Furthermore, we provide an extensive examination of the leading simulation platforms employed in MARL research for CAVs.
    \item We highlight and explore the technical challenges that MARL faces in these applications and discuss potential research directions to address these challenges.
    \item The work, including a well-classified list of relevant papers, has been presented at the following site: \url{https://github.com/huahuaedi/MARL_in_CAV_control_review}.
\end{enumerate}

This paper is organized as follows: Section~\ref{sec:backgrounds} introduces the basics of RL and MARL; A detailed review of MARL applied in CAV control from different control dimensions and mainstream simulation platforms are described in section~\ref{sec:app}. Section~\ref{sec:dis} summarizes the remaining challenges and opportunities. Finally, Section~\ref{sec:conl} concludes the review.

\section{Methodologies}\label{sec:backgrounds}
In this section, we begin with a comprehensive overview of the fundamental principles of RL. We then delve into various prominent MARL algorithms, establishing the context to enhance the understanding of our review.

\subsection{Preliminaries of RL}
RL, frequently modeled through a mathematical framework known as a Markov Decision Process (MDP), has risen as an effective method for data-driven sequential decision-making \cite{kaelbling1996reinforcement}.
Recently, deep neural networks (DNN) have substantially elevated the capability of RL to manage complex problems \cite{mnih2015human}. Key developments include sophisticated algorithms such as the Deep Q-Network (DQN) \cite{mnih2015human}, Deep Deterministic Policy Gradient (DDPG) \cite{lillicrap2015continuous}, and Advantage Actor-Critic (A2C) \cite{mnih2016asynchronous}. For instance, the AlphaStar, which operates on principles similar to DQN, marked a significant milestone by outperforming professional esports players in StarCraft II, illustrating the power of RL in strategic game environments \cite{vinyals2019grandmaster}. Moreover, the development of the autonomous driving system by leveraging DDPG exemplifies the potential of RL, showcasing how vehicles can autonomously navigate through environments with dynamic obstacles \cite{10388472}. 

\begin{figure}[htbp]
\centering
\includegraphics[width=0.9\columnwidth]{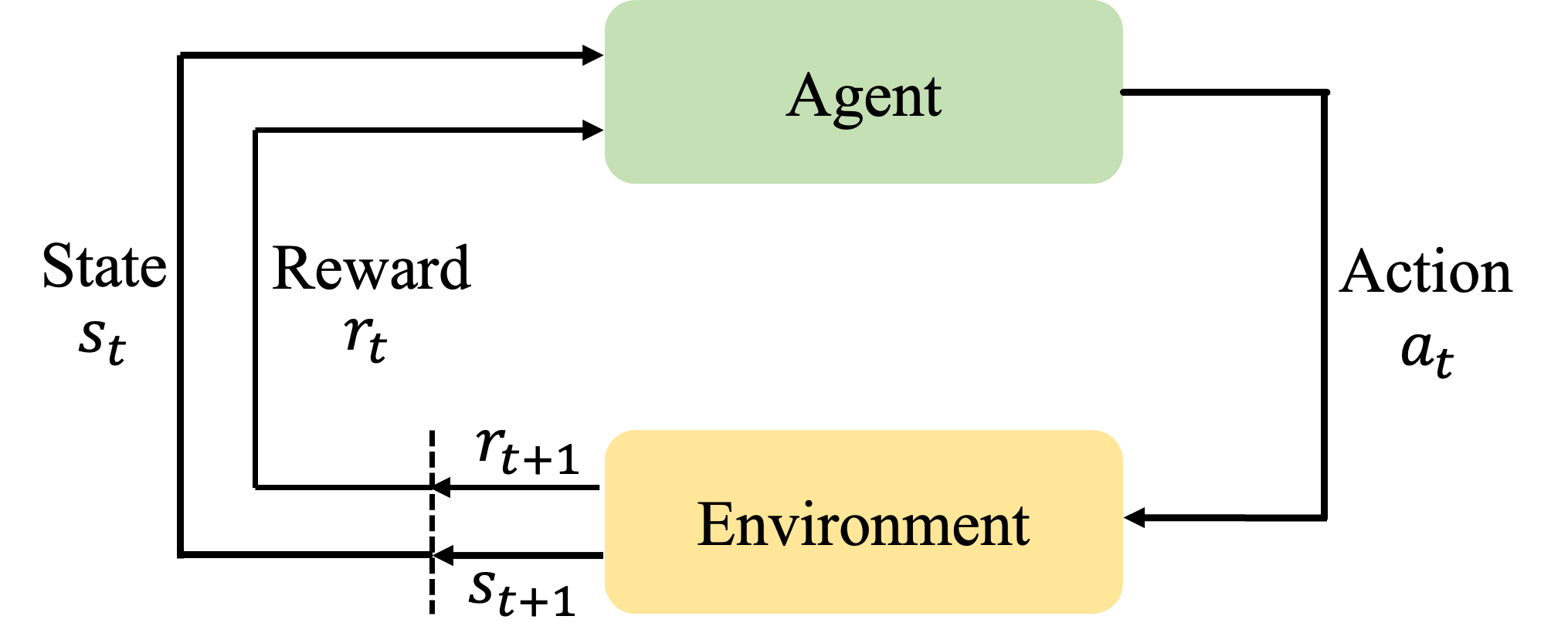}
\caption{Illustration of reinforcement learning (RL). }
\label{fig1}
\end{figure}

In the context of a RL framework (as shown in Fig. \ref{fig1}), a learner known as the \textbf{agent} navigates through the \textbf{environment} using a process of trial and error. This agent assesses situations, undertakes specific \textbf{actions} in the environment, and consequently receives feedback in the form of a \textbf{reward} signal and a \textbf{new state}. The feedback, provided by the environment, serves as an indicator to the agent about the effectiveness of its actions, signaling whether they have a positive or negative impact. 
Many RL problems are framed as \textbf{MDP}, which provide a mathematical framework for modeling decision making where outcomes are partly random and partly under the control of a decision maker. MDPs are characterized by states, actions, transition dynamics, and rewards
with $\mathcal{M} = (\mathcal{S}, \mathcal{A}, \mathcal{P}, \mathcal{R})$, where it is defined as follows:


\begin{enumerate}
\item State space $\mathcal{S}$:  a representation of the environment at a time step$t$. It can include all information necessary for the agent to make a decision. 
An \textbf{observation} $o$ offers an incomplete state description, potentially missing full details. In \textbf{fully observed} environments, the observation is the same as the state. In \textbf{partially observed} environments, the observation might contain less information than the state. 



\item Action space $\mathcal{A}$: the set of possible moves or decisions the agent can make in a given state. The set of actions available can depend on the state. The action spaces fall into two categories:  \textbf{discrete} or \textbf{continuous}.


\item Reward $\mathcal{R}(s_t, a_t, s_{t+1})$: a scalar feedback signal given to the agent from the environment after performing action $a_t$ in state $s_t$. The agent's objective is to maximize the cumulative reward over time.

\item Transition Probability $\mathcal{P}_{ss'}(S_{t+1}=s' | S_{t}=s)$: the chance of an agent transitioning from one state to a different state.
\end{enumerate}


In the following subsections, we will explore three prevalent RL algorithms, including Value-Based Methods, Policy-Based Methods, and Actor-Critic Methods.

\subsubsection{\textbf{Value-Based Methods}}
In the domain of value-based methods within RL, the Q-function, represented as $Q_{\theta}$, is usually parameterized by a set of parameters denoted as $\theta$. This parameterization facilitates the use of different function approximators for estimating Q-values, including traditional Q-tables \cite{watkins1992q} and the complex Deep Neural Networks (DNN) \cite{mnih2015human}. The core mechanism for updating these parameters relies on the temporal difference (TD), expressed as  $(\mathcal{T}Q_{\theta^-} - Q_{\theta})(s_t, a_t)$, where $\mathcal{T}$ signifies the dynamic programming operator and $Q_{\theta^-}$ represents a version of the model parameters $\theta^-$ that is temporarily fixed to stabilize updates \cite{chu2019model}. To mitigate the variance in Q-value estimates and enhance the exploration ability of the algorithm, strategies such as the $\epsilon-\text{greedy}$ method and experience replay mechanism are frequently employed within deep Q-learning frameworks \cite{szepesvari2022algorithms}. The optimal action 
$a^*(s)$ can be described as: 
\begin{equation}\label{eqn:optimal_a}
a^*(s) = \argmax_a{Q^*(s_t=s, a_t=a)},
\end{equation}

Some widely recognized deep Q-learning-based algorithms include DQN \cite{mnih2013playing}, double deep Q-Network (DDQN)\cite{van2016deep}, and hindsight experience replay (HER) \cite{andrychowicz2017hindsight}.

\subsubsection{\textbf{Policy-Based Methods}}
Diverging from the value-based Q-learning approach, policy-based methods prioritize direct manipulation of the policy itself, denoted as $\pi_\theta$, via a distinct set of parameters $\theta$. The primary goal of adjusting $\theta$ revolves around amplifying the likelihood of selected actions along the overall rewards accumulation. This process is guided by a specifically defined loss function, which is mathematically represented as:
\begin{equation}\label{eqn:policy_loss}
\nabla_{\theta} \mathcal{L} (\pi_{\theta}) =\mathop{\mathbb{E}}_{\tau \sim \pi_{\theta}}[\sum_{t=0}^T \nabla_{\theta} \log \pi_{\theta} (a_t|s_t) R_t],
\end{equation}
Subsequently, the parameters of the policy network are updated by employing stochastic gradient ascent, facilitating an iterative improvement of the policy, as detailed by
\begin{equation}\label{eqn:gradient_ascent}
\theta_{k+1} = \theta_{k} + \alpha \nabla_{\theta} \mathcal{L} (\pi_{\theta}),
\end{equation}
In contrast to Q-learning methods, policy gradient techniques demonstrate enhanced resilience to the non-stationary dynamics characterizing individual trajectories, albeit at the cost of potentially higher variance in outcomes \cite{chu2019multi}. Among the most prominent algorithms that embody the principles of policy gradients are Deep Deterministic Policy Gradient (DDPG) \cite{lillicrap2015continuous}, Soft Actor-Critic (SAC) \cite{haarnoja2018soft}, and Asynchronous Advantage Actor-Critic (A3C) \cite{mnih2016asynchronous}.


\subsubsection{\textbf{Actor-Critic Methods}}
Actor-critic methods aim to mitigate the high variability in outcomes often seen with policy gradient techniques by integrating an advantage function. This function enhances the policy gradient methodology by utilizing both the policy update (actor) and a value estimation function (critic) \cite{mnih2016asynchronous}. The advantage function, a pivotal element in this approach, is defined as:
\begin{equation}\label{eqn:advantage_function}
A^\pi (s_t, a_t) = Q^{\pi_\theta}(s_t, a_t) - V_w(s_t),
\end{equation}
where the update of parameters $\theta$ is guided by a policy loss function articulated as:
\begin{equation}\label{eqn:advantage_function_loss}
\nabla_{\theta} \mathcal{L}=\mathop{\mathbb{E}}_{\pi_{\theta}}[\sum_{t=0}^T \nabla_{\theta} \log \pi_{\theta} (a_t|s_t) A_t],
\end{equation}

Concurrently, the critic value function is refined through:
\begin{equation}\label{eqn:advantage_function_value_loss}
\mathcal{L}=\min_{w} \mathop{\mathbb{E}}_{\mathcal{D}}[(R_t + \gamma V_{w^-}(s_t) - V_w(s_t))^2 ],
\end{equation}
with $\mathcal{D}$ denoting the experience replay buffer, a repository for history experiences. This buffer, in collaboration with parameters from preceding iterations often employed within a target network, facilitates the learning process \cite{chen2023deep}.



Despite considerable advancements, single-agent RL algorithms frequently encounter scalability challenges, especially in complex, real-world physical environments that involve multiple agents. These difficulties stem from inherent non-stationarities and the partial observability characteristic on multi-agent scenarios \cite{chu2020multiagent}.


\subsection{Preliminaries of MARL}
\begin{figure}[!ht]
\centerline{\includegraphics[width=1.0\columnwidth]{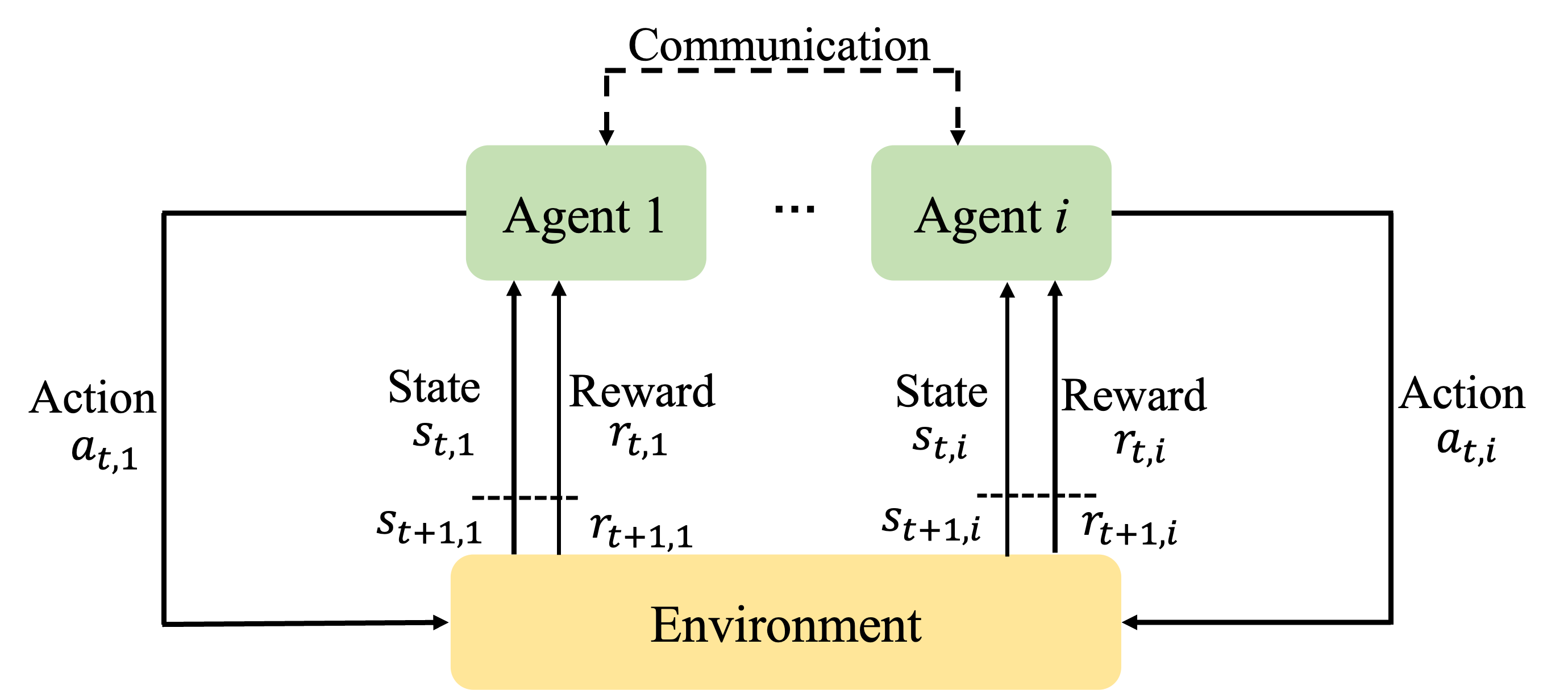}}
\caption{Illustration of multi-agent reinforcement learning (MARL).}
\label{fig3}
\vspace{-10pt}
\end{figure}

In systems with multiple agents (e.g., AVs in Fig.~\ref{fig3}), not only the reward agents themselves but also the reward of their neighbors will be influenced by their actions. MARL has been widely used in various complex systems, including managing traffic \cite{chu2019multi}, strategic gameplay \cite{berner2019dota}, optimizing wireless network resources \cite{naderializadeh2021resource}, and configuring power grids \cite{chen2021powernet}, to highlight a few areas.
A fundamental approach within this field is Independent Q-learning (IQL), where each agent's Q-function is primarily developed based on local actions, as $Q_i(s, a) \approx Q_i(s, a_i)$, simplifying the learning process. Similarly, the Independent Advantage Actor-Critic (IA2C) \cite{chu2019model} represents a variant of MARL, indicating the breadth of strategies explored in this space
However, both IQL and IA2C face challenges in applications due to the partial observability and dynamic environment, as they operate under the assumption that all other agents' actions are part of the environment, which complicates learning when agents' strategies continually evolve \cite{chu2019multi}.


Multi-agent systems are often treated as networks $\mathcal{G} = (\text{\Large $\nu$}, \text{\Large $\varepsilon$})$ without a central controller \cite{chu2019multi, chen2021powernet}, where agents $i \in \text{\Large $\nu$}$  communicate with nearby peers $\mathcal{N}_i:=\{j|\varepsilon_{ij}\in\large\text{\Large $\varepsilon$})\}$ via the edge connections $\text{$\varepsilon_{ij}$}, i \neq j$. 
This decentralized framework, where each agent only perceives part of the environment, reflects real-world scenarios like AVs detecting surrounding vehicles. This structure effectively simulates the dynamic environment as a partially observable MDP (POMDP) $\mathcal{M_G}=(\{\mathcal{A}_i, \mathcal{S}_i, \mathcal{R}_i\}_{i\subseteq \nu}, \mathcal{T})$, with the system's behavior determined by the collective actions ($\mathcal{A} =\mathcal{A}_{1}\times \mathcal{A}_{2}\times \cdots \times \mathcal{A}_{N}$) of all agents and the state transitions (with the state: $\mathcal{S} =\mathcal{S}_1\times \mathcal{S}_{2}\times \cdots\times \mathcal{S}_N$) governed by probability distributions $\mathcal{T}(s'|s, a)$. 
Rewards play a crucial role in guiding agents toward achieving goals, which vary across \textit{cooperative}, \textit{competitive}, and \textit{mixed} scenarios. In cooperative environments, agents work together towards a common goal, possibly sharing a uniform reward ($\mathcal{R}_1=\mathcal{R}_2= \cdots = \mathcal{R}_N$) or aiming to optimize an average reward that accommodates differences among agents (\textit{team-average} reward \cite{zhang2021multi}). Conversely, competitive scenarios feature agents with individualistic goals, often modeled as zero-sum games ($\bar{\mathcal{R}}=\frac{1}{N}\sum_{i \in \Large \nu} \mathcal{R}_i(s, a, s')$), where one agent's gain is the loss of another.  Mixed environments (also known as \textit{general sum} game settings) allow for both cooperation and competition, with agents pursuing individual yet not directly conflicting objectives.

The majority of applications reviewed in Section~\ref{sec:app} of this paper mainly focus on cooperative scenarios, highlighting the importance of collaboration in achieving shared objectives in multi-agent systems.

\subsection{Training and execution strategies in MARL}
In MARL settings, different approaches and frameworks are employed to tackle the complex challenges of coordinating and training agents in decentralized environments. This subsection explores two fundamental paradigms: Centralized Training with Decentralized Execution (CTDE) and Decentralized Training with Decentralized Execution (DTDE), each offering distinct insights and trade-offs in the pursuit of effective multi-agent learning \cite{wang2023hierarchical, MahajanRSW19, li2023multi}.

\subsubsection{\textbf{Centralized training with decentralized execution (CTDE)}}
\begin{figure}[htbp]
\centerline{\includegraphics[width=7cm]{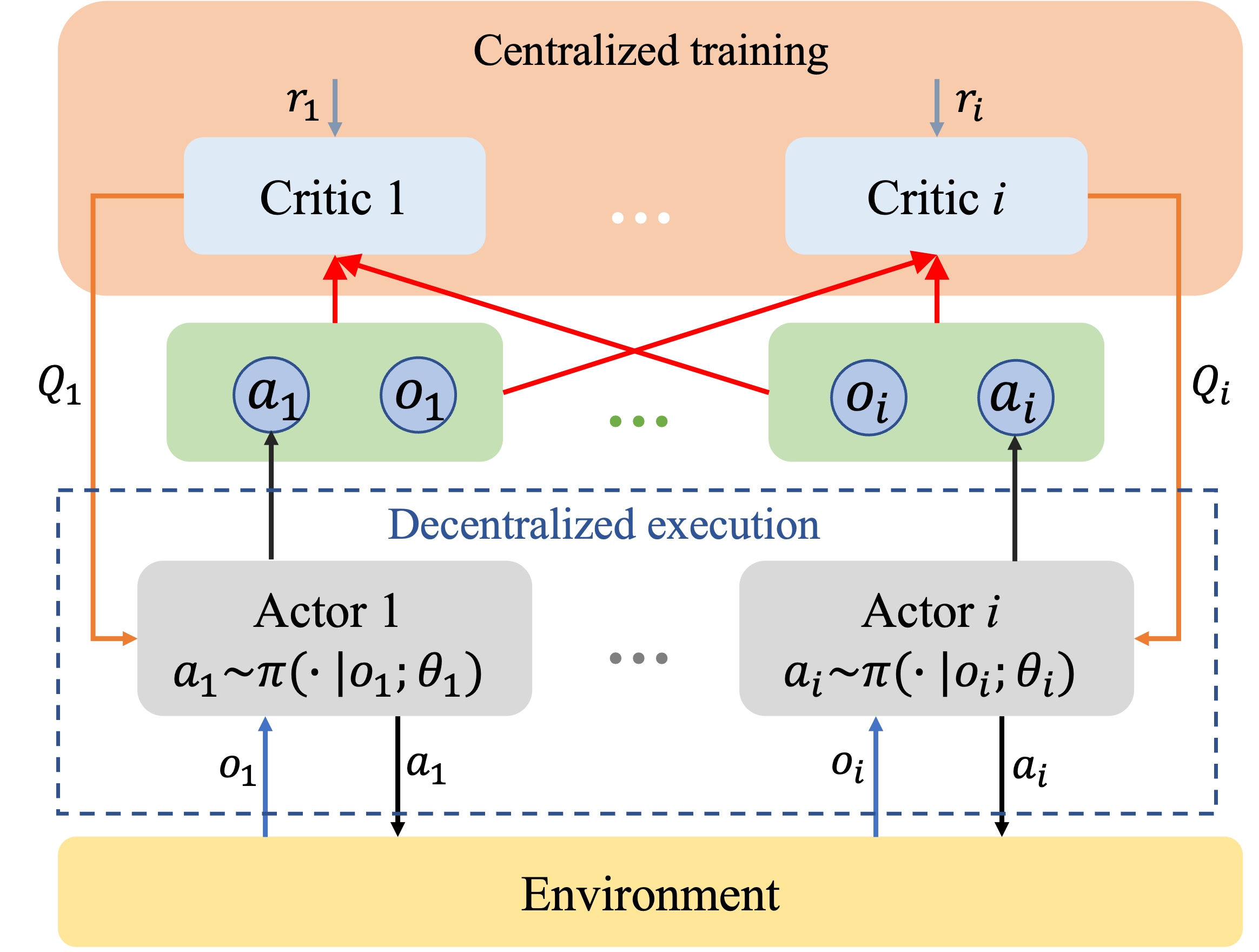}}
\caption{Illustration of centralized training with decentralized execution (CTDE) in MARL.}
\label{fig:CTDE}
\vspace{-5pt}
\end{figure}

The majority of MARL algorithms adhere to centralized training with a decentralized execution (CTDE) framework. In this framework, the decentralized problem is initially transformed into a centralized one, solvable by a central controller, which gathers essential training information, e.g., observation, reward, and global state information, for the training process. Following this data collection, the centralized value functions are learned based on the information of all agents, and then the gradient from the centralized value function is used to train the policy of each agent. But in the process of execution, each agent outputs action according to its individual observation (see Fig.~\ref{fig:CTDE}). 

It is important to acknowledge the inherent trade-offs of the CTDE. One significant strength is the enhanced learning stability enabled by centralized training, which can result in more robust policies \cite{9931995}. However, it is crucial to recognize that maintaining a centralized controller comes with its own set of challenges. Firstly, it can be prohibitively expensive and infeasible in certain scenarios, particularly in large-scale or resource-constrained environments. Additionally, the centralized controller introduces privacy concerns, as it necessitates the sharing of sensitive information among agents. Moreover, the central controller becomes a single point of failure, rendering the whole system vulnerable to disruptions. These trade-offs highlight the need for careful consideration of the CTDE framework's suitability in specific MARL applications, taking into account the advantages and drawbacks it presents.

\subsubsection{\textbf{Decentralized training with decentralized execution (DTDE)}}
\begin{figure}[!ht]
\centerline{\includegraphics[width=7cm]{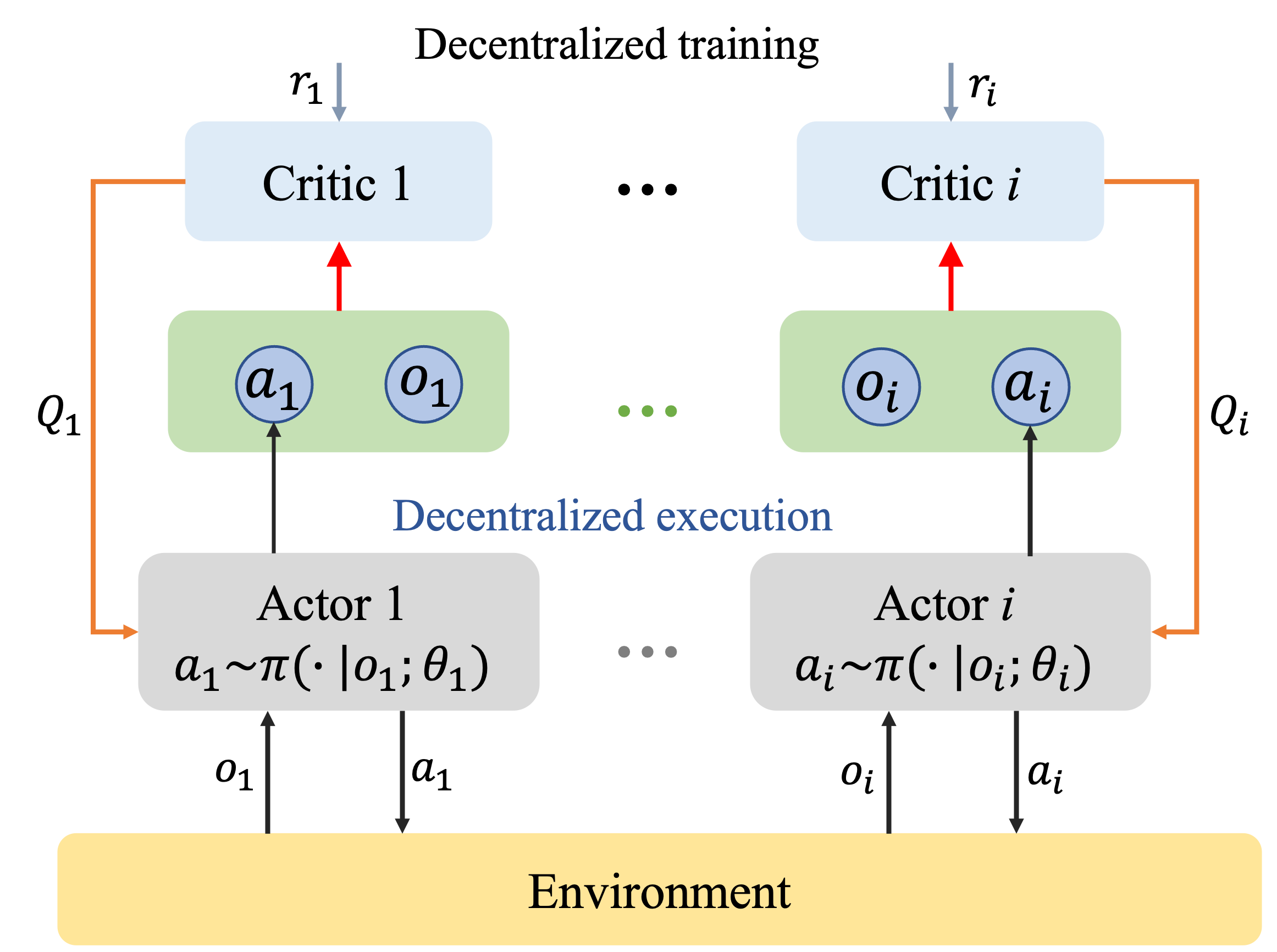}}
\caption{Illustration of decentralized training with decentralized execution (DTDE) in MARL.}
\label{fig:DTDE}
\vspace{-10pt}
\end{figure}

In decentralized settings, individual agents acquire knowledge independently, devoid of direct interactions. Each agent operates with its distinct set of observations, policies, and algorithms, utilizing the environment as the exclusive channel for engagement. Commonly employed algorithms within this framework encompass independent Q-learning (IQL) \cite{tan1993multi} and independent advantage actor-critic (IA2C) \cite{chu2019model}, frequently adopted as foundational benchmarks. Although these algorithms enable autonomous learning by agents, they may not fully harness the collaborative potential \cite{nguyen2020deep}. To address non-stationarity concerns, \cite{zhang2018fully} proposes a fully decentralized MARL framework. In this framework, each agent makes decisions based on both its local observation and the messages exchanged with neighboring agents via the network, offering convergence guarantees. Fig.~\ref{fig:DTDE} shows the DTDE pattern based on the actor-critic framework.

\subsection{MARL algorithm variants}
In this subsection, we have exclusively reviewed the common MARL algorithms that find utility in CAV applications. Further, the MARL algorithms are divided into four categories: value function decomposition, learning to communicate, hierarchical structure, and causal inference, which are rooted in the inherent complexity and diverse requirements of multi-agent environments, particularly in CAV control. 
Table~\ref{table13} offers an extensive overview of key works from four distinct perspectives within various settings.

\begin{table*}[!th]
\centering
\renewcommand{\arraystretch}{1.2}
\caption{An overview of the primary works for MARL algorithms.}

\begin{tabular}{>{\centering}m{0.1\textwidth} | >{\centering}m{0.08\textwidth}  >{\centering}m{0.08\textwidth}  m{0.4\textwidth}}
\hline

\shortstack{\textbf{Categories}} 
& \shortstack{\textbf{Work}}  & \shortstack[c]{\quad \\ \textbf{MARL} \\ \textbf{algorithm}} & \shortstack{\textbf{\qquad \qquad \qquad \qquad \qquad \qquad \qquad \qquad Novelty}} \\
\hline

\multirow{6}{*}{\shortstack{ Value Function \\  Decomposition}} 
                                 & \multicolumn{1}{c}{\shortstack{\hspace{1pt} \cite{SunehagLGCZJLSL18}, 2018}} & \multicolumn{1}{c}{\shortstack[c]{VDN}} & \multicolumn{1}{l}{\shortstack[l]{\quad \\ Decomposing value functions using linear summation}}\\\cline{2-4}
                                 & \multicolumn{1}{c}{\cite{pmlr-v80-rashid18a}, 2018} & \multicolumn{1}{c}{QMIX} & \multicolumn{1}{l}{Enforcing monotonicity constraints between joint and individual value functions}\\ \cline{2-4}    
                                 & \multicolumn{1}{c}{\cite{pmlr-v97-son19a}, 2019} & \multicolumn{1}{c}{QTRAN} & \multicolumn{1}{l}{ Relaxing the non-negative weight network in Qmix}\\\cline{2-4}
                                 & \multicolumn{1}{c}{\cite{RashidFPW20}, 2020} & \multicolumn{1}{c}{WQMIX} & \multicolumn{1}{l}{Introducing weighting functions and non-monotonic mixing networks}\\ \cline{2-4}       
                                 & \multicolumn{1}{c}{\cite{zhang2021avd}, 2021} & \multicolumn{1}{c}{AVD-Net} & \multicolumn{1}{l}{An attention-based approach to capitalize on the coordination relations between agents}\\\cline{2-4}
                                 & \multicolumn{1}{c}{\cite{10103926}, 2023} & \multicolumn{1}{c}{PER} & \multicolumn{1}{l}{Both the reward of an agent acting on its own and cooperating with other observable agents}\\\hline

\multirow{4}{*}{\shortstack{ Learning to \\ Communicate}} 
                                 & \multicolumn{1}{c}{\shortstack{\hspace{2pt}\cite{foerster2017stabilising}, 2017}} & \multicolumn{1}{c}{\shortstack[c]{FingerPrint}} & \multicolumn{1}{l}{\shortstack[l]{\quad \\Merging low-dimensional policy fingerprints in the state space of neighboring agents}}\\ \cline{2-4}
                                 & \multicolumn{1}{c}{\cite{foerster2016learning}, 2016} & \multicolumn{1}{c}{DIAL} & \multicolumn{1}{l}{The agent continuously encodes information and sends it to other recipients}\\ \cline{2-4}
                                 & \multicolumn{1}{c}{\cite{chu2020multiagent}, 2020} & \multicolumn{1}{c}{NeurComm} & \multicolumn{1}{l}{Encoding and concatenating communicating signals rather than aggregating them}\\ \cline{2-4}
                                 & \multicolumn{1}{c}{\cite{wang2020}, 2020} & \multicolumn{1}{c}{IMAC} & \multicolumn{1}{l}{Learning an efficient communication protocol based on the information bottleneck principle}\\\hline

\multirow{6}{*}{\shortstack{Hierarchical \\ Structure}} 
                                 & \multicolumn{1}{c}{\shortstack{\cite{ahilan2019feudal}, 2019}} & \multicolumn{1}{c}{\shortstack[c]{FHM}} & \multicolumn{1}{l}{\shortstack[l]{\quad \\Establishing a feudal hierarchy structure}}\\\cline{2-4}
                                 & \multicolumn{1}{c}{\cite{tang2018hierarchical}, 2018} & \multicolumn{1}{c}{HQMIX} & \multicolumn{1}{l}{Including hierarchical QMIX and hierarchical communication network}\\ \cline{2-4}
                                 & \multicolumn{1}{c}{\cite{yang2019hierarchical}, 2019} & \multicolumn{1}{c}{HSD} & \multicolumn{1}{l}{Training a cooperative decentralized policy for high-level skill selection}\\\cline{2-4}
                                 & \multicolumn{1}{c}{\cite{wang2020rode}, 2019} & \multicolumn{1}{c}{RODE} & \multicolumn{1}{l}{Introducing a method to clearly divide the action space through action clustering}\\\cline{2-4}
                                 & \multicolumn{1}{c}{\cite{xu2023haven}, 2023} & \multicolumn{1}{c}{HAVEN } & \multicolumn{1}{l}{Designing a dual coordination mechanism within a two-level hierarchical structure}\\\hline

\multirow{5}{*}{\shortstack{ Causal \\ Inference}} 
                                 & \multicolumn{1}{c}{\shortstack{\hspace{2pt}\cite{jaques2018intrinsic}, 2018}} & \multicolumn{1}{c}{\shortstack[c]{CIR}} & \multicolumn{1}{l}{\shortstack[l]{\quad \\Agents receive a reward based on their causal influence on the actions of others}}\\\cline{2-4}
                                 & \multicolumn{1}{c}{\cite{pina2023discovering}, 2023} & \multicolumn{1}{c}{ICL} & \multicolumn{1}{l}{Evaluating the causal effect of each agent's observations on team collaboration performance}\\\cline{2-4}
                                 & \multicolumn{1}{c}{\cite{liu2023lazy}, 2023} & \multicolumn{1}{c}{LAIES} & \multicolumn{1}{l}{Calculating the causal effect of their actions on external states using the do-calculus process}\\\cline{2-4}
                                 & \multicolumn{1}{c}{\cite{wang2022fully}, 2022} & \multicolumn{1}{c}{FD-MARL} & \multicolumn{1}{l}{Constructing continuous communication protocols based on causal analysis}\\\cline{2-4}
                                 & \multicolumn{1}{c}{\cite{li2022deconfounded}, 2022} & \multicolumn{1}{c}{DVD} & \multicolumn{1}{l}{Deconfounded value function decomposition based on causal effects}\\\hline

\end{tabular}
\label{table13}
\end{table*}

\subsubsection{\textbf{Value function decomposition}}

The challenge of credit assignment in cooperative settings has emerged as a significant area of research interest. The shared rewards in a fully cooperative environment make it difficult to distinguish the contribution of each agent, some agents tend to be lazy \cite{jiang2023credit}. To solve the above problem, some studies learn different value functions to distinguish the contribution of each agent. Foerster et al. \cite{counterfactual} and Guo et al. \cite{9185035} introduced the counterfactual baseline principle to learn different value functions by centralized learning method in the context of a shared team reward environment. Additionally, Hou et al. have developed a credit allocation mechanism based on role attention, which facilitates the learning process of role policies in multi-agent systems by managing how credit is allocated among agents \cite{10104101}. Liu et al. construct the causal effect of the agent's actions on the external state from the perspective of a causal diagram, solving the problem of lazy agents \cite{liu2023lazy}. 

Among the above methods to solve credit assignment, value function decomposition is considered to be an important research direction. The main concept and feature of value function decomposition is to decompose a joint value function into individual components to identify the distinct contributions of each agent. In the process, many useful value function methods were proposed to solve increasingly complex problems. 
VDN decomposes the joint action-value function by breaking it down into a sum of the action-value functions for individual agents. This linear decomposition method yields favorable results in certain simple problems \cite{SunehagLGCZJLSL18}.
Different from the linear value function decomposition method of VDN, 
QMIX enhances the representational capacity of the joint value function by enforcing monotonicity constraints between joint and the individual value functions. This approach allows for handling increasingly complex scenarios effectively \cite{pmlr-v80-rashid18a}.
QTRAN further improves the representation ability of joint functions by relaxing the non-negative weight network in QMIX \cite{pmlr-v97-son19a}. WQMIX extends the value function method to complex scenarios of non-monotonic rewards by introducing weighting functions and non-monotonic mixing networks \cite{RashidFPW20}. After that, some variant algorithms based on VDN and QMIX were used to solve different specific problems \cite{zhang2021avd, 10103926}.

\subsubsection{\textbf{Learning to Communicate}}
Research in MARL can be categorized into two distinct groups according to their methods of communication \cite{chu2020multiagent}. 
The first group operates without communication among agents and centers its efforts on stabilizing training by employing advanced value estimation methods. For example, MADDPG \cite{lowe2017multi} extends DDPG to the MARL setting by employing a centralized critic network to enable the global value estimate. Similarly, COMA \cite{foerster2018counterfactual} modifies the actor-critic method for use in MARL, calculating the advantage function for each agent by employing a centralized critic in conjunction with a counterfactual baseline.
The second group investigates heuristic communication protocols, which may involve either direct message sharing or learnable communication protocols. For instance, in FingerPrint \cite{foerster2017stabilising}, the low-dimensional policy fingerprints are directly shared and incorporated into the state space of neighboring agents. In DIAL \cite{foerster2016learning}, each DQN agent concurrently produces a communicated message while assessing action values. This message is encoded and subsequently combined with additional input signals at the receiving end.
NeurComm proposed in \cite{chu2020multiagent} argues that encoding and concatenating communicating signals, rather than aggregating them, offers distinct advantages in mitigating information loss during communication. IMAC method learns an efficient communication protocol as well as scheduling based on the information bottleneck principle \cite{wang2020}. Their method involves the development of a novel differentiable communication protocol, which incorporates information related to state, policy, and the encoded data from neighboring agents. Subsequently, this information is encoded and concatenated into the state space of the neighboring agents.

\subsubsection{\textbf{Hierarchical Structure}}
In the realm of multi-agent systems, a plethora of challenges are encountered, encompassing issues like sparse rewards, sequential decision-making, and limited transfer capabilities. To address these challenges effectively, the adoption of a hierarchical approach is unequivocally advantageous. The fundamental tenet of hierarchical multi-agent reinforcement learning revolves around the acquisition of hierarchical strategies for decision-making across varying levels of temporal abstraction. Current research in this domain broadly categorizes hierarchical multi-agent reinforcement learning into two paradigms: option-based hierarchical structures \cite{bacon2017option, harb2018waiting} and goal-based hierarchical structures \cite{vezhnevets2017feudal, nachum2018data}, both of which have found applications in diverse multi-agent scenarios.

Feudal Multi-agent Hierarchies (FMH) establish a feudal hierarchy within a multi-agent context, yielding noteworthy results. However, a primary limitation of this approach is its inherent challenge in tackling fully collaborative tasks with shared rewards \cite{ahilan2019feudal}. To address the predicament of multi-agent scenarios characterized by sparse and delayed rewards, several hierarchical structures with temporal abstraction have been proposed, including hierarchical QMIX and hierarchical communication network methods \cite{tang2018hierarchical}.

Furthermore, recognizing the limitations of traditional approaches, which typically rely on  manually defined high-level action spaces, the Hierarchical Learning with Skill Discovery (HSD) model was conceived. This model aims to autonomously train decentralized policies for high-level skill selection while simultaneously developing independent low-level policies for the execution of these skills, thereby enhancing adaptability to changing scenarios \cite{yang2019hierarchical}. On the other hand, learning roles to decompose (RODE) is introduced for clear action space partitioning through action clustering, where each action corresponds to a distinct subspace, this innovation substantially mitigates the challenge of extensive parameter tuning associated in HSD \cite{wang2020rode}. Moreover, HierArchical Value dEcompositioN (HAVEN) introduces a dual coordination mechanism, encompassing inter-layer and inter-agent strategies, facilitated by the formulation of a reward function within a two-level hierarchical structure. This approach effectively addresses the instability issue through the concurrent optimization of strategies at all levels and inter-agent coordination \cite{xu2023haven}. Hierarchical MARL is recognized as a promising avenue for addressing collaborative decision-making challenges in complex, large-scale scenarios. Nonetheless, it faces the challenge of difficulty in designing complex dynamic hierarchical structures and difficult migration of policies.

\subsubsection{\textbf{Causal Inference}}
There are multiple complex variables in MARL, and it is usually difficult to directly discover their internal relationships. Therefore, some recent research has begun to combine MARL with causal inference to discover the causal relationships between agents or variables, and further understand the operating mechanism of agents through intervention and inference, thereby motivating the agent to conduct more targeted learning \cite{grimbly2021causal}. Pearl's three-level causal model is considered a powerful tool for constructing causal relationships between variables \cite{pearl2018theoretical}. Causal influence reward methods promote the collaborative performance of agents by rewarding those agents that have a causal influence on the actions of other agents, where this causal influence is evaluated by counterfactual reasoning \cite{jaques2018intrinsic}. Rafael Pina et al proposed the independent causal learning (ICL) algorithm to evaluate the causal effect of each agent's observations on team collaboration performance and solve the credit allocation problem in the independent learning framework \cite{pina2023discovering}.

There are also other works that combine causal inference from different perspectives under the framework of MARL. LAIES mathematically define the concept of fully lazy agents and teams by calculating the causal effect of their actions on external states using the do-calculus process, which solved the sparse reward problem in MARL \cite{liu2023lazy}. FD-MARL allows agents to modify communication messages by choosing the counterfactual that bears the most significant influence on others, the continuous communication based on causal analysis enables efficient information transformation in a fully decentralized manner \cite{wang2022fully}. The deconfounded value decomposition (DVD) method investigates value function decomposition from the perspective of causal inference, which cuts off the backdoor confounding path from the global state to the joint value function \cite{li2022deconfounded}. In addition, causal inference has also been applied to some practical application scenarios, such as traffic signal control \cite{yang2023causal} and human-computer interaction decision-making \cite{ho2021human}.

\begin{table*}[!th] \label{table:23}
\centering
\caption{An overview of the primary works from 1-D and 2-D cooperation for MARL applied in CAV control.}

\begin{tabularx}{\textwidth}{c | c | c c c X}
\hline
\shortstack{\textbf{Scenarios} \\ \quad \\ \quad} & \shortstack{\quad \\ \textbf{Typical} \\ \shortstack{\textbf{application}}} & \shortstack{\textbf{Work} \\ \quad \\ \quad}  & \shortstack{\quad \\ \textbf{MARL} \\ \textbf{algorithm}} & \shortstack{\textbf{Novelty} \\ \quad} & \shortstack{\textbf{\qquad \quad Metrics} \\ \quad \\ \quad} \\
\hline

\multirow{16}{*}{\shortstack{ 1-D \\ Cooperation}} & \multirow{16}{*}{\shortstack{Platooning \\ control}} 

& \multicolumn{1}{c}{\shortstack{\cite{peake2020multi}, 2020 \\ \quad }} & \multicolumn{1}{c}{\shortstack[c]{LSTM-Based \\ MA-Reinforce}} & \multicolumn{1}{l}{\shortstack[l]{\quad \\\scalebox{0.8}{$\bullet$} Individual and global rewards design \\ \scalebox{0.8}{$\bullet$} A LSTM-based communication protocol }} & \multicolumn{1}{l}{\shortstack[l]{\scalebox{0.8}{$\bullet$} Convergence stability \\\scalebox{0.8}{$\bullet$} Travel time }} \\\cline{3-6}

&       & \multicolumn{1}{c}{\shortstack{\cite{vu2020multi}, 2020 \\ \quad \\ \quad \\ \quad }} & \multicolumn{1}{c}{\shortstack[c]{MA-DDQN \\ \quad \\ \quad \\ \quad}} & \multicolumn{1}{l}{\shortstack[l]{\quad \\ \scalebox{0.8}{$\bullet$} Local channel data incorporation \\ \scalebox{0.8}{$\bullet$} A different reward design including the weighted \\sum-rates and the transmission time }} & \multicolumn{1}{l}{\shortstack[l]{\scalebox{0.8}{$\bullet$} Sum-rate of V2N \\ \scalebox{0.8}{$\bullet$} Packet probability of V2V  \\ \quad}}\\\cline{3-6}

&       & \multicolumn{1}{c}{\shortstack{\cite{chu2020multiagent}, 2020 \\ \quad \\ \quad \\ \quad}} & \multicolumn{1}{c}{\shortstack[c]{NeurComm \\ \quad \\ \quad \\ \quad}} & \multicolumn{1}{l}{\shortstack[l]{\quad\\\scalebox{0.8}{$\bullet$} A spatial discount factor to stabilize training \\ \scalebox{0.8}{$\bullet$} A new differentiable communication protocol on \\ both agent states and behaviors}} & \multicolumn{1}{l}{\shortstack[l]{\scalebox{0.8}{$\bullet$} Average  headway \\\scalebox{0.8}{$\bullet$} Average  velocity \\ \quad }} \\\cline{3-6}

&       & \multicolumn{1}{c}{\shortstack{\cite{liu2021efficient}, 2021 \\ \quad \\ \quad \\ \quad }} & \multicolumn{1}{c}{\shortstack[c]{C-HARL \\ \quad \\ \quad \\ \quad}} & \multicolumn{1}{l}{\shortstack[l]{\quad \\ \scalebox{0.8}{$\bullet$}  Cooperative hierarchical attention RL (C-HARL)\\
\scalebox{0.8}{$\bullet$} Meta-learning with GNN integration \\ \quad}} & \multicolumn{1}{l}{\shortstack[l]{\scalebox{0.8}{$\bullet$} Convergence \\
\scalebox{0.8}{$\bullet$} Success rate of message \\
\scalebox{0.8}{$\bullet$} Communication delay}} \\\cline{3-6}

&       & \multicolumn{1}{c}{\shortstack{\cite{li2022deep}, 2022 \\ \quad \\ \quad \\ \quad  }} & \multicolumn{1}{c}{\shortstack{MADDPG \\ \quad \\ \quad \\ \quad  }} & \multicolumn{1}{l}{\shortstack[l]{\quad \\ \scalebox{0.8}{$\bullet$} A Stackelberg game to model interactions \\
\scalebox{0.8}{$\bullet$} Decisions sharing with social partners  within the\\ platoon due to social effects}} & \multicolumn{1}{l}{\shortstack[l]{\scalebox{0.8}{$\bullet$} Convergence \\
\scalebox{0.8}{$\bullet$} Defined utility value \\
\scalebox{0.8}{$\bullet$} Social effect }} \\\cline{3-6}

&       & \multicolumn{1}{c}{\shortstack{\cite{shi2023deep}, 2023 \\ \quad \\ \quad \\ \quad }} & \multicolumn{1}{c}{\shortstack[c]{Distributed \\ PPO\\ \quad \\ \quad}} & \multicolumn{1}{l}{\shortstack[l]{\quad \\ \scalebox{0.8}{$\bullet$} State fusion strategy with the equilibrium to \\ stabilize traffic oscillations \\
\scalebox{0.8}{$\bullet$} A reward function design in quadratic form}} & \multicolumn{1}{l}{\shortstack[l]{\scalebox{0.8}{$\bullet$} Driving comfort \\
\scalebox{0.8}{$\bullet$} Stability \\ 
\scalebox{0.8}{$\bullet$} Travel efficiency }} \\\hline

\multirow{19}{*}{\shortstack{2-D \\ Cooperation}} & \multirow{19}{*}{\shortstack{Cooperative \\ lane \\ changing}} 

& \multicolumn{1}{c}{\shortstack{\cite{hou2021decentralized}, 2021 \\ \quad \\ \quad \\ \quad }} & \multicolumn{1}{c}{\shortstack[c]{MAPPO \\ \quad \\ \quad \\ \quad }} & \multicolumn{1}{l}{\shortstack[l]{\quad \\ \scalebox{0.8}{$\bullet$} A decentralized cooperative lane-changing design \\
\scalebox{0.8}{$\bullet$} A novel reward function design considering \\ eliminating traffic shock waves}} & \multicolumn{1}{l}{\shortstack[l]{\scalebox{0.8}{$\bullet$} Traffic throughput\\
\scalebox{0.8}{$\bullet$} Number of Stops \\ 
\scalebox{0.8}{$\bullet$} Fuel efficiency}} \\\cline{3-6}

&       & \multicolumn{1}{c}{\shortstack{\cite{wang2021harmonious}, 2021 \\ \quad \\ \quad \\ \quad \\ \quad \\ \quad}} & \multicolumn{1}{c}{\shortstack[c]{Zero-sum \\ DQN  \\ \quad \\ \quad \\ \quad   }} & \multicolumn{1}{l}{\shortstack[l]{\quad \\ \scalebox{0.8}{$\bullet$} The delay of an individual vehicle and overall \\ traffic efficiency in the reward function\\
\scalebox{0.8}{$\bullet$} A lane change model with limited sensing \\  without V2X communications}} & \multicolumn{1}{l}{\shortstack[l]{\scalebox{0.8}{$\bullet$} Convergence \\
\scalebox{0.8}{$\bullet$} Stability\\
\scalebox{0.8}{$\bullet$} Travel efficiency \\ \quad }} \\\cline{3-6}

&       & \multicolumn{1}{c}{\shortstack{\cite{nagarajan2021lane}, 2021 \\ \quad \\ \quad \\ \quad \\ \quad \\ \quad}} & \multicolumn{1}{c}{\shortstack{ MADQN \\ \quad \\ \quad \\ \quad \\ \quad \\ \quad}} & \multicolumn{1}{l}{\shortstack[l]{\quad \\ \scalebox{0.8}{$\bullet$} Interactions and negotiations model establishment \\between multiple agents  \\
\scalebox{0.8}{$\bullet$} An experience replay mechanism to tackle the \\non-stationarity}} & \multicolumn{1}{l}{\shortstack[l]{\scalebox{0.8}{$\bullet$} Timeout\\
\scalebox{0.8}{$\bullet$} Accident\\
\scalebox{0.8}{$\bullet$} Success rate \\ \quad \\ \quad}} \\\cline{3-6}

&       & \multicolumn{1}{c}{\shortstack{\cite{chen2022multi}, 2022 \\ \quad \\ \quad \\ \quad \\ \quad \\ \quad}} & \multicolumn{1}{c}{\shortstack{QMIX \\ \quad \\ \quad \\ \quad \\ \quad \\ \quad}} & \multicolumn{1}{l}{\shortstack[l]{\quad \\ \scalebox{0.8}{$\bullet$} The simplification of system complexity through \\reinforcement learning \\
\scalebox{0.8}{$\bullet$} The fairness in cooperation to allow for independent \\ lane-changing and overtaking}} & \multicolumn{1}{l}{\shortstack[l]{\scalebox{0.8}{$\bullet$} Collaborative effect \\
\scalebox{0.8}{$\bullet$} Travel efficiency \\ \quad \\ \quad \\ \quad }}  \\\cline{3-6}

&       & \multicolumn{1}{c}{\shortstack{\cite{zhang2022multi}, 2022 \\ \quad \\ \quad \\ \quad \\ \quad \\ \quad}} & \multicolumn{1}{c}{\shortstack{Bi-level \\ DQN \\ \quad \\ \quad \\ \quad \\ \quad }} & \multicolumn{1}{l}{\shortstack[l]{\quad \\ \scalebox{0.8}{$\bullet$} An innovative reward function designed with their \\ own benefits and the impact on overall traffic\\
\scalebox{0.8}{$\bullet$} The driving intentions of the surrounding vehicles \\encoded into the observation space}} & \multicolumn{1}{l}{\shortstack[l]{\scalebox{0.8}{$\bullet$} Comfort\\
\scalebox{0.8}{$\bullet$} Safety\\
\scalebox{0.8}{$\bullet$} Travel efficiency \\ \quad }} \\\cline{3-6}

&       & \multicolumn{1}{c}{\shortstack{\cite{zhou2022multi}, 2022 \\ \quad \\ \quad \\ \quad  }} & \multicolumn{1}{c}{\shortstack[c]{MA2C \\ \quad \\ \quad \\ \quad}} & \multicolumn{1}{l}{\shortstack[l]{\scalebox{0.8}{$\bullet$} A innovative local reward and parameter-sharing\\
\scalebox{0.8}{$\bullet$} Multi-objective reward function \\ \quad}} & \multicolumn{1}{l}{\shortstack[l]{\quad \\\scalebox{0.8}{$\bullet$} Driving comfort\\
\scalebox{0.8}{$\bullet$} Safety\\
\scalebox{0.8}{$\bullet$} Travel efficiency}} \\\hline

\end{tabularx}
\end{table*}

\section{Towards applications of MARL in CAVs} \label{sec:app}
CAVs have garnered substantial interest for their potential to reshape the transportation landscape, ushering in an era marked by heightened efficiency, enhanced safety, and bolstered sustainability. Operating within networked environments, CAVs engage in interactions with multiple vehicles, promoting cooperative driving. For example, CAVs can acquire the capabilities to collaborate, execute seamless merges, and navigate intersections with precision, thereby advancing traffic safety and optimizing flow \cite{wang2023faster}. These transformative capabilities are increasingly being explored in tandem with MARL to further optimize CAV interactions within networked environments. Within the realm of MARL, CAVs are designed to engage in intelligent interactions with multiple vehicles, thus promoting cooperative driving strategies.

In this section, we will comprehensively explore the recent strides made in the utilization of MARL within CAV applications. Our examination will be structured according to various dimensions of cooperation, each correlated with a specific number of control components. 
\begin{itemize}
    \item \textbf{One-dimensional cooperation} corresponds to scenarios involving control along a single control direction, such as either longitudinal control or lateral control.
    \item \textbf{Two-dimensional cooperation} extends the scope to include both longitudinal and lateral control components, reflecting the increased complexity in the coordination and decision-making of CAVs.
    \item \textbf{Three-dimensional cooperation} further augments the challenges and opportunities by incorporating additional constraints, such as time limits, which encompass aspects like traffic light control and on-ramp merging.
\end{itemize}
By categorizing these advancements based on control components, we aim to provide a comprehensive perspective on the evolving landscape of MARL within CAV applications, with a focus on the intricacies of cooperation and control in varying dimensions. Additionally, we provide an extensive examination of the leading simulation platforms employed in MARL research for CAVs.

\subsection{One-dimensional cooperation}
In this subsection, we will conduct a detailed review of one-dimensional cooperation applications, which primarily focus on longitudinal control. One prominent example of such application is \textit{platooning control}, where CAVs operate in closely-knit formations and collaborate to enhance efficiency, safety, and fuel economy. This is achieved by minimizing aerodynamic drag and optimizing traffic flow through synchronized control strategies and inter-vehicle coordination. Platooning control stands as a noteworthy illustration of how MARL contributes to the advancement of transportation systems. 

\begin{figure}[!ht]
\centerline{\includegraphics[width=0.90\columnwidth]{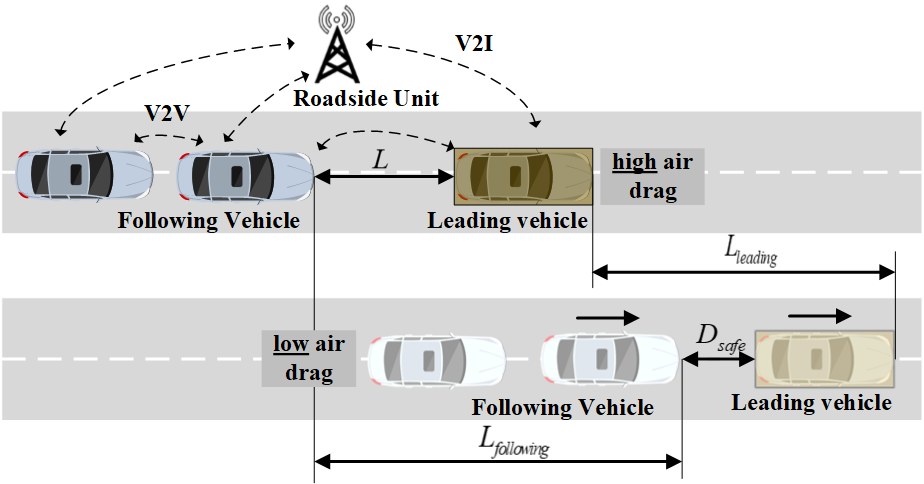}}
\caption{One-dimensional cooperation scenarios: the case of platooning control.}
\label{fig:platoon}
\vspace{-10pt}
\end{figure}

As shown in Fig. \ref{fig:platoon}, the problem of platooning control is commonly addressed through a model-free, multi-agent network approach \cite{chu2020multiagent}.  In this framework, each agent, symbolizing an AV, has the ability to communicate with both preceding and following vehicles via vehicle-to-vehicle (V2V) communication channels.
This setup aims to ensure safe, fuel-efficient, and smooth vehicle following operations, while also maximizing the advantages of driving in close formation \cite{tsugawa2016review}. 
 
Given $N$ vehicles in a platoon, the control objective is to maintain a desired spacing $L_i$ and ensure velocity matching. The state vector \cite{liu2020platoon} for each vehicle $i$ at time $t$ is:

\begin{equation}
\dot{\mathbf{x}}_{i}(t) = \left[\begin{array}{ll}
0 & 1 \\
0 & 0
\end{array}\right] \mathbf{x}_{i}(t) + \left[\begin{array}{l}
0 \\
1
\end{array}\right] \left( k_p \left( \left( x_{i-1}(t) - x_i(t) \right) - L_i \right) \right.
\end{equation}
\begin{equation*}
\left. + k_d \left( v_{i-1}(t) - v_i(t) \right) \right)
\end{equation*}
where $x_i(t)$ is the position of the $ith$ vehicle at time  $t$, $v_i(t)$ is the velocity of the $ith$ vehicle at time $t$,
 $L_i$ is the desired inter-vehicle distance (spacing) between vehicle $i$ and $i+1$, $\mathbf{x}_i(t)=\left[x_i(t), v_i(t)\right]^{T}$ is the state vector, $k_p$ and $k_d$ are the proportional gain for position error and the derivative gain for velocity error respectively. To optimize the platoon control, an objective function  $J$  can be defined, incorporating both spacing error and velocity error over a time horizon $T$:

\begin{equation}
J=\sum_{i=1}^N \int_0^T\left(\alpha e_{x, i}^2(t)+\beta e_{v, i}^2(t)\right) d t
\end{equation}
 where $\alpha$ and $\beta$ are weighting factors that balance the importance of position and velocity errors.
Thus, cooperative strategies would be developed to achieve the collective goal, adapting their behaviors in response to the actions of other vehicles within the network.

One critical task of platooning control is to achieve \textit{string stability}, which refers to the ability of a line of CAVs to maintain a stable and orderly formation (e.g., stable distance and speed) as they travel in close proximity to each other \cite{feng2019string}. Notably, MARL has been proposed to address and enhance the achievement of string stability in the platooning control scenario, contributing to the development of more efficient and coordinated platooning control strategies. For instance, in \cite{peake2020multi}, a MARL approach based on a robust communication protocol with Long Short-Term Memory (LSTM) \cite{hochreiter1997long} is introduced, resulting in stable platooning control. In addition, Li et al. introduce the Communication Proximal Policy Optimization (CommPPO) algorithm to enhance stability, which adapts to varying agent numbers and supports different dynamics of platooning control. CommPPO incorporates a predecessor-leader-follower communication protocol to facilitate the transmission of both global and local state information among agents. Notably, CommPPO introduces a novel reward communication channel, effectively mitigating issues related to spurious rewards and mitigating the problem of ``lazy agents'', which are commonly encountered in other MARL approaches \cite{li2021reinforcement}.
In \cite{chu2020multiagent}, the platooning control problem is formulated as a spatiotemporal MDP. They enhance system stability by introducing a spatial discount factor for local agents. Furthermore, they introduce a novel differentiable communication protocol, NeurComm, which mitigates non-stationarity, leading to improved learning efficiency and control performance. 

\textit{Efficient communication} among CAVs in platooning control also poses a fundamental challenge in achieving seamless coordination and maintaining the desired following properties, such as precise inter-vehicle spacing and synchronized maneuvers. For instance, Liu et al. introduce a Multi-Agent Hierarchical Attention RL (MAHARL) framework, which incorporates the Graph Attention Network (GAT) at each hierarchical level to account for the reciprocal influences between agents. The hierarchical architecture of MAHARL empowers agents with foresight, enabling them to make favorable decisions, even in the absence of immediate incentives, and consider long-term rewards, as elaborated in their work \cite{liu2021efficient}. Additionally, in scenarios where complete information is lacking, Li et al. \cite{li2022deep} develop a game-theoretic framework to capture the strategic interactions between the leading vehicle and the following vehicles. To overcome the absence of full information, the equilibrium solution in platooning control is identified through backward induction, coupled with information collection from the following vehicles.

\textit{Mixed platooning control} involves the simultaneous presence of multiple vehicle platoons within a network, necessitating a sophisticated level of coordination and communication both within (intra-platoon) and between (inter-platooning) these platoons.  \cite{xu2023deep}. 
Parvini et al. have introduced two advanced MARL algorithms specifically designed to address the challenges of mixed platooning control. These include the modified MADDPG and its variant, the modified MADDPG with task decomposition. In these frameworks, the leaders of the platoons are treated as independent agents that interact with the environment to determine the most effective policies for platoon management. Notably, these algorithms employ multiple critics to estimate both global and local rewards, thereby fostering cooperative behavior among the agents. In the second algorithm, individual rewards are further decomposed into task-specific sub-reward functions, offering insights into their work as detailed in \cite{parvini2023aoi}.

Even though CAVs have made significant advancements, there will be an extended period during which CAVs will coexist with human-driven vehicles (HDVs), a situation commonly referred to as \textit{mixed-traffic} scenarios \cite{chu2019model}. Different penetration rates can be assessed to understand the impact of mixed traffic. In the study presented in \cite{wang2019cooperative}, the impact of the penetration rate of CAVs on the energy efficiency of the traffic network is examined, with a specific emphasis on a cooperative eco-driving system that utilizes longitudinal control. A noteworthy aspect of this research is the introduction of a role transition protocol for CAVs, enabling them to smoothly transition between leading and following positions within a vehicle string. Lu et al. enhance the evaluation by integrating several key elements, including altruism control, a quantitative car-following strategy, a refined platoon reward function, and a collision avoidance method, as outlined in their work \cite{lu2023altruistic}.
In \cite{shi2023deep}, the authors introduce the concept of characterizing consecutive HDVs as a collective entity, referred to as AHDV, to minimize stochastic variability and leverage macroscopic characteristics for the control of following CAVs. They employ a control strategy built on distributed proximal policy optimization (DPPO) to anticipate disturbances and downstream traffic conditions in mixed traffic scenarios.
In vehicular environments characterized by high mobility, the use of traditional centralized optimization methods that rely on global channel information can be impractical. In \cite{vu2020multi}, the authors model each CAV as an individual agent, which makes decisions based on local information and communication with neighboring vehicles, without relying on a centralized controller.

\subsection{Two-dimensional cooperation}
\begin{figure}[!ht]
\centerline{\includegraphics[width=0.9\columnwidth]{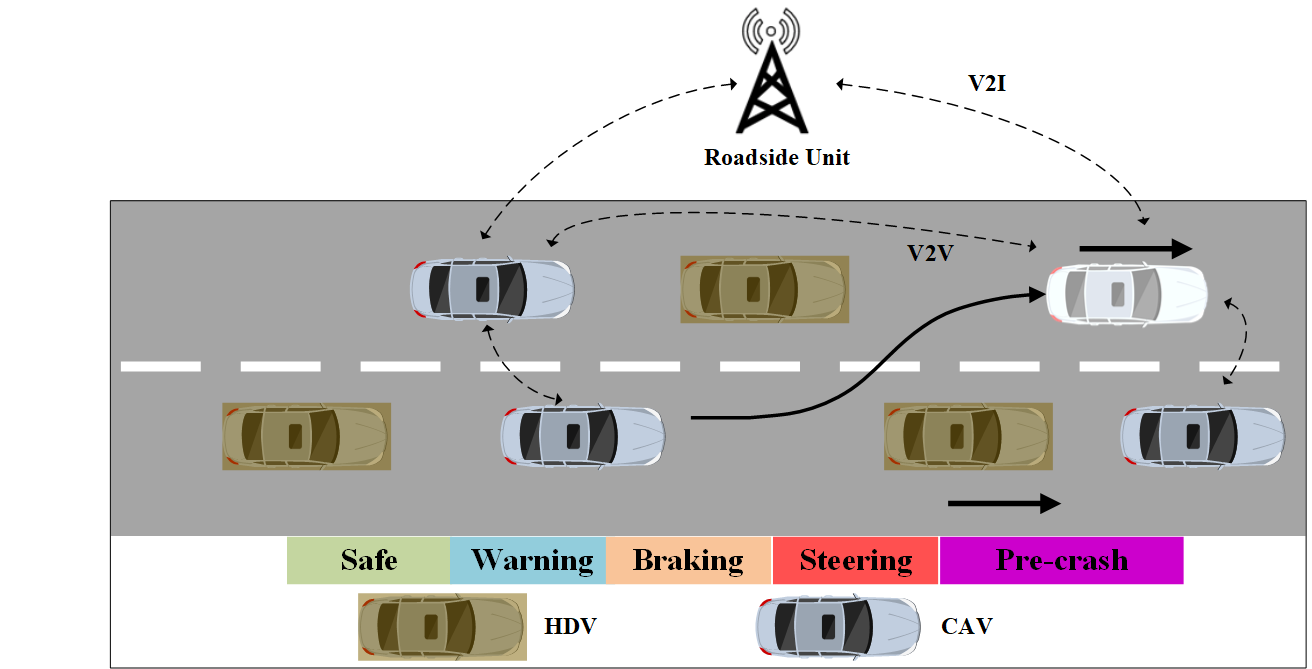}}
\caption{Two-dimensional cooperation scenario.}
\label{fig:two_dim}
\vspace{-10pt}
\end{figure}

Compared to one-dimensional cooperation, two-dimensional cooperation addresses problems related to both longitudinal and lateral control, with its most typical application being \textit{cooperative lane changing}. As shown in Fig.~\ref{fig:two_dim}, CAVs communicate and coordinate through V2V and V2I communication channels, enabling them to make decisions and acquire effective lane-changing strategies within shared driving environments \cite{he2022robust}. 

Generally, the lane-changing mathematical problem \cite{chen2024lane} can be expressed as follows:

\begin{equation}
\begin{aligned}
\min_{\mathbf{u}_i(t)} J &= \int_{t_0}^{t_f} \Bigg[ \alpha_1 \Big( \left( x_{i-1}(t) - x_i(t) - L_i - \Delta x_{\text{safe}} \right)^2 + \\
&\quad \left( x_i(t) - x_{i+1}(t) - L_{i+1} - \Delta x_{\text{safe}} \right)^2 \Big)  + \alpha_2 \left( t_f - t_0 \right) \\
&\quad + \alpha_3 \Big( u_{x,i}(t)^2 + u_{y,i}(t)^2 + \left( \frac{d}{dt} u_{x,i}(t) \right)^2 + \\
&\left( \frac{d}{dt} u_{y,i}(t) \right)^2 \Big) \Bigg] dt
\end{aligned}
\label{two_co_1}
\end{equation}

Subject to the vehicle dynamics:
\begin{equation}
\dot{\mathbf{x}}_i(t)=\mathbf{A} \mathbf{x}_i(t)+\mathbf{B} \mathbf{u}_i(t)
\label{two_co_2}
\end{equation}
where $\alpha_1$, $\alpha_2$, and $\alpha_3$ are the corresponding weights for safety, efficiency, and comfort, respectively. Other objectives may also be considered;
$\mathbf{A}=\left[\begin{array}{llll}0 & 0 & 1 & 0 \\ 0 & 0 & 0 & 1 \\ 0 & 0 & 0 & 0 \\ 0 & 0 & 0 & 0\end{array}\right], \quad \mathbf{B}=\left[\begin{array}{ll}0 & 0 \\ 0 & 0 \\ 1 & 0 \\ 0 & 1\end{array}\right]$,  $\mathbf{x}_{i}(t)=\left[x_i(t),y_i(t), v_{x, i}(t), v_{y, i}(t)\right]^{T}$ is the state vector for vehicle  $i$; $x_i(t)$ and $y_i(t)$ are the longitudinal and lateral position of the $ith$ vehicle at time $t$; $v_{x, i}(t)$ and $v_{y, i}(t)$ are longitudinal and lateral velocity of the  $ith$ vehicle at time $t$.
$\mathbf{u}_{i}(t)=\left[u_{x, i}(t), u_{y, i}(t)\right]^{T}$ is the control input vector for vehicle  $i$, $u_{x, i}(t)$ and $u_{y, i}(t)$ are the control input for longitudinal and lateral acceleration respectively; $L_i$ is the length of the  $ith $ vehicle; $\Delta x_{\text{safe}}$ is the safety distance to avoid collision; $w_{\text{lane}}$ is the width of the lane.

And the constraints are expressed as:
\begin{equation}
\begin{aligned}
& x_{i-1}(t) - x_i(t) \geq L_i + \Delta x_{\text{safe}} \\
& x_i(t) - x_{i+1}(t) \geq L_{i+1} + \Delta x_{\text{safe}} \\
& y_i(t) \in \{\text{current lane, target lane}\} \\
& 0 \leq y_i(t) \leq w_{\text{lane}}
\end{aligned}
\label{two_co_3}
\end{equation}



To achieve a safe, efficient, and comfortable lane-change maneuver,
Hou et al. devise a decentralized cooperative lane-changing controller employing a Multi-Agent Proximal Policy Optimization (MAPPO) approach, which empowers each vehicle to independently learn and assess its policy and action rewards based on local information, while still having access to global state data. The trained policies can be effectively transferred and applied across various traffic conditions, enhancing traffic throughput from uncongested to highly congested scenarios \cite{hou2021decentralized}. 
In \cite{chen2022multi}, the authors put forth a twin-vehicle cooperative driving approach, leveraging the QMIX algorithm. Their method applies RL to dynamically adapt to changing conditions on highways, aiming to find an optimal balance between autonomous decision-making and cooperative interaction among the vehicles. This algorithm allows each vehicle in the pair to independently perform lane changes and overtaking maneuvers, even in dense traffic scenarios, while maintaining a predetermined formation between them. 
An enhanced QMIX algorithm \cite{chen2023multi} is further proposed to enhance the flexibility and effectiveness of the collaborative lane-changing system. 
They implement a stable estimation method to mitigate  the problem of overestimated joint Q-values among agents. This approach strikes a fine balance between maintaining formation and allowing for smooth overtaking, thus facilitating intelligent adaptation to a variety of scenarios, including heavy traffic, light traffic, and emergencies.
%
In an effort to enhance mobility within complex traffic environments, in \cite{vishnu2023improving}, instead of relying on relative distance and semantic maps, the authors suggest the utilization of traffic states that encapsulate the spatio-temporal interactions between neighboring vehicles. Within the MADRL framework, three prediction models, namely the transformer-based (TS-Transformer), generative adversarial network-based (TS-GAN), and conditional variational autoencoder-based (TS-CVAE) models, are developed and compared for traffic state prediction.

\textit{Graph neural networks (GNNs)} \cite{zhou2020graph} have gained significant attention in MARL settings, particularly for cooperative lane-changing applications. Chen et al. introduce a novel algorithm based on MADQN, which combines a GCN with a deep Q-network \cite{chen2021graph}. This approach facilitates effective information fusion and decision-making within the MARL framework, thereby enhancing both safety and mobility in cooperative lane-changing scenarios.
Moreover, in \cite{ha2020leveraging}, the authors leverage GNN in conjunction with MADDPG in their work for multi-agent training, addressing the complexity of real-world inputs and aiding in congestion mitigation efforts.

\textit{Safety} is of paramount importance in the context of CAV scenarios. In \cite{han2020multi}, the authors have introduced a safety-enhancing actor-critic algorithm that incorporates two innovative techniques. The first technique is the `truncated Q-function', which effectively leverages shared information from neighboring CAVs, ensuring scalability for large-scale CAV systems. The second technique, ``safe action mapping'', provides safety guarantees throughout both the training and execution phases by utilizing control barrier functions.
Additionally, a bi-level strategy is proposed in \cite{zhang2022multi} to further enhance safety and efficiency. At the upper level, the MADQN model is utilized for making decisions about lane changes. This approach acknowledges the cooperative aspect of driving by factoring in the intentions of surrounding vehicles, enabling implicit negotiation for right-of-way.
At the lower level, a right-of-way assignment model is utilized to ensure safety. A novel reward function is further proposed to encourage coordination and account for traffic impact. 
Moreover, Li et al. have introduced MetaDrive, a platform that focuses on safe driving by addressing generalizability, safety awareness, and multi-agent decision-making. It offers diverse driving scenarios for benchmarking single and multi-agent RL tasks \cite{9829243}. 


Thoughtful \textit{reward function design} is paramount for the effectiveness and efficiency of lane-changing strategies. In \cite{zhou2022multi}, the authors introduce the multi-agent advantage actor-critic (MA2C) framework, which integrates a novel local reward design and parameter-sharing scheme for cooperative lane changing in mixed traffic scenarios. They also present a multi-objective reward function that considers factors such as fuel efficiency, driving comfort, and safety to facilitate successful and efficient lane changing.
In \cite{wang2021harmonious}, a MARL approach is explored in which agents collaborate to reach a zero-sum game state. In contrast to cooperative driving, this approach, known as harmonious driving, places emphasis on achieving a balance between overall and individual efficiency while utilizing limited sensing data from individual vehicles. They design a reward function that promotes harmony by considering both individual and overall efficiency, as opposed to competitive strategies that solely prioritize individual interests.

\subsection{Three-dimensional cooperation}
\begin{figure*}[!ht]
\centerline{\includegraphics[width=1.0 \textwidth]{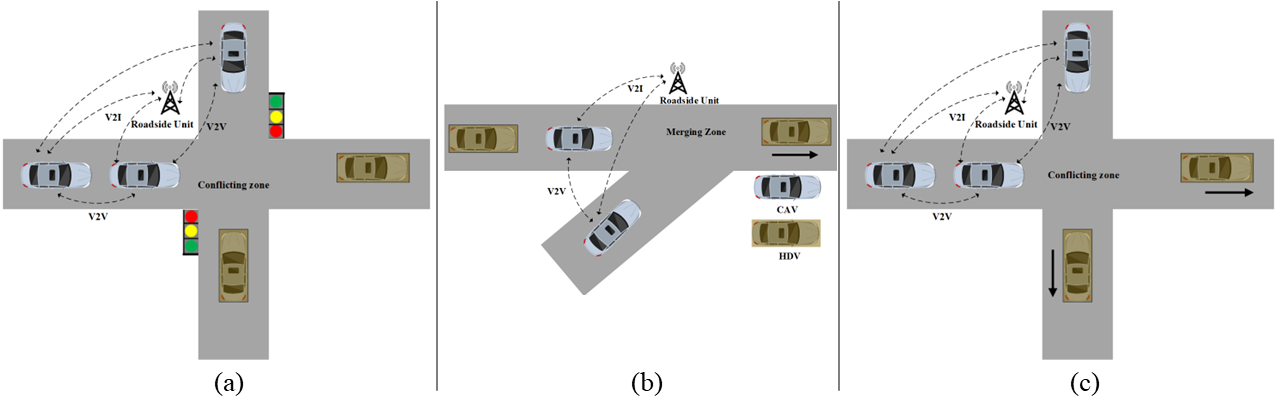}}
\caption{Three-dimensional cooperation scenarios: (a) Traffic Signal Coordination; (b) Merging on-ramps;(c) Unsignalized intersections.}
\label{fig:three_dim}
\vspace{-10pt}
\end{figure*}

Three-dimensional cooperation involves the intricate coordination of longitudinal and lateral movements along with timing, across a variety of traffic scenarios such as traffic signal coordination, on-ramp merging, and navigating through unsignalized intersections, as depicted in Fig.~\ref{fig:three_dim}. To ensure safe, efficient, and effective cooperation, CAVs need to rapidly learn longitudinal and lateral maneuvers within tight time constraints.
The three-dimensional cooperation problem can also be described by Eq. \ref{two_co_1} - Eq. \ref{two_co_3}, with the added time constraints $t_{i, \text { start }} \leq t \leq t_{i, \text { end }}$.


\subsubsection{\textbf{Traffic signal control (TSC) }}
TSC aims to alleviate congestion in saturated road networks by adapting signal timings based on real-time traffic conditions (see Fig.~\ref{fig:three_dim} (a)).
Traditional TSC methods address congestion by solving optimization problems to determine effective coordination and control policies. However, this process can be time-consuming and often requires precise system dynamics \cite{chu2019model}. Recently, model-free MARL approaches have shown a growing interest in precise TSC.


\textit{Value-based MARL} algorithms have been extensively investigated for TSC challenges. For instance, to alleviate the curse of dimensionality and environmental nonstationarity problems, a decentralized coordination MADQN approach is proposed in \cite{wu2019dcl} to explicitly identify and dynamically adapt agent coordination needs during the learning process. Similarly, in \cite{tan2019cooperative}, the complex TSC problem is decomposed into simpler subproblems and tackled using multiple regional agents and a centralized global agent. Each regional agent learns its RL policy for smaller regions with reduced action spaces, while the centralized global agent combines the RL contributions from various regional agents to form a final Q-function for the entire large-scale traffic grid.
Notably, QMIX, an extension of VDN, was investigated in \cite{van2016coordinated} for TSC, achieving promising results. Additionally, Liu et al. introduce a multi-agent dueling-double-deep Q network (MAD3QN) for TSC \cite{liu2021learning}. This framework features an innovative 
$\gamma$-reward design, which combines the traditional $\gamma$-reward with a novel $\gamma$-attention-reward. By employing a spatial differentiation method for agent coordination, their approach enables decentralized control and decoupling of road networks, enhancing scalability and convergence. In \cite{wang2021adaptive}, a cooperative group-based multi-agent Q-learning for ATSC (CGB-MATSC) is proposed. This method enhances the learning process by incorporating a k-nearest-neighbor approach for state representation, a pheromone-based strategy for creating regional green-wave traffic flows, and a spatially discounted reward system. 


On the other hand, \textit{policy-based} MARL algorithms have also been widely studied for TSC. In \cite{chu2019multi}, a fully scalable and decentralized Multi-Agent Actor-Critic (MA2C) algorithm is introduced, which incorporates an advanced discount factor to mitigate the effects of remote agents. This approach demonstrates significant advantages over random and independent controllers. 
Wu et al. introduced the multi-agent recurrent deep deterministic policy gradient (MARDDPG) algorithm, specifically designed for TSC in vehicular networks \cite{wu2020multi}. They utilize a strategy of CTDE, enhancing the efficiency of the training process through parameter sharing among the actor networks. Furthermore, the integration of LSTM networks enables the algorithm to utilize historical data for more effective control decisions.
In \cite{ma2020feudal}, a MA2C approach with a feudal hierarchy concept for TSC is presented. 
This approach segments the traffic network into several regions, each monitored by a manager agent, with traffic signal agents acting as workers. The managers are responsible for high-level coordination and setting objectives for the workers, who then adjust traffic signals to meet these objectives. This hierarchical structure fosters global coordination while ensuring the system scalability.

Efficient \textit{communication} is crucial in applying MARL to TSC as it enables coordinated decision-making, optimal resource allocation, adaptation to dynamic environments, scalability, conflict resolution, and efficient learning, all of which contribute to effective traffic management \cite{zhu2022survey}. 
Liu et al. present a novel algorithm that enhances communication efficiency through a new message exchange method and improves congestion measurement by introducing a more comprehensive reward calculation method. Then a clear and simple representation of traffic conditions is provided, and the synchronization between agents at different intersections with varying cycle lengths needs to be addressed \cite{liu2023traffic}.

\textit{Graph-based MARL} offers advantages in flexible representation, scalability, efficient message passing, a global view, interpretable representations, and adaptability, making it suitable for various multi-agent scenarios, including TSC problems \cite{liu2023graph}. 
In \cite{wei2019colight}, the authors present CoLight, a model leveraging graph-attentional MARL networks. This model facilitates communication among traffic signals, incorporating both temporal and spatial data from neighboring intersections to a target intersection. Crucially, it achieves index-free modeling of these neighboring intersections, enhancing its adaptability and efficiency.
In \cite{yang2021ihg}, Yang et al. propose an Inductive Heterogeneous Graph Multi-agent Actor-Critic (IHG-MA) algorithm  for controlling traffic signals across multiple intersections. This algorithm is distinguished by two main features: 
Firstly, it utilizes an inductive heterogeneous graph (IHG) neural network for representation learning, effectively encoding both heterogeneous features and structural information of nodes and graphs to generate embeddings for previously unseen nodes and new traffic networks. Secondly, IHG-MA adopts the MA2C framework for policy learning, utilizing these embeddings to compute Q-values and policies.

To account for the spatial influence of multi-intersection traffic lights and the temporal dependency on historical traffic conditions, Wang et al. introduce a novel spatio-temporal MARL framework (STMARL)  \cite{wang2020stmarl}. 
This framework captures  spatio-temporal dependencies among multiple traffic lights, coordinating their control via a traffic light adjacency graph. 
This graph is meticulously constructed to reflect the spatial structure of intersections and merges historical traffic data with the prevailing traffic conditions through the use of a recurrent neural network.
Additionally, a GNN model is designed to represent 
the relationships among various traffic lights, drawing on traffic information that varies over time.
Considering the importance of shared information, relying on heuristic methods. Jiang et al. propose a new communication form called UniComm, which incorporates numerous observations gathered at one intersection and predicts their impact on neighboring intersections, enhancing communication efficiency \cite{jiang2022multi}.
Antonio et al. leverage the multi-agent TD3 algorithm by incorporating an LSTM to address the issue of varying observation shapes based on the number of vehicles, and then employ a training method called curriculum through self-play to enable collaborative control of CAVs at intersections \cite{9762548}.

\subsubsection{\textbf{On-ramps merging}}
Highway on-ramp merging task requires the seamless integration of on-ramp CAVs into the main traffic flow without causing collisions (see Fig.~\ref{fig:three_dim} (b)). This complex maneuver necessitates that CAVs traveling in the through lane proactively modify their velocities—slowing down or speeding up as necessary—to create sufficient room for CAVs on the on-ramp to merge safely. Concurrently, CAVs on the ramp must regulate their velocities and make lane changes promptly in a timely manner when conditions are safe, thereby avoiding potential deadlock situations. This dual adjustment ensures a smooth and efficient merging process, critical for maintaining traffic flow and safety on highways \cite{yan2023multi, chen2023deep}.

In \cite{schester2019longitudinal}, a simplified mathematical formulation for an on-ramp merging scenario is presented to model the fundamental interactions and solved with a MADQN algorithm that accounts for the interaction between an on-ramp merging vehicle and a traffic vehicle in the target lane. The results indicate that a multi-agent approach can result in reduced collision rates compared to a single-agent approach, but this improvement is contingent on the optimality of the traffic vehicle's behavior in the target lane.
Zhou et al. introduce a cooperative merging control strategy for CAVs using a distributed MADDPG approach, which accounts for safe merging distances, acceleration limits, and factors like rear-end safety, lateral safety, and energy consumption \cite{zhou2022cooperative}. 
To address the challenge of a dynamic environment resulting from the decentralized learning of CAVs, a decentralized framework employing MADDPG is presented to coordinate CAVs during highway merging  \cite{nakka2022multi}. This framework enables policies learned from a small group of trained CAVs to be transferred and applied to any number of CAVs, with a reward function that promotes high-speed travel, resulting in smoother traffic flow while maintaining safety in terms of rear-end and lateral collisions.

Some studies also examine the influence of \textit{human drivers} in on-ramp merging scenarios. For instance, Hu et al. introduce interaction-aware decision-making with an adaptive strategies (IDAS) approach, which utilizes a modified MA2C method and curriculum learning to train a single policy that enables an AV to navigate the road while taking into account the cooperativeness of other drivers \cite{hu2019interaction}. 
In \cite{chen2023deep}, the authors introduce the MA2C approach, which allows CAVs on both the merge lane and through lane to cooperatively develop a policy. This approach focuses on optimizing traffic flow while also adjusting to the dynamics of HDVs.
Their efficient and scalable MARL framework accommodates dynamic traffic scenarios with varying communication topologies, utilizing parameter sharing, local rewards, and action masking to promote inter-agent cooperation and scalability. 
Furthermore, Chen et al. introduce a priority-based safety supervisor designed to minimize collision rates and accelerate the training process.
Similarly, a game-theoretic multi-agent planning approach called GAMEPLAN is introduced in \cite{chandra2022gameplan}, which incorporates both human drivers and CAVs in merging scenarios. This method leverages game theory and an auction mechanism to compute optimal actions for each agent, considering their driving style inferred from commonly available sensors. The algorithm assigns higher priority to more aggressive or impatient drivers and lower priority to more conservative or patient drivers, ensuring game-theoretic optimality while preventing collisions and deadlocks.

\begin{table*}[!th]
\centering
\caption{An overview of the primary works from four perspectives in different settings.}

\begin{tabularx}{\textwidth}{c | c | c c c X}
\hline
\shortstack{\textbf{Scenarios} \\ \quad \\ \quad} & \shortstack{\quad \\ \textbf{Typical} \\ \shortstack{\textbf{application}}} & \shortstack{\textbf{Work} \\ \quad \\ \quad}  & \shortstack{\textbf{MARL} \\ \textbf{algorithm}} & \shortstack{\textbf{Novelty} \\ \quad } & \shortstack{\textbf{\qquad \quad Metrics} \\ \quad \\ \quad} \\
\hline

\multirow{50}{*}{\shortstack{3-D \\ Cooperation}} & \multirow{15}{*}{\shortstack{Traffic \\ signal \\ control}} 

& \multicolumn{1}{c}{\shortstack{\cite{wu2019dcl}, 2019  \\ \quad \\ \quad \\ \quad \\ \quad \\ \quad}} & \multicolumn{1}{c}{\shortstack[c]{MADQN  \\ \quad \\ \quad \\ \quad \\ \quad \\ \quad}} & \multicolumn{1}{l}{\shortstack[l]{\quad \\ \scalebox{0.8}{$\bullet$} The state space of CAV decomposed into independent \\ and coordinated parts\\ 
\scalebox{0.8}{$\bullet$} Dynamic adaptation to mitigate the curse of \\dimensionality and  nonstationarity}} & \multicolumn{1}{l}{\shortstack[l]{ \scalebox{0.8}{$\bullet$} Travel efficiency\\ \scalebox{0.8}{$\bullet$} Intersection delay \\
\scalebox{0.8}{$\bullet$} Decision quality \\ \quad}} \\\cline{3-6}

&       & \multicolumn{1}{c}{\shortstack{\cite{liu2021learning}, 2021 \\ \quad \\ \quad \\ \quad \\ \quad}} & \multicolumn{1}{c}{\shortstack[c]{MADQN  \\ \quad \\ \quad \\ \quad \\ \quad}} & \multicolumn{1}{l}{\shortstack[l]{\quad \\ \scalebox{0.8}{$\bullet$} The $\gamma$-reward method to extend the  Markov chain to \\ the space-time domain\\ 
\scalebox{0.8}{$\bullet$} The spatial differentiation method  to amend the current\\ reward by recursion}} & \multicolumn{1}{l}{\shortstack[l]{\quad \\\scalebox{0.8}{$\bullet$} Asymmetry road \\ network evaluation\\ 
\scalebox{0.8}{$\bullet$} Attention score\\ 
\scalebox{0.8}{$\bullet$} Training time \\ \quad }}\\\cline{3-6}

&       & \multicolumn{1}{c}{\shortstack{\cite{wang2021adaptive}, 2021 \\ \quad \\ \quad\\ \quad\\ \quad \\ \quad\\ \quad \\ \quad\\ \quad }} & \multicolumn{1}{c}{\shortstack{CG-Based MA-\\Q Learning \\ (CGB-MAQL)\\ \quad \\ \quad \\ \quad \\ \quad\\ \quad}} & \multicolumn{1}{l}{\shortstack[l]{\quad \\ \scalebox{0.8}{$\bullet$} Agents organized into cooperative groups with each \\ agent responsible for its region \\ 
\scalebox{0.8}{$\bullet$} A k-nearest-neighbor-based  joint state representation \\ among neighboring agents\\ 
\scalebox{0.8}{$\bullet$} A heuristic training mechanism by terminating poor\\ behavior strategies}} & \multicolumn{1}{l}{\shortstack[l]{\scalebox{0.8}{$\bullet$} Waiting time\\ 
\scalebox{0.8}{$\bullet$} Road pheromone \\ \quad \\ \quad\\ \quad\\ \quad \\ \quad\\ \quad}} \\\cline{3-6}

&       & \multicolumn{1}{c}{\shortstack{\cite{liu2023traffic}, 2023 \\ \quad \\ \quad \\ \quad \\ \quad\\ \quad \\ \quad}} & \multicolumn{1}{c}{\shortstack{MA-Q\\ Learning \\ \quad\\ \quad\\ \quad\\ \quad}} & \multicolumn{1}{l}{\shortstack[l]{\quad \\ \scalebox{0.8}{$\bullet$} A new message sending and processing \\
\scalebox{0.8}{$\bullet$} A data structure to record the latest and most \\ valuable message \\
\scalebox{0.8}{$\bullet$}  A new reward calculation with both queue length and \\total waiting time}} & \multicolumn{1}{l}{\shortstack[l]{\scalebox{0.8}{$\bullet$} Waiting time\\ 
\scalebox{0.8}{$\bullet$} Reward \\ \quad \\ \quad \\ \quad\\ \quad\\ \quad}} \\\cline{2-6}

& \multirow{11}{*}{\shortstack{On-ramps \\ merging}} 

& \multicolumn{1}{c}{\shortstack{\cite{schester2019longitudinal}, 2019 \\ \quad \\ \quad \\ \quad \\ \quad\\ \quad}} & \multicolumn{1}{c}{\shortstack[c]{MA-Q\\ Learning\\ \quad\\ \quad\\ \quad}} & \multicolumn{1}{l}{\shortstack[l]{\quad \\ \scalebox{0.8}{$\bullet$} Multi-agent simulator proposed to study the interactions \\between a single pair of vehicles\\ 
\scalebox{0.8}{$\bullet$} A simplified mathematical formulation to capture  the \\ complexities of merging scenarios}} & \multicolumn{1}{l}{\shortstack[l]{\scalebox{0.8}{$\bullet$} Collision rates\\ 
\scalebox{0.8}{$\bullet$} Average speed \\ \quad\\ \quad\\ \quad}} \\\cline{3-6}

&       & \multicolumn{1}{c}{\shortstack{\cite{nakka2022multi}, 2022 \\ \quad \\ \quad\\ \quad}} & \multicolumn{1}{c}{\shortstack{MADDPG\\ \quad\\ \quad\\ \quad}} & \multicolumn{1}{l}{\shortstack[l]{\quad \\\scalebox{0.8}{$\bullet$} Transferring the policies through a few learning CAVs \\ in the training process\\
\scalebox{0.8}{$\bullet$} Reward function design at high speeds}} & \multicolumn{1}{l}{\shortstack[l]{\scalebox{0.8}{$\bullet$} Traffic flow \\ smoothness\\ \quad\\ \quad}} \\\cline{3-6}

&       & \multicolumn{1}{c}{\shortstack{\cite{zhou2022cooperative}, 2022 \\ \quad \\ \quad \\ \quad }} & \multicolumn{1}{c}{\shortstack{Distributed \\ MADDPG\\ \quad\\ \quad}} & \multicolumn{1}{l}{\shortstack[l]{\quad \\\scalebox{0.8}{$\bullet$} The interference of human-driven Vehicles (HDV)\\ 
\scalebox{0.8}{$\bullet$} Reward function design including rear-end \\ safety, lateral safety, and energy consumption}} & \multicolumn{1}{l}{\shortstack[l]{\scalebox{0.8}{$\bullet$} Fuel consumption\\
\scalebox{0.8}{$\bullet$} Travel time \\ \quad \\ \quad}} \\\cline{3-6}

&       & \multicolumn{1}{c}{\shortstack{\cite{chen2023deep}, 2023 \\ \quad \\ \quad \\ \quad \\ \quad\\ \quad \\ \quad}} & \multicolumn{1}{c}{\shortstack{MA2C \\ \quad \\ \quad \\ \quad \\ \quad\\ \quad\\ \quad\\ \quad}} & \multicolumn{1}{l}{\shortstack[l]{\quad \\ \scalebox{0.8}{$\bullet$} An action masking scheme \\ 
\scalebox{0.8}{$\bullet$} Parameter sharing and local rewards to encourage \\ inter-agent cooperation\\ 
\scalebox{0.8}{$\bullet$}  A novel priority-based safety supervisor  \\introduced to reduce collision rates}} & \multicolumn{1}{l}{\shortstack[l]{\scalebox{0.8}{$\bullet$} Traffic throughput\\ 
\scalebox{0.8}{$\bullet$} Collision rate \\ \quad\\ \quad\\ \quad\\ \quad\\ \quad}} \\\cline{2-6}

 & \multirow{12}{*}{\shortstack{Unsignalized \\ intersections}} 
 
& \multicolumn{1}{c}{\shortstack{\cite{xu2021leveraging}, 2021 \\ \quad \\ \quad \\ \quad \\ \quad \\ \quad}} & \multicolumn{1}{c}{\shortstack{MA-Q\\ Learning \\ \quad\\ \quad\\ \quad}} & \multicolumn{1}{l}{\shortstack[l]{\quad \\ \scalebox{0.8}{$\bullet$} The central section (CS) and the waiting section (WS) \\ to introduce a regulation scheme \\
\scalebox{0.8}{$\bullet$} Features incorporation such as fixed Q-targets and \\ experienced replay}} & \multicolumn{1}{l}{\shortstack[l]{\scalebox{0.8}{$\bullet$} Traffic throughput\\ \scalebox{0.8}{$\bullet$} Waiting time \\ \quad\\ \quad\\ \quad}} \\\cline{3-6}

&       & \multicolumn{1}{c}{\shortstack{\cite{spatharis2022multiagent}, 2022 \\ \quad \\ \quad \\ \quad \\ \quad\\ \quad\\ \quad\\ \quad\\ \quad}} & \multicolumn{1}{c}{\shortstack{MADDQN \\ \quad\\ \quad\\ \quad\\ \quad\\ \quad\\ \quad\\ \quad\\ \quad}} & \multicolumn{1}{l}{\shortstack[l]{\quad \\ \scalebox{0.8}{$\bullet$} Route-agents introduced to coordinate the actions of \\ multiple vehicles \\
\scalebox{0.8}{$\bullet$} The inclusion of a collision term in the reward function \\ enables cooperation \\
\scalebox{0.8}{$\bullet$} The reuse of knowledge from agents' policies to handle \\ unknown traffic scenarios}} & \multicolumn{1}{l}{\shortstack[l]{\scalebox{0.8}{$\bullet$} Safety\\
\scalebox{0.8}{$\bullet$} Travel time \\\scalebox{0.8}{$\bullet$} Fuel consumption \\ \quad\\ \quad\\ \quad\\ \quad\\ \quad}} \\\cline{3-6}

&       & \multicolumn{1}{c}{\shortstack{\cite{antonio2022multi}, 2022 \\ \quad \\ \quad \\ \quad \\ \quad }} & \multicolumn{1}{c}{\shortstack{QMIX \\ \quad\\ \quad\\ \quad\\ \quad}} & \multicolumn{1}{l}{\shortstack[l]{\quad \\ \scalebox{0.8}{$\bullet$} End-to-end learning to autonomously learn complex \\ real-life traffic dynamics\\ 
\scalebox{0.8}{$\bullet$} Curriculum through Self-Play to adapt and learn in new \\ and complex traffic scenarios}} & \multicolumn{1}{l}{\shortstack[l]{\scalebox{0.8}{$\bullet$} Travel time\\ 
\scalebox{0.8}{$\bullet$} Time lost due \\ to congestion\\ 
\scalebox{0.8}{$\bullet$} Waiting time }}\\\cline{3-6}

&       & \multicolumn{1}{c}{\shortstack{\cite{guo2022coordination}, 2022\\ \quad \\ \quad \\ \quad }} & \multicolumn{1}{c}{\shortstack{QMIX\\ \quad\\ \quad\\ \quad}} & \multicolumn{1}{l}{\shortstack[l]  {\quad \\ \scalebox{0.8}{$\bullet$} Network-level improvements and Q value updates using\\ Temporal Difference (TD) ($\lambda$) \\
\scalebox{0.8}{$\bullet$}  Reward clipping operations to enhance the QMIX}} & \multicolumn{1}{l}{\shortstack[l]{\scalebox{0.8}{$\bullet$} Average speed \\\scalebox{0.8}{$\bullet$} Collision rate \\\scalebox{0.8}{$\bullet$} Fuelconsumption}} \\\hline

\end{tabularx}
\label{table:33}
\end{table*}

\subsubsection{\textbf{Unsignalized intersections}}
Intersections serve as pivotal nodes and potential bottlenecks within urban road networks. Enhancing traffic efficiency at these junctions is essential in ameliorating overall traffic throughput and alleviating congestion. Nevertheless, the management of vehicular interactions at unsignalized intersections, as illustrated in Fig.~\ref{fig:three_dim} (b), presents a notably intricate challenge \cite{spatharis2022multiagent}. Many conventional techniques, such as rule-based \cite{aksjonov2021rule}, planning-based \cite{gu2016motion}, and single-agent RL methods \cite{10186792}, tend to approach the problem of intersection management as if it were a single-agent challenge, which can not capture the complex interactions among CAVs. 
Recently, MARL has emerged as a promising and effective tool for managing traffic at unsignalized intersections.

In \cite{spatharis2022multiagent}, an advanced collaborative multi-agent double Deep Q-Network (DQN) framework is proposed for unsigned traffic management. Their framework incorporates an efficient reward function that accounts for both the safety and efficiency of CAVs. Furthermore, this framework offers adaptability, enabling transfer learning and the reuse of knowledge from agent policies, particularly when dealing with unfamiliar traffic scenarios. 
In \cite{capasso2021end}, the authors introduce a multi-agent delayed-A3C (MAD-A3C) approach, which utilizes continuous, model-free DRL to train a neural network capable of predicting acceleration and steering angles of other road users at each time step. Their study illustrates that these CAV agents can acquire the essential rules for effectively navigating intersections by comprehending the priorities of other learners in the environment, all the while ensuring safe driving along their designated routes.

Yan et al. utilize  a multi-agent policy decomposition strategy, which allows for decentralized control using local observations across an arbitrary number of CAVs. Notably, without the need for reward shaping, the DQN-based method learns to  coordinate CAVs in a way that mimics traffic signal behaviors, attaining nearly optimal throughput with control over 33\% to 50\% of the vehicles.
Furthermore, through multi-task learning and transfer learning, they demonstrate the generalizability of this behavior across different inflow rates and traffic network sizes \cite{yan2021reinforcement}.

Effective \textit{coordination} among CAVs is essential to ensure safe and efficient maneuvers at unsignalized intersections. In \cite{mavrogiannis2020implicit}, a decentralized MADQN algorithm is adopted and learned to make decisions for CAVs. To enable effective coordination among agents, the intent trajectories of other neighboring agents are incorporated into each agent's state space. Furthermore, a decentralized and conflict-free coordination scheme designed for CAVs at unsignalized intersections is proposed in \cite{zheng2022deep}, with the objective of enhancing intersection management precision. They frame the challenge of safely guiding multiple vehicles through unsignalized intersections as a partially observable stochastic game (POSG). To facilitate collaborative decision-making in a distributed fashion, they introduce a cooperative multi-agent proximal optimization algorithm (CMAPPO).
In \cite{hamouda2021multi}, a multi-layer coordination strategy is proposed for the management of unsignalized intersections. This architecture comprises two layers: a low-level layer that employs a dynamics-based algorithm to control individual CAVs, and a high-level layer that integrates a twin delayed deep deterministic policy gradient (TD3) algorithm, trained in a centralized manner but executed in a decentralized fashion by multiple agents for decision-making. Their findings reveal a successful training process, with the MATD3 algorithm achieving a remarkable 100\% success rate in preventing intersection collisions.
To tackle coordination between adjacent intersections, a multi-agent-based deep reinforcement learning scheduling (MA-DRLS) algorithm is proposed in \cite{xu2021leveraging}. This approach allows each intersection agent to independently devise an optimal scheduling strategy through information exchange with other agents. The algorithm employs DQN networks, and the use of fixed Q-targets and experience replay helps improve the neural network's reliability during the training process.
In \cite{tallapragada2023reinforcement}, a multi-agent joint-action DDPG (MAJA-DDPG) framework that integrates learning and sequential optimization is proposed for the management of unsignalized intersections. 
The algorithm involves two primary steps: 
Initially, a shared policy is learned to determine the order in which vehicles cross at the intersection based on traffic state information. Subsequently, vehicle trajectories are optimized sequentially following the determined crossing order. Notably, the proposed algorithm learns a shared policy that can be implemented in a distributed manner.

To address the challenge of limited communication resources, the authors in \cite{li2021learning} introduce an algorithm known as Efficient Communication Method (ECM)-MA2C. This approach aims to maintain coordination performance while making efficient use of constrained communication resources. The ECM-MA2C algorithm employs a variational auto-encoder algorithm combined with advanced multi-head attention mechanisms to extract and retain valuable information from neighboring agents. Experimental results demonstrate that this efficient communication scheme outperforms baseline approaches in bandwidth-constrained environments, highlighting its effectiveness in achieving improved performance with limited communication resources.

In \cite{guo2022coordination}, the authors have concentrated their efforts on the application of a value decomposition-based MARL approach, namely QMIX, for the coordination of multiple CAVs at unsignalized intersections. To enhance the performance of the original QMIX framework, several implementation enhancements have been integrated. These improvements encompass network-level optimizations, Q-value updates using Temporal Difference (TD) with $\gamma$, and reward clipping operations.

\begin{figure*}[!ht]
\centering
\centerline{\includegraphics[width=0.98\textwidth]{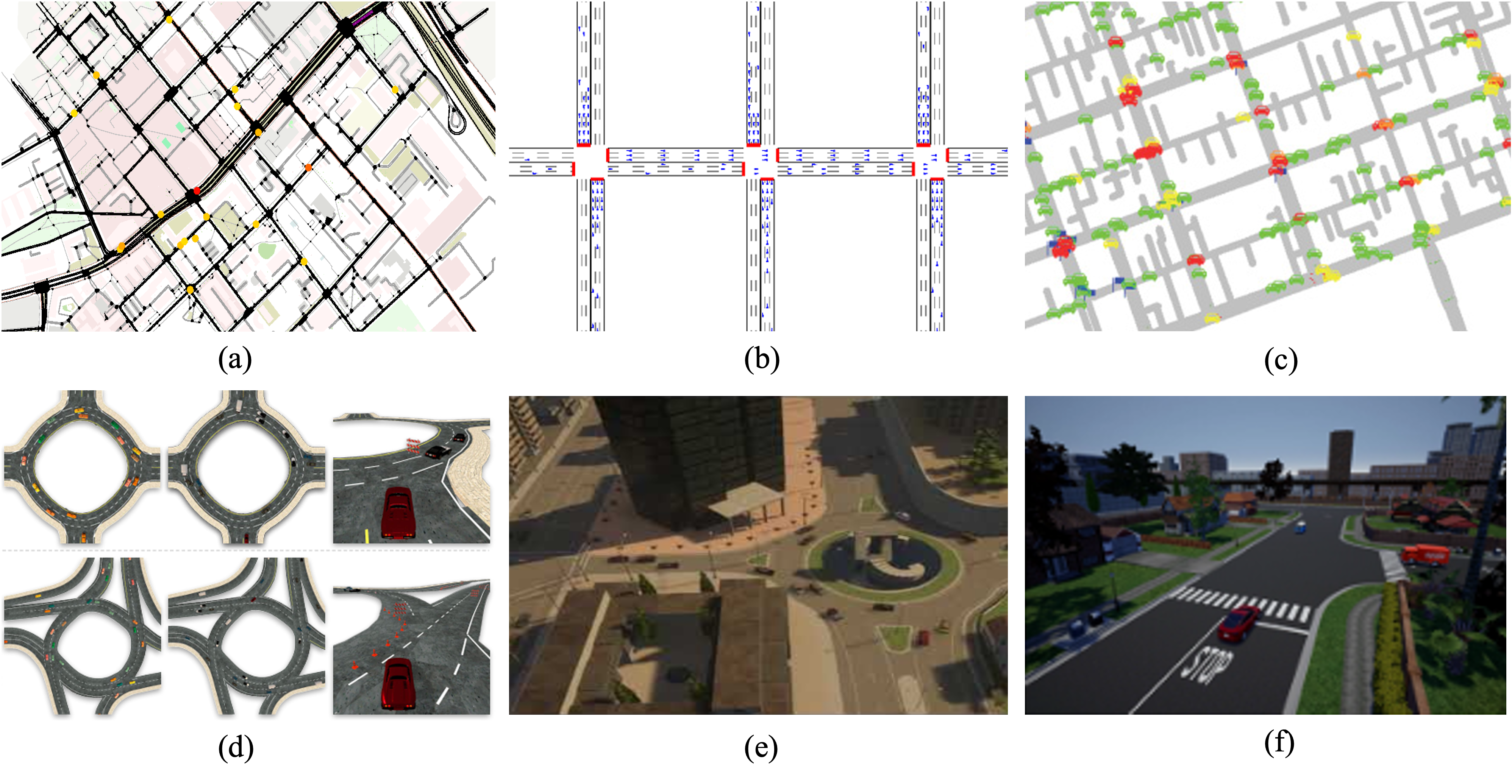}}
\caption{Representative scenarios in different simulators: (a) SUMO \cite{lopez2018microscopic}; (b) CityFlow  \cite{zhang2019cityflow}; (c) SMARTS \cite{zhou2020smarts}; (d) MetaDrive \cite{li2022metadrive}; (e) CARLA \cite{dosovitskiy2017carla}; (f) MACAD \cite{palanisamy2020multi}.}
\label{fig7}
\vspace{-10pt}
\end{figure*}

\subsection{Simulation platforms} 
\label{sec:sim}
The simulation platforms, which act as the training/learning environments, play an important role in the performance and adaptability of the autonomous agents. The learning platforms provide different challenges for the planning and control of the autonomous agents by considering different traffic scenarios and different characters of agents (Fig. \ref{fig7}). These platforms offer a virtual environment to learn and validate the CAV's perception, decision-making, and low-level control systems, where the cost and the safety of the agents can be optimized. The traffic ecosystem/simulation platform incorporates realistic models, physics-based simulations, and efficient numerical solvers, providing a controllable environment to accelerate the development of CAVs \cite{chen2019novel}. Comparisons of different simulation platforms for CAVs are shown in Table \ref{table:3}.


\textbf{SUMO} (Simulation of Urban MObility) is an open-source traffic simulation package used to model CAVs and their interactions \cite{lopez2018microscopic}. It provides tools for simulating traffic scenarios, including vehicle behavior and road networks, enabling researchers to study and analyze traffic dynamics. SUMO aids in evaluating traffic management strategies, routing algorithms, and the impact of CAVs on traffic flow and efficiency. It is widely used for assessing CAV behavior and optimizing transportation systems in a virtual environment.

\textbf{CityFlow}, by optimizing data structures and using efficient algorithms, improves the simulation efficiency and accelerates the learning tasks, allowing for large-scale road networks and traffic flow simulations \cite{zhang2019cityflow}. It supports flexible definitions of road networks using synthetic and real-world data and offers a user-friendly interface for reinforcement learning. CityFlow outperforms SUMO, being over twenty times faster and capable of city-wide traffic simulations with interactive rendering. It opens new possibilities for testing machine learning methods in the intelligent transportation domain, serving as a foundation for transportation studies beyond traffic signal control.

To mitigate the gaps between learning and reality, learning platforms with more realistic agent characters are required, and several simulation platforms considering vehicle dynamics and realistic sensor data are proposed. 

 \textbf{SMARTS} (Scalable Multi-Agent RL Training School) is a dedicated simulation platform for realistic multi-agent interaction in autonomous driving research \cite{zhou2020smarts}. SMARTS supports the training and accumulation of diverse behavior models, enabling the creation of realistic and varied driving interactions. It offers a user-friendly interface, key features, and benchmark tasks. The platform is open-source, encouraging research on multi-agent learning for autonomous driving.

\textbf{MetaDrive} stands out as an efficient and compositional driving simulator, offering infinite scene generation, lightweight operation, realistic physics simulation, and multiple sensory inputs \cite{li2022metadrive}, where a trade-off between the visual rendering and the physical simulation has been made.  With the ability to generate infinite scenes encompassing diverse road maps and traffic settings, researchers can explore a wide range of driving environments, paving the way for the development of generalizable RL algorithms. These features make it a valuable tool for researchers in autonomous driving and robotics, enabling them to conduct impactful studies with comprehensive realism and flexibility.

\textbf{CARLA}, which is an open-source simulator developed by the team at Intel Labs, considers more comprehensive environment and agent variances. It provides a realistic 3D environment for testing and evaluating autonomous driving algorithms, including support for multi-agent scenarios \cite{dosovitskiy2017carla}. With CARLA, researchers and developers can create complex simulations that closely resemble real-world driving conditions, enabling them to assess and refine the performance of their autonomous driving systems. Its high-fidelity physics engine and accurate sensor models allow for realistic perception and control experiments. Furthermore, CARLA offers extensive customization options, facilitating the creation of various urban environments, weather conditions, and traffic scenarios to comprehensively evaluate the robustness and effectiveness of autonomous driving algorithms.

\textbf{MACAD} (Multi-Agent Connected and Autonomous Driving)platform serves as a high-fidelity simulation environment for CAVs, utilizing "partially observable Markov games" to accurately depict the challenges of connected autonomous driving with realistic assumptions and models. This platform offers a comprehensive range of CAD simulation settings, facilitating the exploration and advancement of Deep RL strategies in integrated sensing, perception, planning, and control for CAD systems across diverse operational design domains, under realistic multi-agent conditions, as highlighted by \cite{palanisamy2020multi}. Furthermore, MACAD categorizes multi-agent learning scenarios according to task types, agent characteristics, and environmental factors, providing researchers with customizable options. 

More simulators for CAVs have been designed to consider more detailed characters and versatile scenarios, which provides more tools to mitigate the gaps between the simulation and reality.

A \textbf{PCMA} (pedestrian crash avoidance mitigation) system is developed for interdependent decisions of the vehicles and pedestrians by Trumpp et. al \cite{trumpp2022modeling}. In the system, an AV-pedestrian interaction at an unmarked crosswalk is modeled using a Markov decision process and deep reinforcement learning (DRL). Two pedestrian behaviors are considered: a baseline predefined strategy and an advanced model using DRL, turning the interaction into a multi-agent problem. The PCAM systems are assessed based on collision rates and traffic flow efficiency, focusing on the impact of observation uncertainty. 

\textbf{Nocturne} considers the limited visibility for studying the multi-agent coordination in the 2D driving simulator, as the complex partial observablilities of the multi-agent system also attract the attention of the researchers. It efficiently calculates visible features, allowing for rapid simulations. Using real-world driving data, tests showed that machine-learning agents significantly differed from human-like coordination \cite{vinitsky2022nocturne}.

Generally speaking, the simulation-based evaluations are simpler to implement, reproduce, and scale compared with real experiments for learning. However, simulations may not capture all the challenges associated with an actual deployment. For example, factors such as network delay, vehicle model discrepancies, computation time, and the necessity of implementing clock synchronization and fail-safe routines pose challenges in real-world implementations.

\begin{table*}[!th]\label{tab:sim}
\centering
\caption{Comparasions between some typical CAV simulation platforms.}

\begin{tabularx}{\textwidth}{X c c c c c c c} \hline
  \begin{tabular}{l}
      \textbf{Simulators}
  \end{tabular} & 
  \begin{tabular}{c}
       \textbf{Vehicle} \\
       \textbf{Dynamics}
  \end{tabular} & 
  \begin{tabular}{c}
       \textbf{3D} \\
       \textbf{Scenarios}
  \end{tabular} & 
  \begin{tabular}{c}
       \textbf{Visual} \\
       \textbf{Rendering}
  \end{tabular} &
  \begin{tabular}{c}
       \textbf{Pedestrians}
  \end{tabular} & 
  \begin{tabular}{c}
       \textbf{Light} \\
       \textbf{Weighted}
  \end{tabular} & 
  \begin{tabular}{c}
      \textbf{Customized} \\
      \textbf{Maps}
  \end{tabular} & 
  \begin{tabular}{c}
      \textbf{Data}\\
      \textbf{Importing}
  \end{tabular}
  \\
\hline
\begin{tabular}{@{}c@{}}\textbf{Hightway-env} \cite{highway-env}\end{tabular}
 & - & -& - & - & \checkmark & \checkmark & - \\
\textbf{Nocturne} \cite{vinitsky2022nocturne} &  - & - & - & \checkmark&  \checkmark & - & \checkmark\\
\textbf{SUMO} \cite{lopez2018microscopic} &  - & - & - & \checkmark&  \checkmark & \checkmark & \checkmark\\
\textbf{CityFlow} \cite{zhang2019cityflow} & - & - & - & -& \checkmark & \checkmark & \checkmark\\
\textbf{SMARTS} \cite{zhou2020smarts} & \checkmark & - & - & - & \checkmark & \checkmark & - \\
\textbf{MetaDrive} \cite{li2022metadrive} & \checkmark & \checkmark & - & - & \checkmark & \checkmark &\checkmark \\
\textbf{CARLA} \cite{dosovitskiy2017carla} & \checkmark & \checkmark & \checkmark & \checkmark & - & \checkmark & \checkmark \\
\textbf{MACAD} \cite{palanisamy2020multi} & \checkmark & \checkmark & \checkmark & \checkmark & - & \checkmark & \checkmark \\ \hline
\end{tabularx}
\label{table:3}
\vspace{-10pt}
\end{table*}

\section{Discussion \& Future Work}\label{sec:dis}
CAVs leveraging MARL algorithms present a promising frontier in the evolution of transportation. As urban centers grapple with traffic congestion and road safety, the nuanced and adaptive behaviors offered by MARL algorithms can enable CAVs to respond more fluidly to the actions of other road participants. The dynamism of MARL allows these vehicles to cooperatively optimize traffic flows, potentially alleviating congestion and reducing transit times. Nevertheless, the journey to this future is not without its challenges. Interactions with unpredictable human drivers, managing the delicate balance between exploration and safety in learning, and navigating a complex regulatory landscape are just a few of the hurdles. Yet, with ongoing collaborations among tech giants, startups, and academic circles, there is a collective push to overcome these barriers, signaling an optimistic trajectory for the marriage of MARL and CAVs.
\vspace{-10pt}
\subsection{Macro-micro energy optimization of CAVs using MARL}
Efficient energy management in CAVs is crucial, as it not only ensures the optimal performance of each vehicle but also contributes to overall energy conservation \cite{han2018safe, rios2018impact}. This can be divided into two distinct yet complementary approaches: microscopic energy optimization, which concentrates on maximizing the efficiency of individual components within a single vehicle, and macroscopic energy optimization, which aims to enhance collective energy use across a fleet of AVs.

\textbf{\textit{Microscopic energy optimization}} evaluates the performance of each component, aiming for optimal functioning. Traditional methods, which segregate powertrain, chassis control, and other controls, overlook the synergistic potential between these elements, potentially missing out on energy savings. To overcome this, Hua et al. present a collaborative strategy between two learning agents within the MADRL framework \cite{HUA2023121526}. They conduct a sensitivity analysis to harmonize the settings for two learning agents, identifying an optimal mode that surpasses single-agent models by saving up to 4\% more energy.
Yang et al. develop a multi-objective energy management system (EMS) for hybrid electric vehicles (HEV) using a blend of game theory and RL within a MARL framework \cite{yang2023multi}. Their strategy simultaneously addresses fuel economy, battery charge maintenance, battery longevity, and ultracapacitor constraints, considering the engine–generator set (EGS) and hybrid energy storage system (HESS) as interactive agents.
Shi et al. propose an EMS that uses IQL to sustain battery charge levels and minimize hydrogen consumption, thereby optimizing energy for HESS\cite{shi2022online}. 
Microscopic energy optimization enhances component efficiency but risks overlooking vehicle-wide dynamics. Macroscopic optimization addresses this by aiming to improve the energy efficiency across a fleet of autonomous vehicles.

\textbf{\textit{Macroscopic energy optimization}} has garnered increasing interest, particularly concerning traffic dynamics and their impact on energy efficiency. For instance, Wang et al. have developed a MARL-based strategy for HEVs that synchronizes powertrain management with car-following behaviors, thereby reducing energy use while maintaining safe distances \cite{wang2023cooperative}.
Zhang et al. have designed a MARL method that simultaneously optimizes velocity and energy management for groups of HEVs, employing Markov games and LSTM networks for dynamic group understanding and an innovative asynchronous learning technique for inter-vehicle knowledge sharing\cite{zhang2023integrated}. 
Furthermore, Peng et al. utilize a MADDPG framework with two distinct agents for adaptive cruise control and energy management, respectively, achieving domain-specific optimization and enhancing the MADDPG's effectiveness by more than 10\% with an improved prioritized experience replay \cite{peng2023ecological}.

\begin{figure}[htbp]
\centerline{\includegraphics[width=\columnwidth]{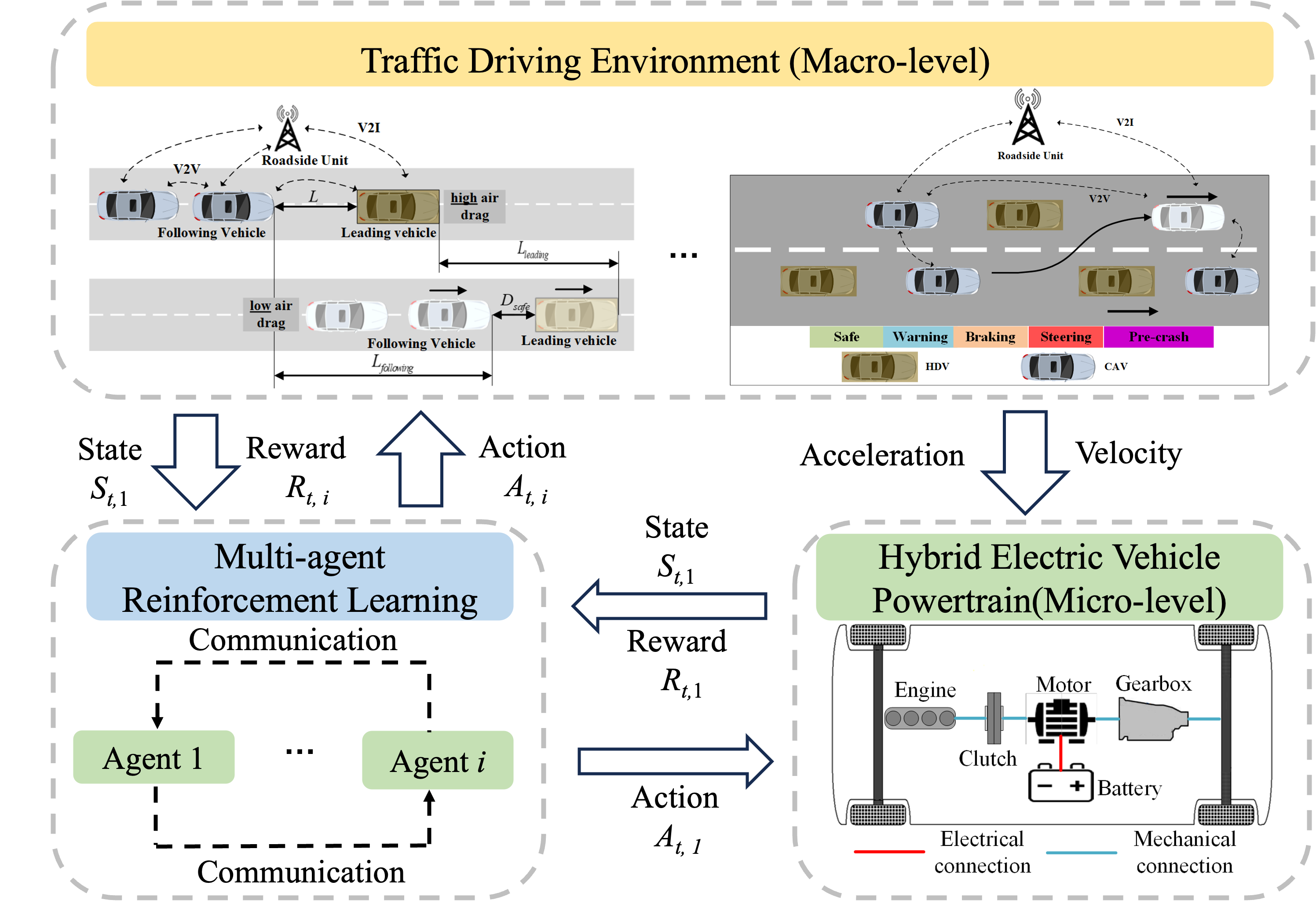}}
\caption{Illustration of the macro-micro energy management system control framework.}
\label{fig:3.1}
\vspace{-10pt}
\end{figure}

\textbf{\textit{Macro-micro energy optimization:}}
The advancement of CAVs is set to enhance sustainable and efficient mobility, with macro-micro energy modeling playing a key role in precise energy consumption analysis, as depicted in Fig.~\ref{fig:3.1}. This approach intertwines macro-level traffic analysis with micro-level HEV powertrain operations. By doing so, it allows for adaptive energy management that responds to real-time traffic conditions and powertrain status, optimizing the use of the electric motor and battery in various driving scenarios. Consequently, this integrated optimization strategy not only improves route efficiency and driving behavior but also leads to notable energy savings and cost reductions \cite{shang2023estimation}.

\vspace{-10pt}
\subsection{Communication challenges}
\textbf{\textit{Current progress:}} Effective communication is pivotal in MARL for CAVs, enabling collaboration, information sharing, and coordinated decision-making through reliable protocols \cite{rahman2021multi}. 
First, the development of advanced communication protocols tailored specifically for CAVs is a pressing need. Existing research focuses on protocols that enable efficient data sharing \cite{hasan2020securing}. For instance, research endeavors have delved into the utilization of graph-based MARL algorithms \cite{chen2021graph}, such as those explored in \cite{liu2023graph} and \cite{yang2021ihg}, to enable efficient and adaptive message exchange among agents. 
Second, the challenge of ensuring robust communication in dynamically evolving traffic scenarios represents another critical aspect that deserves meticulous attention \cite{valiente2022robustness}. For example, in \cite{li2021learning}, a variational auto-encoder algorithm combined with advanced multi-head attention mechanisms is proposed to extract and retain valuable information from neighboring agents while making efficient use of constrained communication resources. 
Last but not least,  it is worth noting that managing unstable communication channels, particularly in terms of latency, emerges as a particularly intriguing consideration in the context of MARL for CAVs \cite{vu2020multi}. In \cite{vu2020multi}, the double deep Q-learning algorithm is proposed to jointly train the agents to maximize the sum rate of V2N links while ensuring the desired packet delivery probability for each V2V link within specified latency constraints.

\textbf{\textit{Research gaps:}} While communication in MARL is a well-explored domain, its adaptation and application within CAVs present distinctive research gaps, which necessitate further exploration \cite{zhu2022survey}. For instance, existing works often overlook pressing real-world concerns in the realm of CAVs, such as communication costs and the challenges posed by noisy environments. These critical factors, when ignored, can introduce substantial obstacles to the practical deployment of CAV technology. MARL algorithms applied in computer science domains are expected to be explored for use in CAV applications within challenging environments marked by limited bandwidth \cite{wang2020learning} and noisy communication channels \cite{freed2020communication}. Moreover, the majority of current MARL algorithms in CAVs rely on Centralized Training and Decentralized Execution (CTDE, Sec.~\ref{sec:backgrounds}), potentially posing a significant privacy threat within the realm of communication. In the field of single-agent RL, security and privacy issues have garnered extensive attention \cite{lei2023new}. However, in the context of MARL, and similarly in the domain of CAVs, these areas have remained relatively unexplored. This suggests a promising area for future research that merits exploration.

\vspace{-10pt}
\subsection{Mixed-traffic challenges}
\textbf{\textit{Current progress:}} One of the formidable challenges posed by CAVs stems from navigating the intricate dynamics of mixed traffic, wherein CAVs share the road with various other road users, including human-driven vehicles, bicycles, e-scooters, and an array of other modes of transportation \cite{valiente2023learning, zhao2019identification, chen2024lane}. This multifaceted coexistence necessitates advanced MARL algorithms to ensure seamless and safe interactions among these diverse road users. Current research into MARL algorithms for CAV applications primarily concentrates on integrating the prediction of human behavior into the decision-making process \cite{hu2019interaction, dai2023bargaining}. For instance, in \cite{chen2023deep}, the authors introduce a novel MARL framework with a priority-based safety supervisor designed to preempt potential collisions. This is achieved by predicting future trajectories, encompassing both CAVs and HDVs and then the unsafe actions generated by MARL agents will be replaced with safe actions, thereby contributing to the enhancement of safety in mixed traffic scenarios. However, it is important to note that human-driven vehicles can introduce an additional layer of unpredictability into the traffic environment, since mixed traffic involves a diverse range of road users, presenting unique challenges for the seamless integration of CAVs into the existing road ecosystem. 

\textbf{\textit{Research gaps:}} Given the diverse spectrum of road users, future research endeavors will encompass the development of real-time, adaptive safety assessment and intervention strategies capable of responding to the dynamic nature of traffic conditions. Another significant aspect entails the establishment of novel coordination and communication mechanisms between CAVs and other road users. For instance, as the presence of e-scooters on the roads continues to surge, there is a growing imperative to develop efficient methods for coordinating the interactions between CAVs and e-scooters \cite{gilroy2022scooter}. Furthermore, the existing MARL algorithms are typically trained within specific scenarios, underscoring the necessity for research aimed at enhancing their capacity to generalize across a wide range of mixed traffic conditions and environmental contexts.

\vspace{-10pt}
\subsection{Sim-to-real challenges}
Deep MARL has achieved significant breakthroughs in various areas of the CAV domain. Given the constraints associated with real-world data collection, such as sample inefficiency and cost considerations, simulation environments have become invaluable for agent training (see Sec.~\ref{sec:sim}). Nevertheless, the disparity between the simulated and real-world settings can lead to a degradation in policy performance when models are transitioned to actual CAV scenarios \cite{zhao2020sim}. In certain simulator platforms, a key objective is to enable the deployment of CAV agents with minimal disparities between their training data and real-world experiences \cite{chu2019multi}. Simultaneously, other research initiatives are dedicated to enhancing safety, recognizing it as a primary obstacle to achieving real-world online training for complex self-driving car agents \cite{chen2023deep}. Conversely, it is crucial to explore advanced MARL techniques, including approaches like multi-agent offline RL \cite{yang2021believe} that leverage pre-existing offline data to update policies with online adaptations. Such future exploration can open new avenues for enhancing the generalization abilities and performance of CAVs in real-world scenarios.

\section{Conclusion}\label{sec:conl}
Recently, MARL has emerged as a focal point of interest within the realm of CAVs. The capability of MARL to tackle intricate control and coordination challenges in CAVs has unlocked new horizons for the development of intelligent and efficient next-generation transportation systems. This review has undertaken a comprehensive exploration of the applications of MARL in the control of CAVs. 

\begin{enumerate}
    \item The review began with an overview of single-agent RL techniques and an extensive survey of the diverse landscape of MARL architectural variants, facilitating a deeper understanding of applications to both individual and collective agent behaviors.
    \item Our analysis categorized these contributions based on different degrees of control, ranging from one-dimensional to two-dimensional and three-dimensional collaboration. Within this multifaceted arena, MARL has exhibited remarkable prowess, demonstrating its promising performance in various CAV control tasks. These include cooperative endeavors such as platooning control, lane-changing maneuvers, on-ramp merging, traffic signal coordination, and unsignalized intersections, among others.
    \item We highlighted significant challenges in designing and evaluating MARL methods within the CAV. Striking a balance between performance, scalability, and safety remains a complex challenge.
\end{enumerate}  

This review serves as a crucial resource, aimed at encouraging further research and the practical application of MARL within the field of CAVs.

\bibliographystyle{IEEEtran}
\bibliography{Main}

\begin{thebibliography}{100}
\providecommand{\url}[1]{#1}
\csname url@samestyle\endcsname
\providecommand{\newblock}{\relax}
\providecommand{\bibinfo}[2]{#2}
\providecommand{\BIBentrySTDinterwordspacing}{\spaceskip=0pt\relax}
\providecommand{\BIBentryALTinterwordstretchfactor}{4}
\providecommand{\BIBentryALTinterwordspacing}{\spaceskip=\fontdimen2\font plus
\BIBentryALTinterwordstretchfactor\fontdimen3\font minus
  \fontdimen4\font\relax}
\providecommand{\BIBforeignlanguage}[2]{{%
\expandafter\ifx\csname l@#1\endcsname\relax
\typeout{** WARNING: IEEEtran.bst: No hyphenation pattern has been}%
\typeout{** loaded for the language `#1'. Using the pattern for}%
\typeout{** the default language instead.}%
\else
\language=\csname l@#1\endcsname
\fi
#2}}
\providecommand{\BIBdecl}{\relax}
\BIBdecl

\bibitem{dimitrakopoulos2010intelligent}
G.~Dimitrakopoulos and P.~Demestichas, ``Intelligent transportation systems,''
  \emph{IEEE Vehicular Technology Magazine}, vol.~5, no.~1, pp. 77--84, 2010.

\bibitem{kehoe2015survey}
B.~Kehoe, S.~Patil, P.~Abbeel, and K.~Goldberg, ``A survey of research on cloud
  robotics and automation,'' \emph{IEEE Transactions on automation science and
  engineering}, vol.~12, no.~2, pp. 398--409, 2015.

\bibitem{proia2021control}
S.~Proia, R.~Carli, G.~Cavone, and M.~Dotoli, ``Control techniques for safe,
  ergonomic, and efficient human-robot collaboration in the digital industry: A
  survey,'' \emph{IEEE Transactions on Automation Science and Engineering},
  vol.~19, no.~3, pp. 1798--1819, 2021.

\bibitem{bi2021hybrid}
J.~Bi, X.~Zhang, H.~Yuan, J.~Zhang, and M.~Zhou, ``A hybrid prediction method
  for realistic network traffic with temporal convolutional network and lstm,''
  \emph{IEEE Transactions on Automation Science and Engineering}, vol.~19,
  no.~3, pp. 1869--1879, 2021.

\bibitem{wu2023finite}
K.~Wu, J.~Hu, Z.~Ding, and F.~Arvin, ``Finite-time fault-tolerant formation
  control for distributed multi-vehicle networks with bearing measurements,''
  \emph{IEEE Transactions on Automation Science and Engineering}, vol.~21,
  no.~2, pp. 1346--1357, 2023.

\bibitem{li2023traffic}
B.~Li, W.~Zhuang, H.~Zhang, H.~Sun, H.~Liu, J.~Zhang, G.~Yin, and B.~Chen,
  ``Traffic-aware ecological cruising control for connected electric vehicle,''
  \emph{IEEE Transactions on Transportation Electrification}, 2023.

\bibitem{hua2020research}
M.~Hua, G.~Chen, C.~Zong, and L.~He, ``Research on synchronous control strategy
  of steer-by-wire system with dual steering actuator motors,''
  \emph{International Journal of Vehicle Autonomous Systems}, vol.~15, no.~1,
  pp. 50--76, 2020.

\bibitem{he2020admittance}
W.~He, C.~Xue, X.~Yu, Z.~Li, and C.~Yang, ``Admittance-based controller design
  for physical human--robot interaction in the constrained task space,''
  \emph{IEEE Transactions on Automation Science and Engineering}, vol.~17,
  no.~4, pp. 1937--1949, 2020.

\bibitem{garg2022systematic}
T.~Garg and G.~Kaur, ``A systematic review on intelligent transport systems,''
  \emph{Journal of Computational and Cognitive Engineering}, 2022.

\bibitem{hua2019hierarchical}
M.~Hua, G.~Chen, B.~Zhang, and Y.~Huang, ``A hierarchical energy efficiency
  optimization control strategy for distributed drive electric vehicles,''
  \emph{Proceedings of the Institution of Mechanical Engineers, Part D: Journal
  of Automobile Engineering}, vol. 233, no.~3, pp. 605--621, 2019.

\bibitem{lin2020comparison}
Y.~Lin, J.~McPhee, and N.~L. Azad, ``Comparison of deep reinforcement learning
  and model predictive control for adaptive cruise control,'' \emph{IEEE
  Transactions on Intelligent Vehicles}, vol.~6, no.~2, pp. 221--231, 2020.

\bibitem{katriniok2022fully}
A.~Katriniok, B.~Rosarius, and P.~M{\"a}h{\"o}nen, ``Fully distributed model
  predictive control of connected automated vehicles in intersections: Theory
  and vehicle experiments,'' \emph{IEEE Transactions on Intelligent
  Transportation Systems}, vol.~23, no.~10, pp. 18\,288--18\,300, 2022.

\bibitem{chen2024communication}
D.~Chen, K.~Zhang, Y.~Wang, X.~Yin, Z.~Li, and D.~Filev,
  ``Communication-efficient decentralized multi-agent reinforcement learning
  for cooperative adaptive cruise control,'' \emph{IEEE Transactions on
  Intelligent Vehicles}, 2024.

\bibitem{liu2023systematic}
W.~Liu, M.~Hua, Z.~Deng, Z.~Meng, Y.~Huang, C.~Hu, S.~Song, L.~Gao, C.~Liu,
  B.~Shuai \emph{et~al.}, ``A systematic survey of control techniques and
  applications in connected and automated vehicles,'' \emph{IEEE Internet of
  Things Journal}, 2023.

\bibitem{silver2016mastering}
D.~Silver, A.~Huang, C.~J. Maddison, A.~Guez, L.~Sifre, G.~Van Den~Driessche,
  J.~Schrittwieser, I.~Antonoglou, V.~Panneershelvam, M.~Lanctot \emph{et~al.},
  ``Mastering the game of go with deep neural networks and tree search,''
  \emph{nature}, vol. 529, no. 7587, pp. 484--489, 2016.

\bibitem{lee2020learning}
J.~Lee, J.~Hwangbo, L.~Wellhausen, V.~Koltun, and M.~Hutter, ``Learning
  quadrupedal locomotion over challenging terrain,'' \emph{Science robotics},
  vol.~5, no.~47, p. eabc5986, 2020.

\bibitem{bai2022group}
X.~Bai, A.~Fielbaum, M.~Kronm{\"u}ller, L.~Knoedler, and J.~Alonso-Mora,
  ``Group-based distributed auction algorithms for multi-robot task
  assignment,'' \emph{IEEE Transactions on Automation Science and Engineering},
  vol.~20, no.~2, pp. 1292--1303, 2022.

\bibitem{wang2024prescribed}
X.~Wang, D.~Ye, L.~Zhang, and X.~Zhao, ``Prescribed performance tracking
  control for nonlinear multiagent systems with distributed observation errors
  compensation,'' \emph{IEEE Transactions on Automation Science and
  Engineering}, 2024.

\bibitem{ganesh2022review}
A.~H. Ganesh and B.~Xu, ``A review of reinforcement learning based energy
  management systems for electrified powertrains: Progress, challenge, and
  potential solution,'' \emph{Renewable and Sustainable Energy Reviews}, vol.
  154, p. 111833, 2022.

\bibitem{liu2023safe}
Z.~E. Liu, Q.~Zhou, Y.~Li, S.~Shuai, and H.~Xu, ``Safe deep reinforcement
  learning-based constrained optimal control scheme for hev energy
  management,'' \emph{IEEE Transactions on Transportation Electrification},
  2023.

\bibitem{liu2024deep}
Z.~E. Liu, Y.~Li, Q.~Zhou, Y.~Li, B.~Shuai, H.~Xu, M.~Hua, G.~Tan, and L.~Xu,
  ``Deep reinforcement learning based energy management for heavy duty hev
  considering discrete-continuous hybrid action space,'' \emph{IEEE
  Transactions on Transportation Electrification}, 2024.

\bibitem{chen2020autonomous}
D.~Chen, L.~Jiang, Y.~Wang, and Z.~Li, ``Autonomous driving using safe
  reinforcement learning by incorporating a regret-based human lane-changing
  decision model,'' in \emph{2020 American Control Conference (ACC)}.\hskip 1em
  plus 0.5em minus 0.4em\relax IEEE, 2020, pp. 4355--4361.

\bibitem{shu2021driving}
H.~Shu, T.~Liu, X.~Mu, and D.~Cao, ``Driving tasks transfer using deep
  reinforcement learning for decision-making of autonomous vehicles in
  unsignalized intersection,'' \emph{IEEE Transactions on Vehicular
  Technology}, vol.~71, no.~1, pp. 41--52, 2021.

\bibitem{van2016deep}
H.~Van~Hasselt, A.~Guez, and D.~Silver, ``Deep reinforcement learning with
  double q-learning,'' in \emph{Proceedings of the AAAI conference on
  artificial intelligence}, vol.~30, no.~1, 2016.

\bibitem{zhou2019development}
M.~Zhou, Y.~Yu, and X.~Qu, ``Development of an efficient driving strategy for
  connected and automated vehicles at signalized intersections: A reinforcement
  learning approach,'' \emph{IEEE Transactions on Intelligent Transportation
  Systems}, vol.~21, no.~1, pp. 433--443, 2019.

\bibitem{shi2021connected}
H.~Shi, Y.~Zhou, K.~Wu, X.~Wang, Y.~Lin, and B.~Ran, ``Connected automated
  vehicle cooperative control with a deep reinforcement learning approach in a
  mixed traffic environment,'' \emph{Transportation Research Part C: Emerging
  Technologies}, vol. 133, p. 103421, 2021.

\bibitem{qu2020jointly}
X.~Qu, Y.~Yu, M.~Zhou, C.-T. Lin, and X.~Wang, ``Jointly dampening traffic
  oscillations and improving energy consumption with electric, connected and
  automated vehicles: a reinforcement learning based approach,'' \emph{Applied
  Energy}, vol. 257, p. 114030, 2020.

\bibitem{chandra2022gameplan}
R.~Chandra and D.~Manocha, ``Gameplan: Game-theoretic multi-agent planning with
  human drivers at intersections, roundabouts, and merging,'' \emph{IEEE
  Robotics and Automation Letters}, vol.~7, no.~2, pp. 2676--2683, 2022.

\bibitem{amrouni2021abides}
S.~Amrouni, A.~Moulin, J.~Vann, S.~Vyetrenko, T.~Balch, and M.~Veloso,
  ``Abides-gym: gym environments for multi-agent discrete event simulation and
  application to financial markets,'' in \emph{Proceedings of the Second ACM
  International Conference on AI in Finance}, 2021, pp. 1--9.

\bibitem{yang2022intelligent}
J.~Yang, J.~Ni, M.~Xi, J.~Wen, and Y.~Li, ``Intelligent path planning of
  underwater robot based on reinforcement learning,'' \emph{IEEE Transactions
  on Automation Science and Engineering}, vol.~20, no.~3, pp. 1983--1996, 2022.

\bibitem{palacios2023enhancing}
E.~Palacios-Morocho, S.~Inca, and J.~F. Monserrat, ``Enhancing cooperative
  multi-agent systems with self-advice and near-neighbor priority collision
  control,'' \emph{IEEE Transactions on Intelligent Vehicles}, 2023.

\bibitem{chen2023deep}
D.~Chen, M.~R. Hajidavalloo, Z.~Li, K.~Chen, Y.~Wang, L.~Jiang, and Y.~Wang,
  ``Deep multi-agent reinforcement learning for highway on-ramp merging in
  mixed traffic,'' \emph{IEEE Transactions on Intelligent Transportation
  Systems}, 2023.

\bibitem{chen2024data}
Z.~Chen, F.~Renda, A.~Le~Gall, L.~Mocellin, M.~Bernabei, T.~Dangel, G.~Ciuti,
  M.~Cianchetti, and C.~Stefanini, ``Data-driven methods applied to soft robot
  modeling and control: A review,'' \emph{IEEE Transactions on Automation
  Science and Engineering}, 2024.

\bibitem{chu2019multi}
T.~Chu, J.~Wang, L.~Codec{\`a}, and Z.~Li, ``Multi-agent deep reinforcement
  learning for large-scale traffic signal control,'' \emph{IEEE Transactions on
  Intelligent Transportation Systems}, vol.~21, no.~3, pp. 1086--1095, 2019.

\bibitem{zhou2020smarts}
M.~Zhou, J.~Luo, J.~Villella, Y.~Yang, D.~Rusu, J.~Miao, W.~Zhang, M.~Alban,
  I.~Fadakar, Z.~Chen \emph{et~al.}, ``Smarts: Scalable multi-agent
  reinforcement learning training school for autonomous driving,'' \emph{arXiv
  preprint arXiv:2010.09776}, 2020.

\bibitem{hernandez2019survey}
P.~Hernandez-Leal, B.~Kartal, and M.~E. Taylor, ``A survey and critique of
  multiagent deep reinforcement learning,'' \emph{Autonomous Agents and
  Multi-Agent Systems}, vol.~33, no.~6, pp. 750--797, 2019.

\bibitem{yadav2023comprehensive}
P.~Yadav, A.~Mishra, and S.~Kim, ``A comprehensive survey on multi-agent
  reinforcement learning for connected and automated vehicles,''
  \emph{Sensors}, vol.~23, no.~10, p. 4710, 2023.

\bibitem{kaelbling1996reinforcement}
L.~P. Kaelbling, M.~L. Littman, and A.~W. Moore, ``Reinforcement learning: A
  survey,'' \emph{Journal of artificial intelligence research}, vol.~4, pp.
  237--285, 1996.

\bibitem{mnih2015human}
V.~Mnih, K.~Kavukcuoglu, D.~Silver, A.~A. Rusu, J.~Veness, M.~G. Bellemare,
  A.~Graves, M.~Riedmiller, A.~K. Fidjeland, G.~Ostrovski \emph{et~al.},
  ``Human-level control through deep reinforcement learning,'' \emph{nature},
  vol. 518, no. 7540, pp. 529--533, 2015.

\bibitem{lillicrap2015continuous}
T.~P. Lillicrap, J.~J. Hunt, A.~Pritzel, N.~Heess, T.~Erez, Y.~Tassa,
  D.~Silver, and D.~Wierstra, ``Continuous control with deep reinforcement
  learning,'' \emph{arXiv preprint arXiv:1509.02971}, 2015.

\bibitem{mnih2016asynchronous}
V.~Mnih, A.~P. Badia, M.~Mirza, A.~Graves, T.~Lillicrap, T.~Harley, D.~Silver,
  and K.~Kavukcuoglu, ``Asynchronous methods for deep reinforcement learning,''
  in \emph{International conference on machine learning}, 2016, pp. 1928--1937.

\bibitem{vinyals2019grandmaster}
O.~Vinyals, I.~Babuschkin, W.~M. Czarnecki, M.~Mathieu, A.~Dudzik, J.~Chung,
  D.~H. Choi, R.~Powell, T.~Ewalds, P.~Georgiev \emph{et~al.}, ``Grandmaster
  level in starcraft ii using multi-agent reinforcement learning,''
  \emph{Nature}, vol. 575, no. 7782, pp. 350--354, 2019.

\bibitem{10388472}
B.~Zhang, X.~Lin, Y.~Zhu, J.~Tian, and Z.~Zhu, ``Enhancing multi-uav
  reconnaissance and search through double critic ddpg with belief probability
  maps,'' \emph{IEEE Transactions on Intelligent Vehicles}, pp. 1--16, 2024.

\bibitem{watkins1992q}
C.~J. Watkins and P.~Dayan, ``Q-learning,'' \emph{Machine learning}, vol.~8,
  pp. 279--292, 1992.

\bibitem{chu2019model}
T.~Chu and U.~Kalabi{\'c}, ``Model-based deep reinforcement learning for cacc
  in mixed-autonomy vehicle platoon,'' in \emph{2019 IEEE 58th Conference on
  Decision and Control (CDC)}.\hskip 1em plus 0.5em minus 0.4em\relax IEEE,
  2019, pp. 4079--4084.

\bibitem{szepesvari2022algorithms}
C.~Szepesv{\'a}ri, \emph{Algorithms for reinforcement learning}.\hskip 1em plus
  0.5em minus 0.4em\relax Springer Nature, 2022.

\bibitem{mnih2013playing}
V.~Mnih, K.~Kavukcuoglu, D.~Silver, A.~Graves, I.~Antonoglou, D.~Wierstra, and
  M.~Riedmiller, ``Playing atari with deep reinforcement learning,''
  \emph{arXiv preprint arXiv:1312.5602}, 2013.

\bibitem{andrychowicz2017hindsight}
M.~Andrychowicz, F.~Wolski, A.~Ray, J.~Schneider, R.~Fong, P.~Welinder,
  B.~McGrew, J.~Tobin, O.~Pieter~Abbeel, and W.~Zaremba, ``Hindsight experience
  replay,'' \emph{Advances in neural information processing systems}, vol.~30,
  2017.

\bibitem{haarnoja2018soft}
T.~Haarnoja, A.~Zhou, P.~Abbeel, and S.~Levine, ``Soft actor-critic: Off-policy
  maximum entropy deep reinforcement learning with a stochastic actor,'' in
  \emph{International conference on machine learning}.\hskip 1em plus 0.5em
  minus 0.4em\relax PMLR, 2018, pp. 1861--1870.

\bibitem{chu2020multiagent}
\BIBentryALTinterwordspacing
T.~Chu, S.~Chinchali, and S.~Katti, ``Multi-agent reinforcement learning for
  networked system control,'' in \emph{International Conference on Learning
  Representations}, 2020. [Online]. Available:
  \url{https://openreview.net/forum?id=Syx7A3NFvH}
\BIBentrySTDinterwordspacing

\bibitem{berner2019dota}
OpenAI, :, C.~Berner, G.~Brockman, B.~Chan, V.~Cheung, P.~Dębiak, C.~Dennison,
  D.~Farhi, Q.~Fischer, S.~Hashme, C.~Hesse, R.~Józefowicz, S.~Gray,
  C.~Olsson, J.~Pachocki, M.~Petrov, H.~P. d.~O.~Pinto, J.~Raiman, T.~Salimans,
  J.~Schlatter, J.~Schneider, S.~Sidor, I.~Sutskever, J.~Tang, F.~Wolski, and
  S.~Zhang, ``Dota 2 with large scale deep reinforcement learning,'' 2019.

\bibitem{naderializadeh2021resource}
N.~Naderializadeh, J.~Sydir, M.~Simsek, and H.~Nikopour, ``Resource management
  in wireless networks via multi-agent deep reinforcement learning,''
  \emph{IEEE Transactions on Wireless Communications}, 2021.

\bibitem{chen2021powernet}
D.~Chen, K.~Chen, Z.~Li, T.~Chu, R.~Yao, F.~Qiu, and K.~Lin, ``Powernet:
  Multi-agent deep reinforcement learning for scalable powergrid control,''
  \emph{IEEE Transactions on Power Systems}, vol.~37, no.~2, pp. 1007--1017,
  2021.

\bibitem{zhang2021multi}
K.~Zhang, Z.~Yang, and T.~Ba{\c{s}}ar, ``Multi-agent reinforcement learning: A
  selective overview of theories and algorithms,'' \emph{Handbook of
  reinforcement learning and control}, pp. 321--384, 2021.

\bibitem{wang2023hierarchical}
J.~Wang, M.~Yuan, Y.~Li, and Z.~Zhao, ``Hierarchical attention master--slave
  for heterogeneous multi-agent reinforcement learning,'' \emph{Neural
  Networks}, pp. 359--368, 2023.

\bibitem{MahajanRSW19}
A.~Mahajan, T.~Rashid, M.~Samvelyan, and S.~Whiteson, ``{MAVEN:} multi-agent
  variational exploration,'' in \emph{Advances in Neural Information Processing
  Systems}, 2019, pp. 7611--7622.

\bibitem{li2023multi}
W.~Li, S.~He, X.~Mao, B.~Li, C.~Qiu, J.~Yu, F.~Peng, and X.~Tan, ``Multi-agent
  evolution reinforcement learning method for machining parameters optimization
  based on bootstrap aggregating graph attention network simulated
  environment,'' \emph{Journal of Manufacturing Systems}, pp. 424--438, 2023.

\bibitem{9931995}
D.~Qiu, J.~Wang, Z.~Dong, Y.~Wang, and G.~Strbac, ``Mean-field multi-agent
  reinforcement learning for peer-to-peer multi-energy trading,'' \emph{IEEE
  Transactions on Power Systems}, pp. 1--13, 2022.

\bibitem{tan1993multi}
M.~Tan, ``Multi-agent reinforcement learning: Independent vs. cooperative
  agents,'' in \emph{Proceedings of the tenth international conference on
  machine learning}, 1993, pp. 330--337.

\bibitem{nguyen2020deep}
T.~T. Nguyen, N.~D. Nguyen, and S.~Nahavandi, ``Deep reinforcement learning for
  multiagent systems: A review of challenges, solutions, and applications,''
  \emph{IEEE transactions on cybernetics}, vol.~50, no.~9, pp. 3826--3839,
  2020.

\bibitem{zhang2018fully}
K.~Zhang, Z.~Yang, H.~Liu, T.~Zhang, and T.~Basar, ``Fully decentralized
  multi-agent reinforcement learning with networked agents,'' in
  \emph{International Conference on Machine Learning}.\hskip 1em plus 0.5em
  minus 0.4em\relax PMLR, 2018, pp. 5872--5881.

\bibitem{SunehagLGCZJLSL18}
P.~Sunehag, G.~Lever, A.~Gruslys, W.~M. Czarnecki, V.~F. Zambaldi,
  M.~Jaderberg, M.~Lanctot, N.~Sonnerat, J.~Z. Leibo, K.~Tuyls, and T.~Graepel,
  ``Value-decomposition networks for cooperative multi-agent learning based on
  team reward,'' \emph{Proceedings of the 17th International Conference on
  Autonomous Agents and MultiAgent Systems}, pp. 2085--2087, 2018.

\bibitem{pmlr-v80-rashid18a}
T.~Rashid, M.~Samvelyan, C.~Schroeder, G.~Farquhar, J.~Foerster, and
  S.~Whiteson, ``{QMIX}: Monotonic value function factorisation for deep
  multi-agent reinforcement learning,'' \emph{Proceedings of the 35th
  International Conference on Machine Learning}, pp. 4295--4304, 2018.

\bibitem{pmlr-v97-son19a}
K.~Son, D.~Kim, W.~J. Kang, D.~E. Hostallero, and Y.~Yi, ``Qtran: Learning to
  factorize with transformation for cooperative multi-agent reinforcement
  learning,'' \emph{International Conference on Machine Learning}, pp.
  5887--5896, 2019.

\bibitem{RashidFPW20}
T.~Rashid, G.~Farquhar, B.~Peng, and S.~Whiteson, ``Weighted qmix: Expanding
  monotonic value function factorisation for deep multi-agent reinforcement
  learning,'' \emph{Advances in neural information processing systems}, pp.
  10\,199--10\,210, 2020.

\bibitem{zhang2021avd}
Y.~Zhang, H.~Ma, and Y.~Wang, ``Avd-net: Attention value decomposition network
  for deep multi-agent reinforcement learning,'' \emph{25th International
  Conference on Pattern Recognition}, pp. 7810--7816, 2021.

\bibitem{10103926}
S.~Liu, W.~Liu, W.~Chen, G.~Tian, J.~Chen, Y.~Tong, J.~Cao, and Y.~Liu,
  ``Learning multi-agent cooperation via considering actions of teammates,''
  \emph{IEEE Transactions on Neural Networks and Learning Systems}, pp. 1--12,
  2023.

\bibitem{foerster2017stabilising}
J.~Foerster, N.~Nardelli, G.~Farquhar, T.~Afouras, P.~H. Torr, P.~Kohli, and
  S.~Whiteson, ``Stabilising experience replay for deep multi-agent
  reinforcement learning,'' in \emph{International conference on machine
  learning}.\hskip 1em plus 0.5em minus 0.4em\relax PMLR, 2017, pp. 1146--1155.

\bibitem{foerster2016learning}
J.~Foerster, I.~A. Assael, N.~De~Freitas, and S.~Whiteson, ``Learning to
  communicate with deep multi-agent reinforcement learning,'' \emph{Advances in
  neural information processing systems}, vol.~29, 2016.

\bibitem{wang2020}
R.~Wang, X.~He, R.~Yu, W.~Qiu, B.~An, and Z.~Rabinovich, ``Learning efficient
  multi-agent communication: An information bottleneck approach,'' in
  \emph{International Conference on Machine Learning}.\hskip 1em plus 0.5em
  minus 0.4em\relax PMLR, 2020, pp. 9908--9918.

\bibitem{ahilan2019feudal}
S.~Ahilan and P.~Dayan, ``Feudal multi-agent hierarchies for cooperative
  reinforcement learning,'' \emph{arXiv preprint arXiv:1901.08492}, 2019.

\bibitem{tang2018hierarchical}
H.~Tang, J.~Hao, T.~Lv, Y.~Chen, Z.~Zhang, H.~Jia, C.~Ren, Y.~Zheng, Z.~Meng,
  C.~Fan \emph{et~al.}, ``Hierarchical deep multiagent reinforcement learning
  with temporal abstraction,'' \emph{arXiv preprint arXiv:1809.09332}, 2018.

\bibitem{yang2019hierarchical}
J.~Yang, I.~Borovikov, and H.~Zha, ``Hierarchical cooperative multi-agent
  reinforcement learning with skill discovery,'' \emph{arXiv preprint
  arXiv:1912.03558}, 2019.

\bibitem{wang2020rode}
T.~Wang, T.~Gupta, A.~Mahajan, B.~Peng, S.~Whiteson, and C.~Zhang, ``Rode:
  Learning roles to decompose multi-agent tasks,'' \emph{arXiv preprint
  arXiv:2010.01523}, 2020.

\bibitem{xu2023haven}
Z.~Xu, Y.~Bai, B.~Zhang, D.~Li, and G.~Fan, ``Haven: hierarchical cooperative
  multi-agent reinforcement learning with dual coordination mechanism,'' in
  \emph{Proceedings of the AAAI Conference on Artificial Intelligence},
  vol.~37, no.~10, 2023, pp. 11\,735--11\,743.

\bibitem{jaques2018intrinsic}
\BIBentryALTinterwordspacing
N.~Jaques, A.~Lazaridou, E.~Hughes, C.~Gulcehre, P.~A. Ortega, D.~Strouse,
  J.~Z. Leibo, and N.~de~Freitas, ``Intrinsic social motivation via causal
  influence in multi-agent rl,'' 2018. [Online]. Available:
  \url{https://openreview.net/forum?id=B1lG42C9Km}
\BIBentrySTDinterwordspacing

\bibitem{pina2023discovering}
R.~Pina, V.~De~Silva, and C.~Artaud, ``Discovering causality for efficient
  cooperation in multi-agent environments,'' \emph{arXiv preprint
  arXiv:2306.11846}, 2023.

\bibitem{liu2023lazy}
B.~Liu, Z.~Pu, Y.~Pan, J.~Yi, Y.~Liang, and D.~Zhang, ``Lazy agents: a new
  perspective on solving sparse reward problem in multi-agent reinforcement
  learning,'' in \emph{International Conference on Machine Learning}.\hskip 1em
  plus 0.5em minus 0.4em\relax PMLR, 2023, pp. 21\,937--21\,950.

\bibitem{wang2022fully}
H.~Wang, Y.~Yu, and Y.~Jiang, ``Fully decentralized multiagent communication
  via causal inference,'' \emph{IEEE Transactions on Neural Networks and
  Learning Systems}, 2022.

\bibitem{li2022deconfounded}
J.~Li, K.~Kuang, B.~Wang, F.~Liu, L.~Chen, C.~Fan, F.~Wu, and J.~Xiao,
  ``Deconfounded value decomposition for multi-agent reinforcement learning,''
  in \emph{International Conference on Machine Learning}.\hskip 1em plus 0.5em
  minus 0.4em\relax PMLR, 2022, pp. 12\,843--12\,856.

\bibitem{jiang2023credit}
K.~Jiang, W.~Liu, Y.~Wang, L.~Dong, and C.~Sun, ``Credit assignment in
  heterogeneous multi-agent reinforcement learning for fully cooperative
  tasks,'' \emph{Applied Intelligence}, pp. 1--18, 2023.

\bibitem{counterfactual}
J.~Foerster, G.~Farquhar, T.~Afouras, N.~Nardelli, and S.~Whiteson,
  ``Counterfactual multi-agent policy gradients,'' \emph{Proceedings of the
  AAAI Conference on Artificial Intelligence}, pp. 2974--2982, 2018.

\bibitem{9185035}
D.~Guo, L.~Tang, X.~Zhang, and Y.-C. Liang, ``Joint optimization of handover
  control and power allocation based on multi-agent deep reinforcement
  learning,'' \emph{IEEE Transactions on Vehicular Technology}, pp.
  13\,124--13\,138, 2020.

\bibitem{10104101}
Y.~Hou, M.~Sun, Y.~Zeng, Y.-S. Ong, Y.~Jin, H.~Ge, and Q.~Zhang, ``A
  multi-agent cooperative learning system with evolution of social roles,''
  \emph{IEEE Transactions on Evolutionary Computation}, 2023.

\bibitem{lowe2017multi}
R.~Lowe, Y.~I. Wu, A.~Tamar, J.~Harb, O.~Pieter~Abbeel, and I.~Mordatch,
  ``Multi-agent actor-critic for mixed cooperative-competitive environments,''
  \emph{Advances in neural information processing systems}, vol.~30, 2017.

\bibitem{foerster2018counterfactual}
J.~Foerster, G.~Farquhar, T.~Afouras, N.~Nardelli, and S.~Whiteson,
  ``Counterfactual multi-agent policy gradients,'' in \emph{Proceedings of the
  AAAI conference on artificial intelligence}, vol.~32, no.~1, 2018.

\bibitem{bacon2017option}
P.-L. Bacon, J.~Harb, and D.~Precup, ``The option-critic architecture,'' in
  \emph{Proceedings of the AAAI conference on artificial intelligence},
  vol.~31, no.~1, 2017.

\bibitem{harb2018waiting}
J.~Harb, P.-L. Bacon, M.~Klissarov, and D.~Precup, ``When waiting is not an
  option: Learning options with a deliberation cost,'' in \emph{Proceedings of
  the AAAI Conference on Artificial Intelligence}, vol.~32, no.~1, 2018.

\bibitem{vezhnevets2017feudal}
A.~S. Vezhnevets, S.~Osindero, T.~Schaul, N.~Heess, M.~Jaderberg, D.~Silver,
  and K.~Kavukcuoglu, ``Feudal networks for hierarchical reinforcement
  learning,'' in \emph{International Conference on Machine Learning}.\hskip 1em
  plus 0.5em minus 0.4em\relax PMLR, 2017, pp. 3540--3549.

\bibitem{nachum2018data}
O.~Nachum, S.~S. Gu, H.~Lee, and S.~Levine, ``Data-efficient hierarchical
  reinforcement learning,'' \emph{Advances in neural information processing
  systems}, vol.~31, 2018.

\bibitem{grimbly2021causal}
S.~J. Grimbly, J.~Shock, and A.~Pretorius, ``Causal multi-agent reinforcement
  learning: Review and open problems,'' \emph{arXiv preprint arXiv:2111.06721},
  2021.

\bibitem{pearl2018theoretical}
J.~Pearl, ``Theoretical impediments to machine learning with seven sparks from
  the causal revolution,'' \emph{arXiv preprint arXiv:1801.04016}, 2018.

\bibitem{yang2023causal}
S.~Yang, B.~Yang, Z.~Zeng, and Z.~Kang, ``Causal inference multi-agent
  reinforcement learning for traffic signal control,'' \emph{Information
  Fusion}, vol.~94, pp. 243--256, 2023.

\bibitem{ho2021human}
J.~Ho and C.-M. Wang, ``Human-centered ai using ethical causality and learning
  representation for multi-agent deep reinforcement learning,'' in \emph{2021
  IEEE 2nd International Conference on Human-Machine Systems (ICHMS)}.\hskip
  1em plus 0.5em minus 0.4em\relax IEEE, 2021, pp. 1--6.

\bibitem{peake2020multi}
A.~Peake, J.~McCalmon, B.~Raiford, T.~Liu, and S.~Alqahtani, ``Multi-agent
  reinforcement learning for cooperative adaptive cruise control,'' in
  \emph{2020 IEEE 32nd International Conference on Tools with Artificial
  Intelligence (ICTAI)}.\hskip 1em plus 0.5em minus 0.4em\relax IEEE, 2020, pp.
  15--22.

\bibitem{vu2020multi}
H.~V. Vu, M.~Farzanullah, Z.~Liu, D.~H. Nguyen, R.~Morawski, and T.~Le-Ngoc,
  ``Multi-agent reinforcement learning for channel assignment and power
  allocation in platoon-based c-v2x systems,'' \emph{arXiv preprint
  arXiv:2011.04555}, 2020.

\bibitem{liu2021efficient}
B.~Liu, W.~Han, E.~Wang, X.~Ma, S.~Xiong, C.~Qiao, and J.~Wang, ``An efficient
  message dissemination scheme for cooperative drivings via multi-agent
  hierarchical attention reinforcement learning,'' in \emph{2021 IEEE 41st
  International Conference on Distributed Computing Systems (ICDCS)}.\hskip 1em
  plus 0.5em minus 0.4em\relax IEEE, 2021, pp. 326--336.

\bibitem{li2022deep}
B.~Li, K.~Xie, X.~Huang, Y.~Wu, and S.~Xie, ``Deep reinforcement learning based
  incentive mechanism design for platoon autonomous driving with social
  effect,'' \emph{IEEE Transactions on Vehicular Technology}, vol.~71, no.~7,
  pp. 7719--7729, 2022.

\bibitem{shi2023deep}
H.~Shi, D.~Chen, N.~Zheng, X.~Wang, Y.~Zhou, and B.~Ran, ``A deep reinforcement
  learning based distributed control strategy for connected automated vehicles
  in mixed traffic platoon,'' \emph{Transportation Research Part C: Emerging
  Technologies}, vol. 148, p. 104019, 2023.

\bibitem{hou2021decentralized}
Y.~Hou and P.~Graf, ``Decentralized cooperative lane changing at freeway
  weaving areas using multi-agent deep reinforcement learning,'' \emph{arXiv
  preprint arXiv:2110.08124}, 2021.

\bibitem{wang2021harmonious}
G.~Wang, J.~Hu, Z.~Li, and L.~Li, ``Harmonious lane changing via deep
  reinforcement learning,'' \emph{IEEE Transactions on Intelligent
  Transportation Systems}, vol.~23, no.~5, pp. 4642--4650, 2021.

\bibitem{nagarajan2021lane}
K.~Nagarajan and Z.~Yi, ``Lane changing using multi-agent dqn,'' in \emph{2021
  IEEE International Conference on Autonomous Systems (ICAS)}.\hskip 1em plus
  0.5em minus 0.4em\relax IEEE, 2021, pp. 1--6.

\bibitem{chen2022multi}
S.~Chen, M.~Wang, W.~Song, Y.~Yang, and M.~Fu, ``Multi-agent reinforcement
  learning-based twin-vehicle fair cooperative driving in dynamic highway
  scenarios,'' in \emph{2022 IEEE 25th International Conference on Intelligent
  Transportation Systems (ITSC)}.\hskip 1em plus 0.5em minus 0.4em\relax IEEE,
  2022, pp. 730--736.

\bibitem{zhang2022multi}
J.~Zhang, C.~Chang, X.~Zeng, and L.~Li, ``Multi-agent drl-based lane change
  with right-of-way collaboration awareness,'' \emph{IEEE Transactions on
  Intelligent Transportation Systems}, vol.~24, no.~1, pp. 854--869, 2022.

\bibitem{zhou2022multi}
W.~Zhou, D.~Chen, J.~Yan, Z.~Li, H.~Yin, and W.~Ge, ``Multi-agent reinforcement
  learning for cooperative lane changing of connected and autonomous vehicles
  in mixed traffic,'' \emph{Autonomous Intelligent Systems}, vol.~2, no.~1,
  p.~5, 2022.

\bibitem{wang2023faster}
H.~Wang, W.~Hao, J.~So, X.~Xiao, Z.~Chen, and J.~Hu, ``A faster cooperative
  lane change controller enabled by formulating in spatial domain,'' \emph{IEEE
  Transactions on Intelligent Vehicles}, 2023.

\bibitem{tsugawa2016review}
S.~Tsugawa, S.~Jeschke, and S.~E. Shladover, ``A review of truck platooning
  projects for energy savings,'' \emph{IEEE Transactions on Intelligent
  Vehicles}, vol.~1, no.~1, pp. 68--77, 2016.

\bibitem{liu2020platoon}
B.~Liu, Z.~Ding, and C.~Lv, ``Platoon control of connected autonomous vehicles:
  A distributed reinforcement learning method by consensus,''
  \emph{IFAC-PapersOnLine}, vol.~53, no.~2, pp. 15\,241--15\,246, 2020.

\bibitem{feng2019string}
S.~Feng, Y.~Zhang, S.~E. Li, Z.~Cao, H.~X. Liu, and L.~Li, ``String stability
  for vehicular platoon control: Definitions and analysis methods,''
  \emph{Annual Reviews in Control}, vol.~47, pp. 81--97, 2019.

\bibitem{hochreiter1997long}
S.~Hochreiter and J.~Schmidhuber, ``Long short-term memory,'' \emph{Neural
  computation}, vol.~9, no.~8, pp. 1735--1780, 1997.

\bibitem{li2021reinforcement}
M.~Li, Z.~Cao, and Z.~Li, ``A reinforcement learning-based vehicle platoon
  control strategy for reducing energy consumption in traffic oscillations,''
  \emph{IEEE Transactions on Neural Networks and Learning Systems}, vol.~32,
  no.~12, pp. 5309--5322, 2021.

\bibitem{xu2023deep}
Y.~Xu, K.~Zhu, H.~Xu, and J.~Ji, ``Deep reinforcement learning for
  multi-objective resource allocation in multi-platoon cooperative vehicular
  networks,'' \emph{IEEE Transactions on Wireless Communications}, 2023.

\bibitem{parvini2023aoi}
M.~Parvini, M.~R. Javan, N.~Mokari, B.~Abbasi, and E.~A. Jorswieck, ``Aoi-aware
  resource allocation for platoon-based c-v2x networks via multi-agent
  multi-task reinforcement learning,'' \emph{IEEE Transactions on Vehicular
  Technology}, 2023.

\bibitem{wang2019cooperative}
Z.~Wang, G.~Wu, and M.~J. Barth, ``Cooperative eco-driving at signalized
  intersections in a partially connected and automated vehicle environment,''
  \emph{IEEE Transactions on Intelligent Transportation Systems}, vol.~21,
  no.~5, pp. 2029--2038, 2019.

\bibitem{lu2023altruistic}
S.~Lu, Y.~Cai, L.~Chen, H.~Wang, X.~Sun, and H.~Gao, ``Altruistic cooperative
  adaptive cruise control of mixed traffic platoon based on deep reinforcement
  learning,'' \emph{IET Intelligent Transport Systems}, 2023.

\bibitem{he2022robust}
X.~He, H.~Yang, Z.~Hu, and C.~Lv, ``Robust lane change decision making for
  autonomous vehicles: An observation adversarial reinforcement learning
  approach,'' \emph{IEEE Transactions on Intelligent Vehicles}, vol.~8, no.~1,
  pp. 184--193, 2022.

\bibitem{chen2024lane}
G.~Chen, Z.~Gao, M.~Hua, B.~Shuai, and Z.~Gao, ``Lane change trajectory
  prediction considering driving style uncertainty for autonomous vehicles,''
  \emph{Mechanical Systems and Signal Processing}, vol. 206, p. 110854, 2024.

\bibitem{chen2023multi}
S.~Chen, M.~Wang, W.~Song, Y.~Yang, and M.~Fu, ``Multi-agent reinforcement
  learning-based decision making for twin-vehicles cooperative driving in
  stochastic dynamic highway environments,'' \emph{IEEE Transactions on
  Vehicular Technology}, 2023.

\bibitem{vishnu2023improving}
C.~Vishnu, V.~Abhinav, D.~Roy, C.~K. Mohan, and C.~S. Babu, ``Improving
  multi-agent trajectory prediction using traffic states on interactive driving
  scenarios,'' \emph{IEEE Robotics and Automation Letters}, vol.~8, no.~5, pp.
  2708--2715, 2023.

\bibitem{zhou2020graph}
J.~Zhou, G.~Cui, S.~Hu, Z.~Zhang, C.~Yang, Z.~Liu, L.~Wang, C.~Li, and M.~Sun,
  ``Graph neural networks: A review of methods and applications,'' \emph{AI
  open}, vol.~1, pp. 57--81, 2020.

\bibitem{chen2021graph}
S.~Chen, J.~Dong, P.~Ha, Y.~Li, and S.~Labi, ``Graph neural network and
  reinforcement learning for multi-agent cooperative control of connected
  autonomous vehicles,'' \emph{Computer-Aided Civil and Infrastructure
  Engineering}, vol.~36, no.~7, pp. 838--857, 2021.

\bibitem{ha2020leveraging}
P.~Y.~J. Ha, S.~Chen, J.~Dong, R.~Du, Y.~Li, and S.~Labi, ``Leveraging the
  capabilities of connected and autonomous vehicles and multi-agent
  reinforcement learning to mitigate highway bottleneck congestion,''
  \emph{arXiv preprint arXiv:2010.05436}, 2020.

\bibitem{han2020multi}
S.~Han, S.~Zhou, J.~Wang, L.~Pepin, C.~Ding, J.~Fu, and F.~Miao, ``A
  multi-agent reinforcement learning approach for safe and efficient behavior
  planning of connected autonomous vehicles,'' \emph{arXiv preprint
  arXiv:2003.04371}, 2020.

\bibitem{9829243}
Q.~Li, Z.~Peng, L.~Feng, Q.~Zhang, Z.~Xue, and B.~Zhou, ``Metadrive: Composing
  diverse driving scenarios for generalizable reinforcement learning,''
  \emph{IEEE Transactions on Pattern Analysis and Machine Intelligence},
  vol.~45, no.~3, pp. 3461--3475, 2023.

\bibitem{wu2019dcl}
Y.~Wu, H.~Chen, and F.~Zhu, ``Dcl-aim: Decentralized coordination learning of
  autonomous intersection management for connected and automated vehicles,''
  \emph{Transportation Research Part C: Emerging Technologies}, vol. 103, pp.
  246--260, 2019.

\bibitem{tan2019cooperative}
T.~Tan, F.~Bao, Y.~Deng, A.~Jin, Q.~Dai, and J.~Wang, ``Cooperative deep
  reinforcement learning for large-scale traffic grid signal control,''
  \emph{IEEE transactions on cybernetics}, vol.~50, no.~6, pp. 2687--2700,
  2019.

\bibitem{van2016coordinated}
E.~Van~der Pol and F.~A. Oliehoek, ``Coordinated deep reinforcement learners
  for traffic light control,'' \emph{Proceedings of learning, inference and
  control of multi-agent systems (at NIPS 2016)}, vol.~8, pp. 21--38, 2016.

\bibitem{liu2021learning}
J.~Liu, H.~Zhang, Z.~Fu, and Y.~Wang, ``Learning scalable multi-agent
  coordination by spatial differentiation for traffic signal control,''
  \emph{Engineering Applications of Artificial Intelligence}, vol. 100, p.
  104165, 2021.

\bibitem{wang2021adaptive}
T.~Wang, J.~Cao, and A.~Hussain, ``Adaptive traffic signal control for
  large-scale scenario with cooperative group-based multi-agent reinforcement
  learning,'' \emph{Transportation research part C: emerging technologies},
  vol. 125, p. 103046, 2021.

\bibitem{wu2020multi}
T.~Wu, P.~Zhou, K.~Liu, Y.~Yuan, X.~Wang, H.~Huang, and D.~O. Wu, ``Multi-agent
  deep reinforcement learning for urban traffic light control in vehicular
  networks,'' \emph{IEEE Transactions on Vehicular Technology}, vol.~69, no.~8,
  pp. 8243--8256, 2020.

\bibitem{ma2020feudal}
J.~Ma and F.~Wu, ``Feudal multi-agent deep reinforcement learning for traffic
  signal control,'' in \emph{Proceedings of the 19th International Conference
  on Autonomous Agents and Multiagent Systems (AAMAS)}, 2020, pp. 816--824.

\bibitem{zhu2022survey}
C.~Zhu, M.~Dastani, and S.~Wang, ``A survey of multi-agent reinforcement
  learning with communication,'' \emph{arXiv preprint arXiv:2203.08975}, 2022.

\bibitem{liu2023traffic}
D.~Liu and L.~Li, ``A traffic light control method based on multi-agent deep
  reinforcement learning algorithm,'' \emph{Scientific Reports}, vol.~13,
  no.~1, p. 9396, 2023.

\bibitem{liu2023graph}
Q.~Liu, X.~Li, Y.~Tang, X.~Gao, F.~Yang, and Z.~Li, ``Graph reinforcement
  learning-based decision-making technology for connected and autonomous
  vehicles: Framework, review, and future trends,'' \emph{Sensors}, vol.~23,
  no.~19, p. 8229, 2023.

\bibitem{wei2019colight}
H.~Wei, N.~Xu, H.~Zhang, G.~Zheng, X.~Zang, C.~Chen, W.~Zhang, Y.~Zhu, K.~Xu,
  and Z.~Li, ``Colight: Learning network-level cooperation for traffic signal
  control,'' in \emph{Proceedings of the 28th ACM International Conference on
  Information and Knowledge Management}, 2019, pp. 1913--1922.

\bibitem{yang2021ihg}
S.~Yang, B.~Yang, Z.~Kang, and L.~Deng, ``Ihg-ma: Inductive heterogeneous graph
  multi-agent reinforcement learning for multi-intersection traffic signal
  control,'' \emph{Neural networks}, vol. 139, pp. 265--277, 2021.

\bibitem{wang2020stmarl}
Y.~Wang, T.~Xu, X.~Niu, C.~Tan, E.~Chen, and H.~Xiong, ``Stmarl: A
  spatio-temporal multi-agent reinforcement learning approach for cooperative
  traffic light control,'' \emph{IEEE Transactions on Mobile Computing},
  vol.~21, no.~6, pp. 2228--2242, 2020.

\bibitem{jiang2022multi}
Q.~Jiang, M.~Qin, S.~Shi, W.~Sun, and B.~Zheng, ``Multi-agent reinforcement
  learning for traffic signal control through universal communication method,''
  \emph{arXiv preprint arXiv:2204.12190}, 2022.

\bibitem{9762548}
G.-P. Antonio and C.~Maria-Dolores, ``Multi-agent deep reinforcement learning
  to manage connected autonomous vehicles at tomorrow's intersections,''
  \emph{IEEE Transactions on Vehicular Technology}, vol.~71, no.~7, pp.
  7033--7043, 2022.

\bibitem{yan2023multi}
Y.~Yan, L.~Peng, T.~Shen, J.~Wang, D.~Pi, D.~Cao, and G.~Yin, ``A multi-vehicle
  game-theoretic framework for decision making and planning of autonomous
  vehicles in mixed traffic,'' \emph{IEEE Transactions on Intelligent
  Vehicles}, 2023.

\bibitem{schester2019longitudinal}
L.~Schester and L.~E. Ortiz, ``Longitudinal position control for highway
  on-ramp merging: A multi-agent approach to automated driving,'' in \emph{2019
  IEEE Intelligent Transportation Systems Conference (ITSC)}.\hskip 1em plus
  0.5em minus 0.4em\relax IEEE, 2019, pp. 3461--3468.

\bibitem{zhou2022cooperative}
S.~Zhou, W.~Zhuang, G.~Yin, H.~Liu, and C.~Qiu, ``Cooperative on-ramp merging
  control of connected and automated vehicles: Distributed multi-agent deep
  reinforcement learning approach,'' in \emph{2022 IEEE 25th International
  Conference on Intelligent Transportation Systems (ITSC)}.\hskip 1em plus
  0.5em minus 0.4em\relax IEEE, 2022, pp. 402--408.

\bibitem{nakka2022multi}
S.~K.~S. Nakka, B.~Chalaki, and A.~A. Malikopoulos, ``A multi-agent deep
  reinforcement learning coordination framework for connected and automated
  vehicles at merging roadways,'' in \emph{2022 American Control Conference
  (ACC)}.\hskip 1em plus 0.5em minus 0.4em\relax IEEE, 2022, pp. 3297--3302.

\bibitem{hu2019interaction}
Y.~Hu, A.~Nakhaei, M.~Tomizuka, and K.~Fujimura, ``Interaction-aware decision
  making with adaptive strategies under merging scenarios,'' in \emph{2019
  IEEE/RSJ International Conference on Intelligent Robots and Systems
  (IROS)}.\hskip 1em plus 0.5em minus 0.4em\relax IEEE, 2019, pp. 151--158.

\bibitem{xu2021leveraging}
Y.~Xu, H.~Zhou, T.~Ma, J.~Zhao, B.~Qian, and X.~Shen, ``Leveraging multiagent
  learning for automated vehicles scheduling at nonsignalized intersections,''
  \emph{IEEE Internet of Things Journal}, vol.~8, no.~14, pp. 11\,427--11\,439,
  2021.

\bibitem{spatharis2022multiagent}
C.~Spatharis and K.~Blekas, ``Multiagent reinforcement learning for autonomous
  driving in traffic zones with unsignalized intersections,'' \emph{Journal of
  Intelligent Transportation Systems}, pp. 1--17, 2022.

\bibitem{antonio2022multi}
G.-P. Antonio and C.~Maria-Dolores, ``Multi-agent deep reinforcement learning
  to manage connected autonomous vehicles at tomorrow's intersections,''
  \emph{IEEE Transactions on Vehicular Technology}, vol.~71, no.~7, pp.
  7033--7043, 2022.

\bibitem{guo2022coordination}
Z.~Guo, Y.~Wu, L.~Wang, and J.~Zhang, ``Coordination for connected and
  automated vehicles at non-signalized intersections: A value
  decomposition-based multiagent deep reinforcement learning approach,''
  \emph{IEEE Transactions on Vehicular Technology}, vol.~72, no.~3, pp.
  3025--3034, 2022.

\bibitem{aksjonov2021rule}
A.~Aksjonov and V.~Kyrki, ``Rule-based decision-making system for autonomous
  vehicles at intersections with mixed traffic environment,'' in \emph{2021
  IEEE International Intelligent Transportation Systems Conference
  (ITSC)}.\hskip 1em plus 0.5em minus 0.4em\relax IEEE, 2021, pp. 660--666.

\bibitem{gu2016motion}
Y.~Gu, Y.~Hashimoto, L.-T. Hsu, and S.~Kamijo, ``Motion planning based on
  learning models of pedestrian and driver behaviors,'' in \emph{2016 IEEE 19th
  International Conference on Intelligent Transportation Systems (ITSC)}.\hskip
  1em plus 0.5em minus 0.4em\relax IEEE, 2016, pp. 808--813.

\bibitem{10186792}
C.~Huang, J.~Zhao, H.~Zhou, H.~Zhang, X.~Zhang, and C.~Ye, ``Multi-agent
  decision-making at unsignalized intersections with reinforcement learning
  from demonstrations,'' in \emph{2023 IEEE Intelligent Vehicles Symposium
  (IV)}, 2023, pp. 1--6.

\bibitem{capasso2021end}
A.~P. Capasso, P.~Maramotti, A.~Dell'Eva, and A.~Broggi, ``End-to-end
  intersection handling using multi-agent deep reinforcement learning,'' in
  \emph{2021 IEEE Intelligent Vehicles Symposium (IV)}.\hskip 1em plus 0.5em
  minus 0.4em\relax IEEE, 2021, pp. 443--450.

\bibitem{yan2021reinforcement}
Z.~Yan and C.~Wu, ``Reinforcement learning for mixed autonomy intersections,''
  in \emph{2021 IEEE International Intelligent Transportation Systems
  Conference (ITSC)}.\hskip 1em plus 0.5em minus 0.4em\relax IEEE, 2021, pp.
  2089--2094.

\bibitem{mavrogiannis2020implicit}
C.~Mavrogiannis, J.~A. DeCastro, and S.~S. Srinivasa, ``Implicit multiagent
  coordination at unsignalized intersections via multimodal inference enabled
  by topological braids,'' \emph{arXiv preprint arXiv:2004.05205}, 2020.

\bibitem{zheng2022deep}
J.~Zheng, K.~Zhu, and R.~Wang, ``Deep reinforcement learning for autonomous
  vehicles collaboration at unsignalized intersections,'' in \emph{GLOBECOM
  2022-2022 IEEE Global Communications Conference}.\hskip 1em plus 0.5em minus
  0.4em\relax IEEE, 2022, pp. 1115--1120.

\bibitem{hamouda2021multi}
A.~H. Hamouda, D.~M. Mahfouz, C.~M. Elias, and O.~M. Shehata, ``Multi-layer
  control architecture for unsignalized intersection management via nonlinear
  mpc and deep reinforcement learning,'' in \emph{2021 IEEE International
  Intelligent Transportation Systems Conference (ITSC)}.\hskip 1em plus 0.5em
  minus 0.4em\relax IEEE, 2021, pp. 1990--1996.

\bibitem{tallapragada2023reinforcement}
P.~Tallapragada \emph{et~al.}, ``Reinforcement learning aided sequential
  optimization for unsignalized intersection management of robot traffic,''
  \emph{arXiv preprint arXiv:2302.05082}, 2023.

\bibitem{li2021learning}
Z.~Li, Q.~Yuan, G.~Luo, and J.~Li, ``Learning effective multi-vehicle
  cooperation at unsignalized intersection via bandwidth-constrained
  communication,'' in \emph{2021 IEEE 94th Vehicular Technology Conference
  (VTC2021-Fall)}.\hskip 1em plus 0.5em minus 0.4em\relax IEEE, 2021, pp. 1--7.

\bibitem{lopez2018microscopic}
P.~A. Lopez, M.~Behrisch, L.~Bieker-Walz, J.~Erdmann, Y.-P. Fl{\"o}tter{\"o}d,
  R.~Hilbrich, L.~L{\"u}cken, J.~Rummel, P.~Wagner, and E.~Wie{\ss}ner,
  ``Microscopic traffic simulation using sumo,'' in \emph{2018 21st
  international conference on intelligent transportation systems (ITSC)}.\hskip
  1em plus 0.5em minus 0.4em\relax IEEE, 2018, pp. 2575--2582.

\bibitem{zhang2019cityflow}
H.~Zhang, S.~Feng, C.~Liu, Y.~Ding, Y.~Zhu, Z.~Zhou, W.~Zhang, Y.~Yu, H.~Jin,
  and Z.~Li, ``Cityflow: A multi-agent reinforcement learning environment for
  large scale city traffic scenario,'' in \emph{The world wide web conference},
  2019, pp. 3620--3624.

\bibitem{li2022metadrive}
Q.~Li, Z.~Peng, L.~Feng, Q.~Zhang, Z.~Xue, and B.~Zhou, ``Metadrive: Composing
  diverse driving scenarios for generalizable reinforcement learning,''
  \emph{IEEE transactions on pattern analysis and machine intelligence},
  vol.~45, no.~3, pp. 3461--3475, 2022.

\bibitem{dosovitskiy2017carla}
A.~Dosovitskiy, G.~Ros, F.~Codevilla, A.~Lopez, and V.~Koltun, ``Carla: An open
  urban driving simulator,'' in \emph{Conference on robot learning}.\hskip 1em
  plus 0.5em minus 0.4em\relax PMLR, 2017, pp. 1--16.

\bibitem{palanisamy2020multi}
P.~Palanisamy, ``Multi-agent connected autonomous driving using deep
  reinforcement learning,'' in \emph{2020 International Joint Conference on
  Neural Networks (IJCNN)}.\hskip 1em plus 0.5em minus 0.4em\relax IEEE, 2020,
  pp. 1--7.

\bibitem{chen2019novel}
S.~Chen, Y.~Chen, S.~Zhang, and N.~Zheng, ``A novel integrated simulation and
  testing platform for self-driving cars with hardware in the loop,''
  \emph{IEEE Transactions on Intelligent Vehicles}, vol.~4, no.~3, pp.
  425--436, 2019.

\bibitem{trumpp2022modeling}
R.~Trumpp, H.~Bayerlein, and D.~Gesbert, ``Modeling interactions of autonomous
  vehicles and pedestrians with deep multi-agent reinforcement learning for
  collision avoidance,'' in \emph{2022 IEEE Intelligent Vehicles Symposium
  (IV)}.\hskip 1em plus 0.5em minus 0.4em\relax IEEE, 2022, pp. 331--336.

\bibitem{vinitsky2022nocturne}
E.~Vinitsky, N.~Lichtl{\'e}, X.~Yang, B.~Amos, and J.~Foerster, ``Nocturne: a
  scalable driving benchmark for bringing multi-agent learning one step closer
  to the real world,'' \emph{Advances in Neural Information Processing
  Systems}, vol.~35, pp. 3962--3974, 2022.

\bibitem{highway-env}
E.~Leurent, ``An environment for autonomous driving decision-making,''
  \url{https://github.com/eleurent/highway-env}, 2018.

\bibitem{han2018safe}
J.~Han, A.~Sciarretta, L.~L. Ojeda, G.~De~Nunzio, and L.~Thibault, ``Safe-and
  eco-driving control for connected and automated electric vehicles using
  analytical state-constrained optimal solution,'' \emph{IEEE Transactions on
  Intelligent Vehicles}, vol.~3, no.~2, pp. 163--172, 2018.

\bibitem{rios2018impact}
J.~Rios-Torres and A.~A. Malikopoulos, ``Impact of partial penetrations of
  connected and automated vehicles on fuel consumption and traffic flow,''
  \emph{IEEE Transactions on Intelligent Vehicles}, vol.~3, no.~4, pp.
  453--462, 2018.

\bibitem{HUA2023121526}
M.~Hua, C.~Zhang, F.~Zhang, Z.~Li, X.~Yu, H.~Xu, and Q.~Zhou, ``Energy
  management of multi-mode plug-in hybrid electric vehicle using multi-agent
  deep reinforcement learning,'' \emph{Applied Energy}, vol. 348, p. 121526,
  2023.

\bibitem{yang2023multi}
N.~Yang, L.~Han, R.~Liu, Z.~Wei, H.~Liu, and C.~Xiang, ``Multi-objective
  intelligent energy management for hybrid electric vehicles based on
  multi-agent reinforcement learning,'' \emph{IEEE Transactions on
  Transportation Electrification}, 2023.

\bibitem{shi2022online}
W.~Shi, Y.~Huangfu, L.~Xu, and S.~Pang, ``Online energy management strategy
  considering fuel cell fault for multi-stack fuel cell hybrid vehicle based on
  multi-agent reinforcement learning,'' \emph{Applied Energy}, vol. 328, p.
  120234, 2022.

\bibitem{wang2023cooperative}
Y.~Wang, Y.~Wu, Y.~Tang, Q.~Li, and H.~He, ``Cooperative energy management and
  eco-driving of plug-in hybrid electric vehicle via multi-agent reinforcement
  learning,'' \emph{Applied Energy}, vol. 332, p. 120563, 2023.

\bibitem{zhang2023integrated}
H.~Zhang, J.~Peng, H.~Dong, F.~Ding, and H.~Tan, ``Integrated velocity
  optimization and energy management strategy for hybrid electric vehicle
  platoon: A multi-agent reinforcement learning approach,'' \emph{IEEE
  Transactions on Transportation Electrification}, 2023.

\bibitem{peng2023ecological}
J.~Peng, W.~Chen, Y.~Fan, H.~He, Z.~Wei, and C.~Ma, ``Ecological driving
  framework of hybrid electric vehicle based on heterogeneous multi agent deep
  reinforcement learning,'' \emph{IEEE Transactions on Transportation
  Electrification}, 2023.

\bibitem{shang2023estimation}
W.-L. Shang, M.~Zhang, G.~Wu, L.~Yang, S.~Fang, and W.~Ochieng, ``Estimation of
  traffic energy consumption based on macro-micro modelling with sparse data
  from connected and automated vehicles,'' \emph{Applied Energy}, vol. 351, p.
  121916, 2023.

\bibitem{rahman2021multi}
M.~H. Rahman, M.~Abdel-Aty, and Y.~Wu, ``A multi-vehicle communication system
  to assess the safety and mobility of connected and automated vehicles,''
  \emph{Transportation research part C: emerging technologies}, vol. 124, p.
  102887, 2021.

\bibitem{hasan2020securing}
M.~Hasan, S.~Mohan, T.~Shimizu, and H.~Lu, ``Securing vehicle-to-everything
  (v2x) communication platforms,'' \emph{IEEE Transactions on Intelligent
  Vehicles}, vol.~5, no.~4, pp. 693--713, 2020.

\bibitem{valiente2022robustness}
R.~Valiente, B.~Toghi, R.~Pedarsani, and Y.~P. Fallah, ``Robustness and
  adaptability of reinforcement learning-based cooperative autonomous driving
  in mixed-autonomy traffic,'' \emph{IEEE Open Journal of Intelligent
  Transportation Systems}, vol.~3, pp. 397--410, 2022.

\bibitem{wang2020learning}
R.~Wang, X.~He, R.~Yu, W.~Qiu, B.~An, and Z.~Rabinovich, ``Learning efficient
  multi-agent communication: An information bottleneck approach,'' in
  \emph{International Conference on Machine Learning}.\hskip 1em plus 0.5em
  minus 0.4em\relax PMLR, 2020, pp. 9908--9918.

\bibitem{freed2020communication}
B.~Freed, G.~Sartoretti, J.~Hu, and H.~Choset, ``Communication learning via
  backpropagation in discrete channels with unknown noise,'' in
  \emph{Proceedings of the AAAI conference on artificial intelligence},
  vol.~34, no.~05, 2020, pp. 7160--7168.

\bibitem{lei2023new}
Y.~Lei, D.~Ye, S.~Shen, Y.~Sui, T.~Zhu, and W.~Zhou, ``New challenges in
  reinforcement learning: a survey of security and privacy,'' \emph{Artificial
  Intelligence Review}, vol.~56, no.~7, pp. 7195--7236, 2023.

\bibitem{valiente2023learning}
R.~Valiente, B.~Toghi, M.~Razzaghpour, R.~Pedarsani, and Y.~P. Fallah,
  ``Learning-based social coordination to improve safety and robustness of
  cooperative autonomous vehicles in mixed traffic,'' in \emph{Machine Learning
  and Optimization Techniques for Automotive Cyber-Physical Systems}.\hskip 1em
  plus 0.5em minus 0.4em\relax Springer, 2023, pp. 671--707.

\bibitem{zhao2019identification}
S.~Zhao, G.~Chen, M.~Hua, and C.~Zong, ``An identification algorithm of driver
  steering characteristics based on backpropagation neural network,''
  \emph{Proceedings of the Institution of Mechanical Engineers, Part D: Journal
  of Automobile Engineering}, vol. 233, no.~9, pp. 2333--2342, 2019.

\bibitem{dai2023bargaining}
C.~Dai, C.~Zong, D.~Zhang, M.~Hua, H.~Zheng, and K.~Chuyo, ``A bargaining
  game-based human--machine shared driving control authority allocation
  strategy,'' \emph{IEEE Transactions on Intelligent Transportation Systems},
  2023.

\bibitem{gilroy2022scooter}
S.~Gilroy, D.~Mullins, E.~Jones, A.~Parsi, and M.~Glavin, ``E-scooter rider
  detection and classification in dense urban environments,'' \emph{Results in
  Engineering}, vol.~16, p. 100677, 2022.

\bibitem{zhao2020sim}
W.~Zhao, J.~P. Queralta, and T.~Westerlund, ``Sim-to-real transfer in deep
  reinforcement learning for robotics: a survey,'' in \emph{2020 IEEE symposium
  series on computational intelligence (SSCI)}.\hskip 1em plus 0.5em minus
  0.4em\relax IEEE, 2020, pp. 737--744.

\bibitem{yang2021believe}
Y.~Yang, X.~Ma, C.~Li, Z.~Zheng, Q.~Zhang, G.~Huang, J.~Yang, and Q.~Zhao,
  ``Believe what you see: Implicit constraint approach for offline multi-agent
  reinforcement learning,'' \emph{Advances in Neural Information Processing
  Systems}, vol.~34, pp. 10\,299--10\,312, 2021.

\end{thebibliography}

\end{document}